\documentclass[runningheads]{llncs}

\usepackage{graphicx}
\usepackage{amsmath,amssymb} % define this before the line numbering.
\usepackage{color}
\usepackage[T1]{fontenc}
\usepackage[latin9]{inputenc}
\usepackage{xcolor}
\usepackage{array}
\usepackage{verbatim}
\usepackage{units}
\usepackage{textcomp}
\usepackage{bbding}
\usepackage{url}
\usepackage{multirow}
\usepackage{times}
\usepackage{epsfig}
\usepackage{placeins}
\usepackage{colortbl}
\usepackage{rotating}
\usepackage{hyperref}

\definecolor{lightgray}{gray}{0.6}
\definecolor{verylightgray}{gray}{0.9}
\definecolor{myred}{rgb}{ 0.5020,0, 0}
\definecolor{mygreen}{rgb}{0,  0.5020, 0}
\definecolor{myyellow}{rgb}{ 0.5020,  0.5020, 0.1}
\definecolor{myblue}{rgb}{ 0.1,  1, 1}
\definecolor{myblue2}{rgb}{ 0.1,  0.1, 1}
\definecolor{mypink}{rgb}{ 1,  0.1, 1}

\usepackage{caption} 
\usepackage{hyphenat}
\usepackage[british]{babel}
\usepackage{tablefootnote}
\usepackage{wrapfig}
    \makeatletter
  \renewcommand\subsubsection{\@startsection{subsubsection}{3}{\z@}
  	{-3\p@ \@plus -4\p@ \@minus -4\p@}
  	{-0.5em \@plus -0.22em \@minus -0.1em}	
  	{\normalfont\normalsize\bfseries\boldmath}}
  
  \renewcommand\paragraph{\@startsection{paragraph}{4}{\z@}
  	{-3\p@ \@plus -4\p@ \@minus -4\p@}
  	{-0.5em \@plus -0.22em \@minus -0.1em} 	
  	{\normalfont\normalsize\itshape}}
  
  \renewcommand\section{\@startsection {section}{1}{\z@}%
  	{-1.8ex \@plus -4\p@ \@minus -4\p@}%
  	{1.2ex \@plus 4\p@ \@minus 4\p@}%
    	{\normalfont\large\bfseries\boldmath
  	\rightskip=\z@ \@plus 8em\pretolerance=10000 }}
  
  \renewcommand\subsection{\@startsection {subsection}{2}{\z@}%
  	{-1.2ex \@plus -4\p@ \@minus -4\p@}%
  	{1ex \@plus -4\p@ \@minus -4\p@}%
  {\normalfont\normalsize\bfseries\boldmath
  	\rightskip=\z@ \@plus 8em\pretolerance=10000 }}

   \makeatother

\begin{document}

\title{Video Object Segmentation with\\ Language Referring Expressions} % Replace with your title

\titlerunning{Video Object Segmentation with Language Referring Expressions}

\author{Anna Khoreva\inst{1,2} \and Anna Rohrbach\inst{3} \and Bernt Schiele\inst{1}}
%
%Please include author names in full in the paper, 
%If any authors have names that can be parsed into FirstName LastName in multiple ways, please include the correct parsing, in a comment to the volume editors:
%\index{Lastnames, Firstnames}

\authorrunning{A. Khoreva et al.} % A shorter version of authors' name

%===========================================================

%\institute{Max Planck Institute for Informatics, Germany \email{\{khoreva,schiele\}@mpi-inf.mpg.de} \and Bosch Center for Artificial Intelligence, Germany
%	 \and University of California, Berkeley \email{anna.rohrbach@berkeley.edu}}

\institute{\textsuperscript{1}Max Planck Institute for Informatics \hspace{1.5em}\textsuperscript{2}Bosch Center for Artificial Intelligence\\ \textsuperscript{3}University of California, Berkeley}
\maketitle

\begin{abstract}
Most state-of-the-art semi-supervised video object segmentation methods rely on a pixel-accurate mask of a target object provided for the first frame of a video. 
However, obtaining a detailed segmentation mask is expensive and time-consuming. 
In this work we explore an alternative way of identifying a target object, namely by employing language referring expressions. Besides being a more practical and natural way of pointing out a target object, 
using language specifications can help to avoid drift as well as make the system more robust to complex dynamics and appearance variations. 
Leveraging recent advances of language grounding models designed for images, we propose an approach to extend them to video data, ensuring temporally coherent predictions. 
To evaluate our approach we augment the popular video object segmentation  benchmarks, $\text{DAVIS}_{\text{16}}$ and $\text{DAVIS}_{\text{17}}$ with language descriptions of target objects. 
We show that our language-supervised approach performs on par with the methods which have access to a pixel-level mask of the target object on $\text{DAVIS}_{\text{16}}$ 
and is competitive to methods using scribbles on the challenging $\text{DAVIS}_{\text{17}}$ dataset.
%\keywords{Video Object Segmentation, Referring Expression Comprehension}
\end{abstract}

\begin{figure}
\begin{centering}

\includegraphics[width=0.9\textwidth]{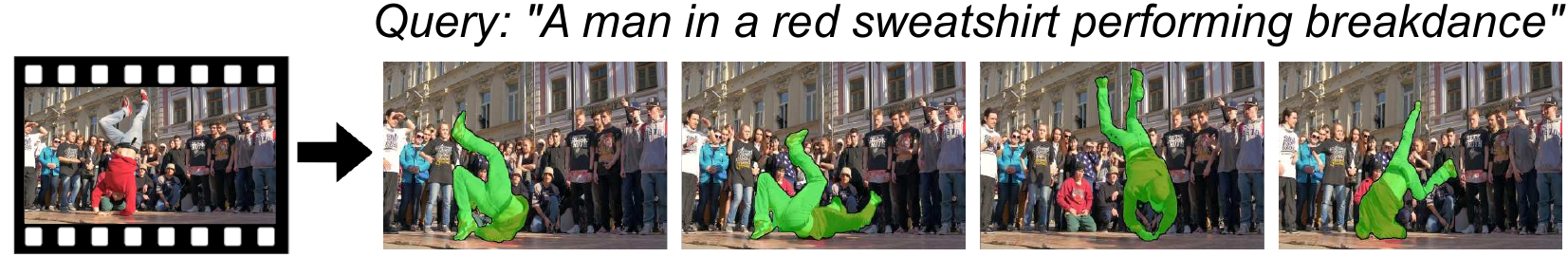}
\par\end{centering}
\caption{\label{fig:Teaser} Examples of the proposed approach. Classical semi-supervised video object segmentation relies on an expensive pixel-level mask annotation of a target object in the first frame of a video. 
We explore a more natural and more practical way of pointing out a target object by providing a language referring expression.}

\end{figure}

\section{Introduction}

Video object segmentation has recently witnessed growing interest \cite{Caelles2017Cvpr,Hu2017MaskRNNIL,Cheng_ICCV_2017,Pont-Tuset_arXiv_2018}. % considerable progress
Segmenting objects at pixel level provides a finer understanding of video
and is relevant for many applications, e.g. augmented reality, video editing, and rotoscoping. %, and summarisation. %and video-based advertisement

Ideally, one would like to obtain a pixel-accurate segmentation of objects in video with no human input during test time.
However, the current state-of-the-art
unsupervised video object segmentation methods \cite{xiao2016cvpr,Jain2017ArxivFusionSeg,TokmakovAS17} have troubles segmenting the target objects in videos containing multiple objects and cluttered backgrounds 
without any guidance from the user.
Hence, many recent works \cite{Caelles2017Cvpr,Hu2017MaskRNNIL,Voigtlaender2017OnlineAO} employ a semi-supervised approach, where a pixel-level mask of the target object is manually annotated in the first frame 
and the task is to accurately segment the object in successive frames.
Although this setting has proven to be successful, it can be prohibitive for many applications. It is tedious and time-consuming for the user to provide a pixel-accurate segmentation 
and usually takes more than a minute to annotate a single instance (\cite{lin2014eccvcoco} reports $79$s for polygon annotations, precisely delineating an object would take even more).
To make video object segmentation more applicable in practice, instead of costly pixel-level masks \cite{Pont-Tuset_arXiv_2018,Maninis_arxiv17,Benard_arxiv17} propose to employ point clicks or scribbles 
to specify the target object in the first frame. 
This is much faster and takes an annotator on average $7.5$s to label an object with point clicks \cite{Maninis_arxiv17} and $10$s with scribbles \cite{lin2016cvprscribblesup}. 
However, on small touchscreen devices, such as tablets or phones, providing precise clicks or drawing scribbles using fingers could be cumbersome and inconvenient for the user.
 
To overcome these limitations we propose a new task - segmenting objects in video using language referring expressions - which is a more natural way of human-computer interaction.
It is much easier for a user to say: ``Segment the man in a red sweatshirt performing breakdance'' (see Figure \ref{fig:Teaser}), 
than to provide a tedious pixel-level segmentation mask or struggle with drawing a scribble which does not straddle the object boundary. 
Moreover, employing language specifications can make the system more robust to background clutter, help to avoid drift 
and better adapt to the complex dynamics inherent to videos, while not over-fitting to a particular view in the first frame (see Table \ref{tab:comparative-result-VOS-mask-language}).

We aim to investigate the capabilities and limitations of existing techniques on the proposed task and explore how far one can go while leveraging the advances in image-level language grounding and pixel-level segmentation in videos.
We start by analyzing the performance of the state-of-the-art language grounding models \cite{2017-cvpr-dbnet,yu2018mattnet} for localization of objects in videos via bounding boxes. 
We discover that they suffer from a number of issues, predicting temporally inconsistent and jittery boxes, and show a way to enhance their predictions by enforcing temporal coherency (see Figure \ref{fig:qualitative-results-grounding}). 
Next we propose a convnet-based framework that utilizes referring expressions for video object segmentation task, 
where the output of the grounding model (bounding box) is used as a guidance for pixel-wise segmentation of the object. % in each video frame. 
We also show that video object segmentation using the mask annotation on the first frame can be further improved by using language supervision, highlighting the complementarity of both modalities.

To evaluate the proposed approach we extend the popular benchmarks for segmenting single and multiple objects in videos, $\text{DAVIS}_{\text{16}}$ \cite{Perazzi2016Cvpr} and $\text{DAVIS}_{\text{17}}$ \cite{Pont-Tuset_arXiv_2017},
with language descriptions of the target objects. 
We collect the annotations using two different settings, asking the annotators to provide a description of the target object based on the first frame only as well as on the full video. Future work may choose which setting they prefer to use.
On average each video has been annotated with $7.5$ referring expressions and it takes the annotator around $5$s to provide a referring expression for a target object.

Our language-supervised approach performs on par with semi-supervised methods which have access to the pixel-accurate object mask on $\text{DAVIS}_{\text{16}}$ and 
shows comparable results to the techniques that employ scribbles on the challenging $\text{DAVIS}_{\text{17}}$ dataset.

In summary, our contributions are the following.
We present a new task of segmenting objects in video using natural language referring expressions for which we augment two well-known video segmentation benchmarks with textual descriptions of target objects.
We conduct an extensive analysis of the performance of the state-of-the-art language grounding models on video data and propose a way to improve their temporal coherency. To the best of our knowledge we are the first to perform an analysis of transferability of image-based grounding models to video.
We show that high quality video object segmentation results can be obtained by employing language referring expressions, allowing a more natural and practical human-computer interaction. Moreover, we show that 
language descriptions are complementary to visual forms of supervision, such as masks, and can be exploited as an additional source of guidance for object segmentation. 
Thus, while proposing the new task and accompanying dataset, our work contributes the necessary benchmark analysis, a very competitive baseline and valuable insights for future work.
We hope our findings would further promote the research in the field of video object segmentation via language expressions and help to discover better techniques that can be used in realistic scenarios.

\section{\label{sec:Related-work} Related Work}

\subsection{Grounding natural language expressions}
There has been an increasing interest in the task of grounding natural language expressions over the last few years \cite{yu2016modeling,liu2017recurrent,Li_2018_CVPR}. 
We group the existing works by the type of visual domain: images and video.

\subsubsection{Image domain.} 
Grounding natural language expressions is a task of localizing a given expression in an image with a bounding box \cite{2017-cvpr-dbnet,mao2016generation} or a segmentation mask \cite{liu2017recurrent,Li_2018_CVPR}. 
Referring expression comprehension is a closely related task, 
where the goal is to localize the non-ambiguous referring expression. 
Most existing approaches rely on external bounding box proposals which are scored to determine the top scoring box as the correct region \cite{luo17cvpr,yu2018mattnet}. 
A few recent works explore methods of inferring object regions by proposal generation network \cite{chen2017query} or efficient subwindow search \cite{yeh2017interpretable}. 
Multiple existing approaches model relationships between objects present in the scene \cite{nagaraja16eccv,hu17cvpr}. In this work we choose two state-of-the-art grounding models for experimentation and analysis \cite{2017-cvpr-dbnet,yu2018mattnet}. 
DBNet \cite{2017-cvpr-dbnet} frames grounding as a classification task, where an expression and an image region serve as input and a binary classification decision is an output. 
A key component of this approach is utilization of negative expressions and image regions to ensure discriminative training. DBNet currently leads on Visual Genome \cite{krishnavisualgenome}. 
MattNet \cite{yu2018mattnet} is a modular network which ``softly'' decomposes referring expressions in three parts: subject, location, and relationship, each of which is processed by a different visual module. 
This allows MattNet to process referring expressions of general forms, as each module can be ``enabled'' or ``disabled'' depending on the expression. 
MattNet achieves top performance on RefCOCO(g/+) \cite{yu2016modeling,mao2016generation} both in terms of bounding box localization and pixel-wise segmentation accuracy.

\subsubsection{Video domain.} 
The progress made in image-level natural language grounding leads to an increasing interest in application to video. 
The recent work of \cite{LiCVPR2017} studies object tracking in video using language expressions. 
They introduce a dynamic convolutional layer, where a language query is used to predict visual convolutional filters. 
\cite{ORVideoGaze} addresses object tracking in video with the language descriptions and human gaze as input. 
Our work falls in the same line of research, as we are exploring natural language as input for video object segmentation. To the best of our knowledge, this is the first work to apply natural language to this task. A concurrent work by \cite{gavrilyuk2018actor} has addressed a task of actor/action segmentation in video based on sentence input. Their work focuses on seven classes of actors (adult, baby, etc.) and mostly action-oriented descriptions. In contrast, we consider arbitrary objects and unconstrained referring expressions.

\subsection{Video Object Segmentation}

Video object segmentation has witnessed considerable progress \cite{Papazoglou2013Iccv,Tsai2016Cvpr,TokmakovAS17,Koh_CVPR_2017,Caelles2017Cvpr,Voigtlaender2017OnlineAO}.
In the following, we group the related work into unsupervised and semi-supervised.

\subsubsection{Unsupervised methods.} 
Unsupervised methods assume no human input on the video during test time.
They aim to group pixels consistent in both appearance and motion and extract the most salient spatio-temporal object tube.
Several techniques exploit object proposals \cite{xiao2016cvpr,Koh_CVPR_2017}, saliency \cite{Faktor2014Bmvc} and optical flow \cite{Papazoglou2013Iccv}.
Convnet-based approaches \cite{Cheng_ICCV_2017,Jain2017ArxivFusionSeg,TokmakovAS17} cast video object segmentation as a foreground/background classification problem and feed to the network both appearance and motion cues.
Because these methods do not have any knowledge of the target object, they have difficulties in videos with multiple moving and dominant objects and cluttered backgrounds.

\subsubsection{Semi-supervised methods.} 
Semi-supervised methods assume human input for the first frame, 
either by providing a pixel-accurate mask \cite{Tsai2016Cvpr,Caelles2017Cvpr}, clicks \cite{Maninis_arxiv17} or scribbles \cite{Pont-Tuset_arXiv_2018},
and then propagate the information to the successive frames. 
Existing approaches focus on leveraging superpixels \cite{wen2015cvpr}, constructing graphical models \cite{Tsai2016Cvpr}, utilizing object proposals \cite{Perazzi2015Iccv}
or employing optical flow and long-term trajectories \cite{Wang2017ArxivSTV}.
Lately, convnets have been considered for the task \cite{Caelles2017Cvpr,Khoreva2017CvprMaskTrack,Voigtlaender2017OnlineAO}. 
These methods usually build the architecture upon the semantic segmentation networks \cite{Long2015Cvpr} and process each frame of the video individually.
\cite{Caelles2017Cvpr} proposes to fine-tune a pre-trained generic object segmentation network on the first frame mask of the test video
to make it sensitive to the target object. \cite{Khoreva2017CvprMaskTrack} employs a similar strategy, but also provides a temporal context by feeding the previous
frame mask to the network.
Several methods extend the work of \cite{Caelles2017Cvpr} by incorporating the semantic information \cite{Caelles2017SemanticallyGuidedVO} or 
by integrating online adaptation \cite{Voigtlaender2017OnlineAO}.
\cite{Hu2017MaskRNNIL} proposes to employ a recurrent network to exploit the long-term temporal information. 

The above methods employ a pixel-level mask on the first frame. However, for many applications, particularly on small touchscreen devices, it can be prohibitive to provide a pixel-accurate segmentation.
Hence, there has been a growing interest to integrate cheaper forms of supervision, such as point clicks \cite{Benard_arxiv17,Maninis_arxiv17} or scribbles \cite{Pont-Tuset_arXiv_2018}, 
into convnet-based techniques. In spirit with these approaches, we aim to reduce the annotation effort on the first frame by using language referring expressions to specify the object.
Our approach also builds upon convnets and exploits both linguistic and visual modalities.

\section{\label{sec:Method}Method}

In this section we provide an overview of the proposed approach.
Given a video $V =\{f_1, . . . , f_N\}$ with
N frames and a textual query of the target object $Q$, our aim is to obtain a pixel-level segmentation mask of the target object in
every frame that it appears. 

\begin{figure}[t]

\begin{centering}
\includegraphics[width=0.85\textwidth]{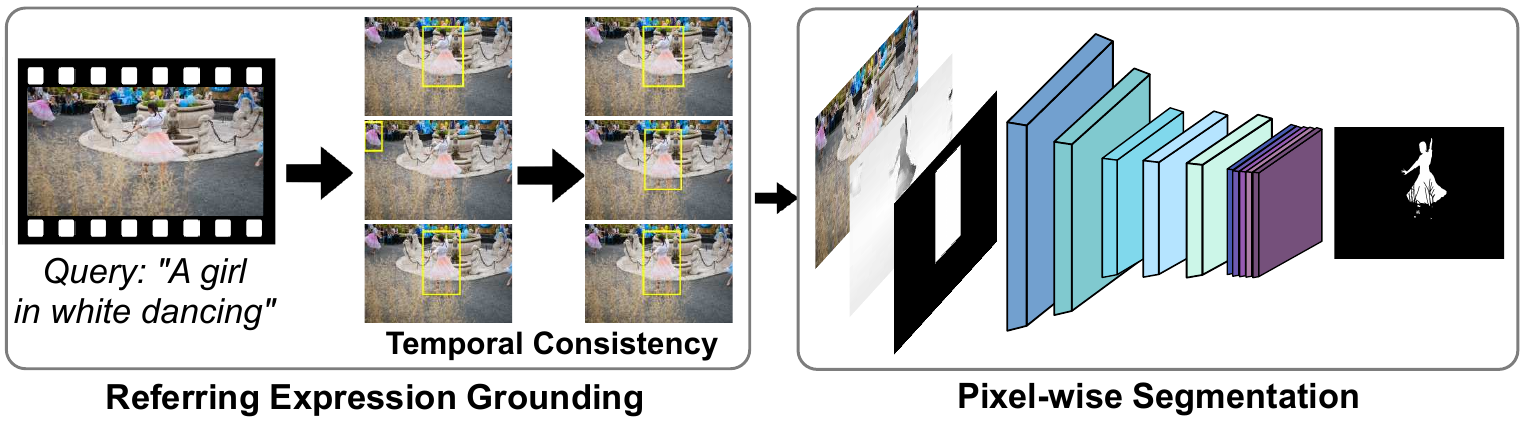}
\par\end{centering}
\caption{\label{fig:System} System overview. We first localize the target object via grounding model using the given referring expression and enforce
temporal consistency of bounding boxes across frames. Next we apply a segmentation convnet to recover detailed object masks.}

\end{figure}

We leverage recent advances in grounding referring expressions in images \cite{2017-cvpr-dbnet,yu2018mattnet} and pixel-level segmentation in videos \cite{Khoreva2017CvprMaskTrack,Jain2017ArxivFusionSeg}.
Our method consists of two main steps (see Figure \ref{fig:System}). 
Using as input the textual query $Q$ provided by the user, we first generate target object bounding box proposals for every frame of the video by exploiting referring expression grounding models, designed for images only. 
%In particular, we experiment with two models, DBNet \cite{2017-cvpr-dbnet} and MattNet \cite{yu2018mattnet}.
Applying these models off-the-shelf results in temporally inconsistent and jittery box predictions (see Figure \ref{fig:qualitative-results-grounding}).
Therefore, to mitigate this issue and make them more applicable for video data, we next employ temporal consistency, which enforces bounding boxes to be coherent across frames. 
As a second step, using as guidance the obtained box predictions of the target object on every frame of the video we apply a convnet-based pixel-wise segmentation model to recover detailed object masks in each frame.

\subsection{\label{subsec:Grounding}Grounding objects in video by referring expressions} 
As discussed in \S\ref{sec:Related-work}, the task of natural language grounding is to automatically localize a region described by a given language expression.
It is typically formulated as measuring the compatibility between a set of object proposals $O =\{o_{i}\}_{i=1}^{M}$ and a given textual query $Q$. 
The grounding model provides as output a set of matching scores $S =\{s_{i}\}_{i=1}^{M}$ between a box proposal and a textual query $Q$.
The box proposal with the highest matching score is selected as the predicted region.

We employ two state-of-the-art referring expression grounding models -- DBNet \cite{2017-cvpr-dbnet} and MattNet \cite{yu2018mattnet}, to localize the object in each frame.
Mask R-CNN \cite{He_2017_ICCV} bounding box proposals are exploited as an initial set of proposals for both models, although originally DBNet has been designed to utilize EdgeBox proposals \cite{Dollar2015Pami}.
However, using the grounding models designed for images and picking the highest scoring proposal for each video frame lead to temporally incoherent results. 
Even with simple textual queries for adjacent frames that from a human perspective look very much alike, the referring model often outputs inconsistent predictions (see Figure \ref{fig:qualitative-results-grounding}).
This indicates the inherent instability of the grounding models trained on the image domain.
To resolve this problem we propose to re-rank the object proposals by exploiting temporal structure along with the original matching scores given by a grounding model.

\subsubsection{Temporal consistency.} 
The goal of the temporal smoothing step is to improve temporal consistency and to reduce id-switches for target object predictions across frames.
Since objects tend to move smoothly through space and in time, there should be little changes from frame to frame and the box proposals should have high overlap between neighboring frames.
By finding temporally coherent tracks of an object that are spread-out in time, we can focus on the predictions that consistently appear
throughout the video and give less emphasis to objects that appear for only a short period of time.

The grounding model provides the likeliness of each box proposal to be the target object by outputting a matching score $s_{i}$.
Then each box proposal is re-ranked based on its overlap with the proposals in other frames, the original objectness score given by \cite{He_2017_ICCV} and its matching score from the grounding model.
Specifically, for each proposal we compute a new score: $\hat{s}_{i}=s_{i}*(\sum_{j=1, j\neq i}^{M} r_{ij}*d_{j}*s_{j}/t_{ij})$, where $r_{ij}$ measures an intersection-over-union ratio between box proposals $i$ and $j$, 
$t_{ij}$ denotes the temporal distance between two proposals ($t_{ij}= |f_i-f_j|$) and $d_j$ is the original objectness score. Then, in each frame we select the proposals with the highest new score.
The new scoring rewards temporally coherent predictions which likely belong to the target object and form a spatio-temporal tube. This step allows to improve temporal coherence boosting grounding and video segmentation performance (see Table \ref{tab:comparative-result-grounding-davis16-17} in \S\ref{subsec:Grounding-results} and Table \ref{tab:ablation-study} in \S\ref{subsec:VOS-results}) while being computational efficient (takes only a fraction of second).

\subsection{\label{subsec:Segmentation}Pixel-level video object segmentation}

We next show how to output pixel-level object masks, exploiting the bounding boxes from grounding as a guidance for the segmentation network.
The boxes are used as the input to the network to guide the network towards the target object, providing its rough location and extent. The task of the network
is to obtain a pixel-level foreground/background segmentation mask using appearance and motion cues.

\subsubsection{Approach.} 
We model pixel-level segmentation as a box refinement task.
The bounding box is transformed into a binary image (255 for the interior of the box, 0 for the background) and concatenated with the
RGB channels of the input image and optical flow magnitude, forming a 5-channel input for the network.  
Thus we ask the network to learn to refine the provided boxes into accurate masks.
Fusing appearance and motion cues allows to better exploit video data and handle better both static and moving objects.

We make one single pass over the video, applying the model per-frame. The network does not keep a notion of the specific appearance of the object in contrast to \cite{Khoreva2017CvprMaskTrack,Caelles2017Cvpr}, 
where the model is fine-tuned during the test time
to learn the appearance of the target object. Neither do we do an online adaptation as in \cite{Voigtlaender2017OnlineAO}, where the model is updated on its previous predictions while processing video frames.
This makes the system more efficient during the inference time, which is more suitable for real-world applications.

Similar to \cite{Khoreva2017CvprMaskTrack}, we train the network on static images, employing the saliency segmentation dataset \cite{cheng2015pami} which contains a diverse set of objects.  
The bounding box is obtained from the ground truth masks.
To make the system robust during test time to sloppy boxes from the grounding model, we augment the ground truth box by randomly jittering its coordinates (uniformly, $\pm 20\%$ of the original box width and height).
We synthesize optical flow from static images by applying affine transformations for both background and foreground object to simulate the camera and object motion in the neighboring frames, as in \cite{khoreva_lucid_dreams17}.
This simple strategy allows us to train on diverse set of static images, while exploiting motion information during test time.
We train the network on many triplets of RGB images, synthesized flow magnitude images and loose boxes in order for the model generalize well to different localization quality of boxes given by the grounding model 
and different dynamics of the object. 

During inference we use the state-of-the-art optical flow estimation method Flow{-}Net2.0 \cite{ilgcvpr17}. We compute the optical flow
magnitude by subtracting the median motion for each frame and averaging the magnitude of the forward and backward flow. 
The obtained image is further scaled to [0; 255] to maintain the same range as RGB channels.

\begin{figure*}[t]

\begin{centering}
\setlength{\tabcolsep}{0em}
\renewcommand{\arraystretch}{0}
\par\end{centering}
\begin{centering}

\hfill{}%
\begin{tabular}{c@{}c@{}c@{ \hskip 0.08in }c@{}c@{}c}

%\multicolumn{6}{c}{} \tabularnewline
\multicolumn{6}{c}{\textit{ Query: "A woman with a stroller."} } \tabularnewline
\includegraphics[width=0.15\linewidth]{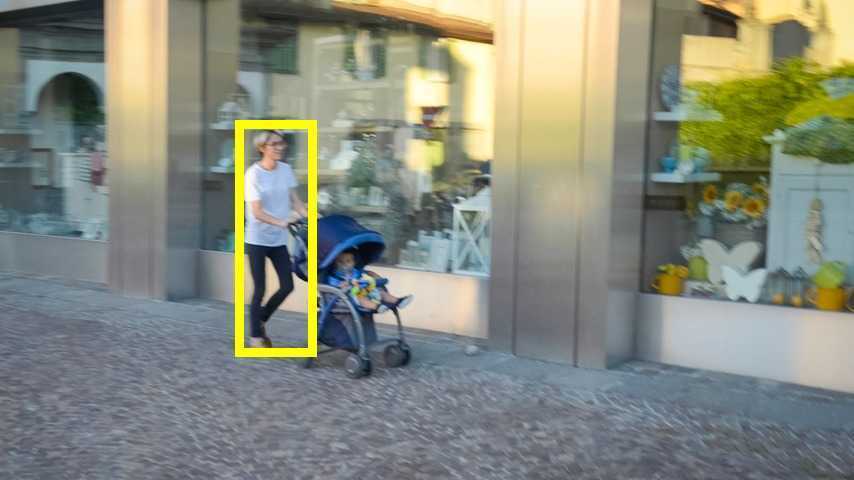} & {\footnotesize{}}\includegraphics[width=0.15\textwidth]{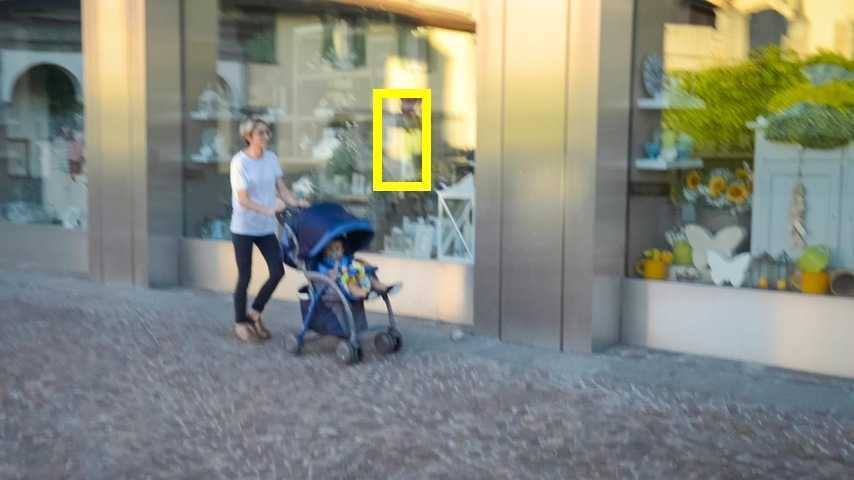} & {\footnotesize{}}\includegraphics[width=0.15\textwidth]{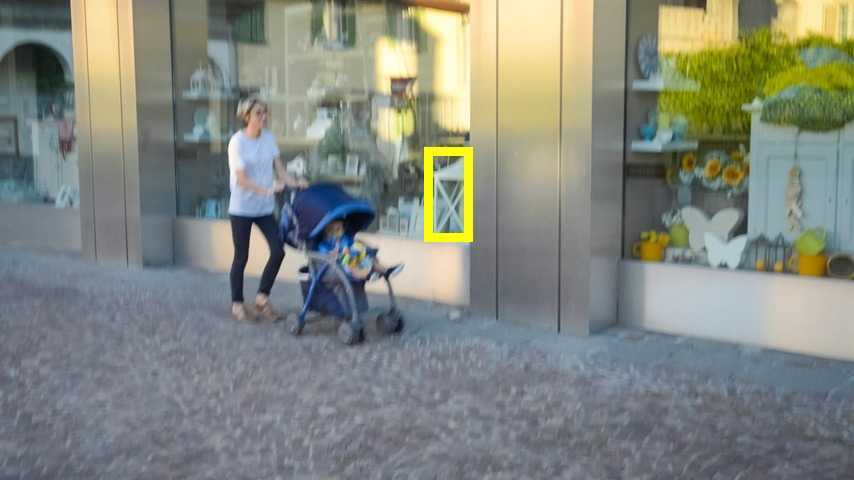} 
& \includegraphics[width=0.15\linewidth]{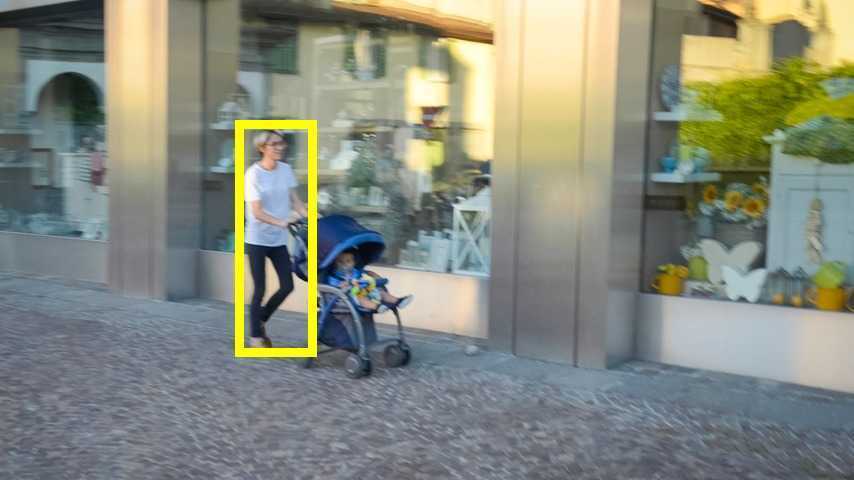} & {\footnotesize{}}\includegraphics[width=0.15\textwidth]{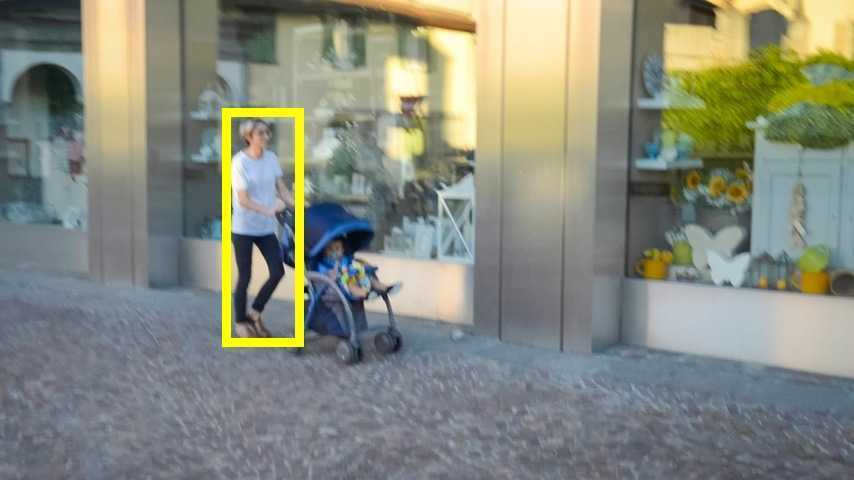} & {\footnotesize{}}\includegraphics[width=0.15\textwidth]{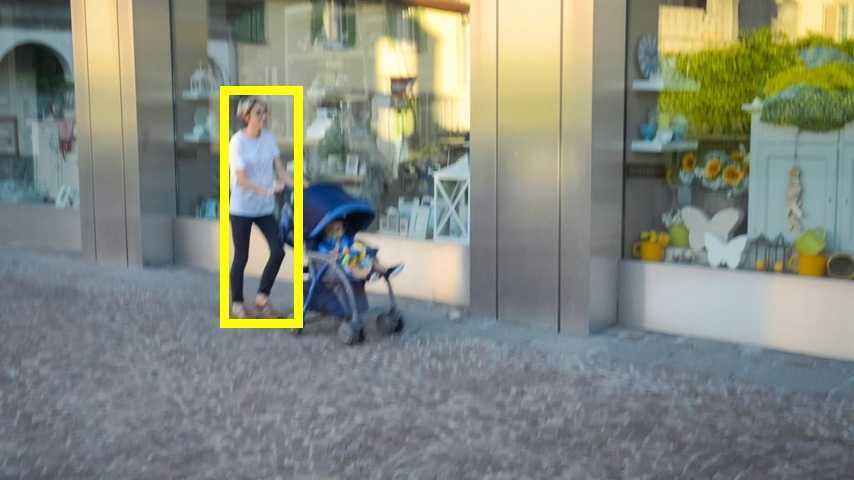}\tabularnewline

\multicolumn{6}{c}{\textit{ Query: "A girl riding a horse."}}  \tabularnewline
\includegraphics[width=0.15\linewidth]{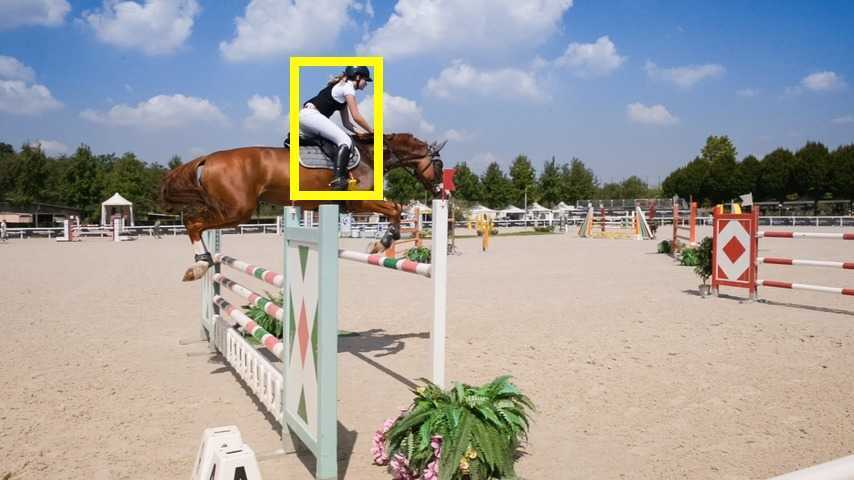} & {\footnotesize{}}\includegraphics[width=0.15\textwidth]{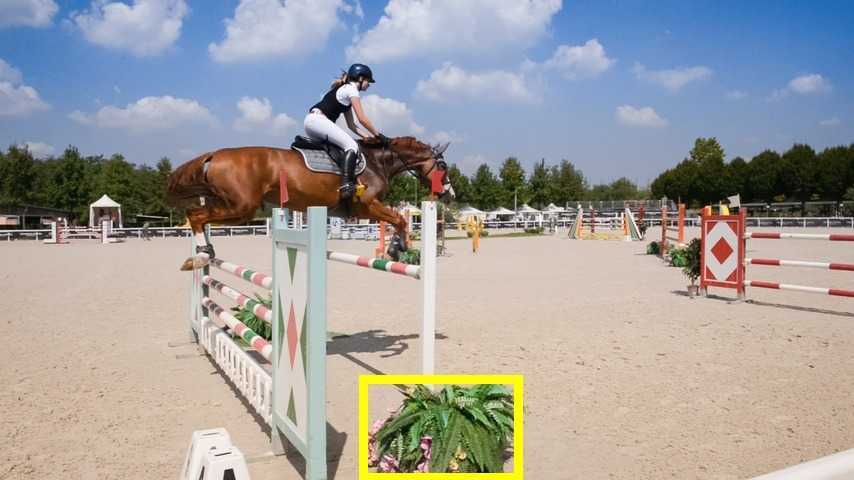} & {\footnotesize{}}\includegraphics[width=0.15\textwidth]{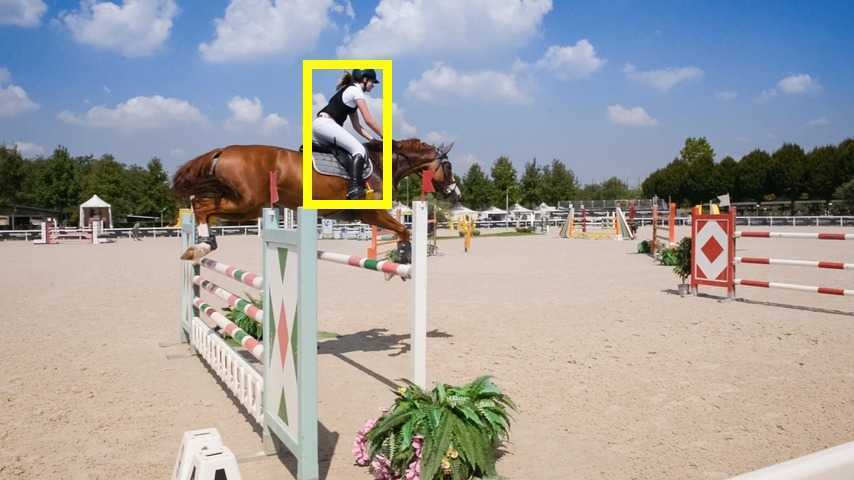} 
& \includegraphics[width=0.15\linewidth]{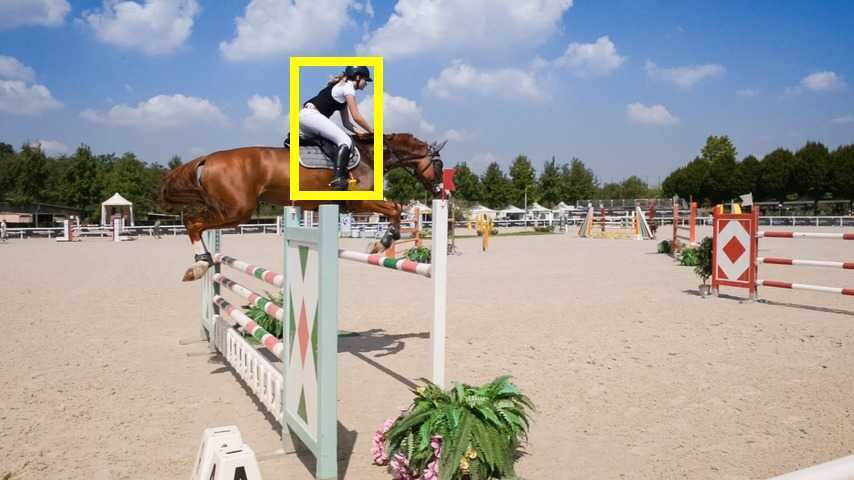} & {\footnotesize{}}\includegraphics[width=0.15\textwidth]{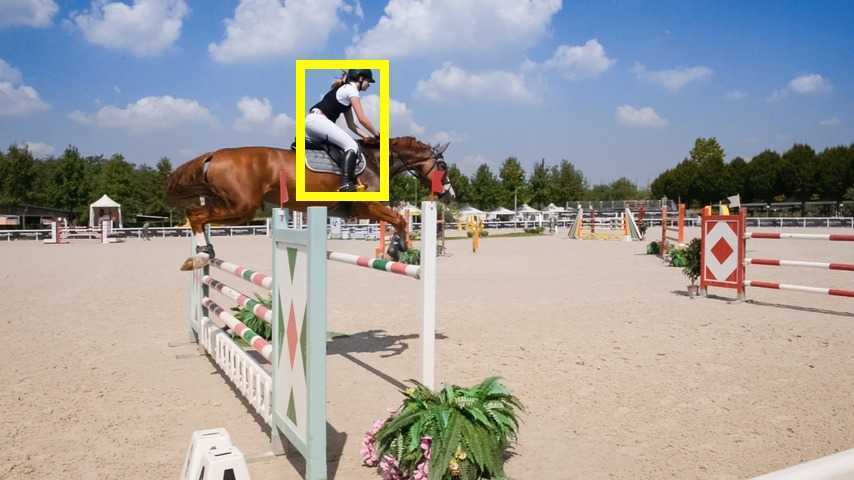} & {\footnotesize{}}\includegraphics[width=0.15\textwidth]{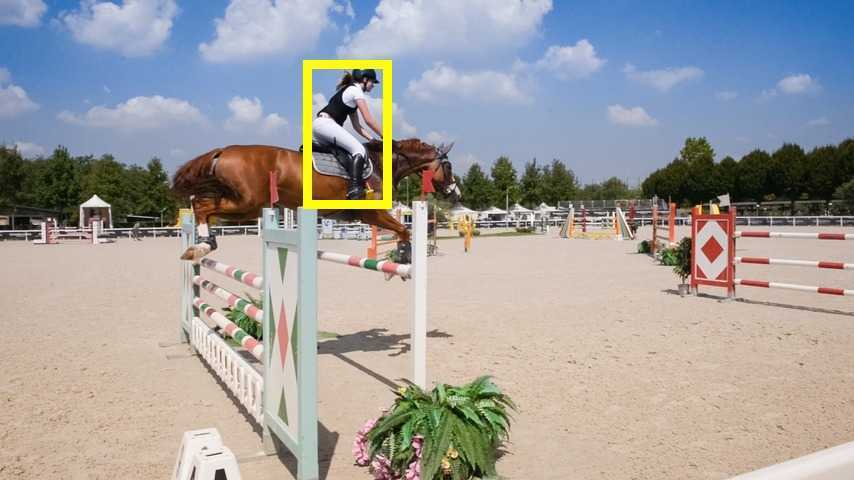}\tabularnewline

\multicolumn{3}{c} {W/o temporal consistency} & \multicolumn{3}{c} {With temporal consistency} \tabularnewline
\end{tabular}\hfill{}
\par\end{centering}
\caption{\label{fig:qualitative-results-grounding}Qualitative results of language grounding with and w/o temporal consistency
on $\text{DAVIS}_{\text{17}}$. The results are obtained using MattNet \cite{yu2018mattnet} trained on RefCOCO \cite{yu2016modeling}.}

\end{figure*}

\subsubsection{Network.} 
As our network architecture we use ResNet-101 \cite{he2016cvpr}. We adapt the network to the segmentation task following the procedure of \cite{Long2015Cvpr} and employing atrous convolutions \cite{chen2016arxivdeeplabv2} with hybrid rates \cite{wang2017understanding} within the last two blocks of
ResNet to enlarge the receptive field as well as to alleviate the "gridding" issue.
After the last block, we apply spatial pyramid pooling \cite{chen2016arxivdeeplabv2}, which aggregates features at
multiple scales by applying atrous convolutions with different rates, and augment it with the image-level features \cite{LiuRB15} to exploit better global context.
The network is trained using a standard cross-entropy loss (all pixels are equally weighted).
The final logits are upsampled to the ground truth resolution to preserve finer details for back-propagation.

For network initialization we use a model pre-trained on ImageNet \cite{he2016cvpr}. The new layers are initialized using the "Xavier"
strategy \cite{glorot10understandingthe}.
The network is trained on MSRA \cite{cheng2015pami} for segmentation.
To avoid the domain shift we fine-tune the model on the training sets of $\text{DAVIS}_{\text{16}}$ \cite{Perazzi2016Cvpr} and $\text{DAVIS}_{\text{17}}$ \cite{Pont-Tuset_arXiv_2017} respectively. 
We employ SGD with a polynomial learning policy with initial learning rate of $0.001$, crop size of $513\times513$, random scale data augmentation (from $0.5$ to $2.0$) and left-right flipping during training. 
The network is trained for $20k$ iterations on MSRA and $20k$ iterations on the training set of $\text{DAVIS}_{\text{16}}$/$\text{DAVIS}_{\text{17}}$.
During inference we employ test time augmentation as in \cite{chen2016arxivdeeplabv2}.

\subsubsection{Other sources of supervision.} 
Additionally we consider variants of the proposed model using different sources of supervision.
Our approach is flexible and can take advantage of the first frame mask annotation as well as language. 
We describe how language can be used on top of the mask supervision, improving the robustness of the system against occlusions and dynamic backgrounds (see \S\ref{subsec:VOS-results} for results).

\paragraph{Mask.} 
Here we discuss a variant that uses only the first frame mask supervision during test time.
The network is initialized with the bounding box obtained from the object mask in the 1st frame and for successive frames uses the prediction from the preceding frame warped with the optical flow (as in \cite{Khoreva2017CvprMaskTrack}) 
to get the input box for the next frame. Following \cite{Khoreva2017CvprMaskTrack,Caelles2017Cvpr} we fine-tune the model for $1k$ iterations on an augmented set obtained 
from the first frame image and mask, to learn the specific properties of the object.

\paragraph{Mask + Language.} 
We show that using language supervision is complementary to the first frame mask.
Instead of relying on the preceding frame prediction as in the previous paragraph, we use the bounding boxes obtained from the grounding model after the temporal consistency step.
We initialize with the ground truth box in the first frame and fine-tune the network on the 1st frame.

\section{\label{sec:Datasets}Collecting referring expressions for video}

Our task is to localize and provide a pixel-level mask of an object on all video frames given a language referring expression obtained either by looking at the first frame only or the full video. To validate our approach we employ two popular video object segmentation datasets, $\text{DAVIS}_{\text{16}}$ \cite{Perazzi2016Cvpr} and $\text{DAVIS}_{\text{17}}$ \cite{Pont-Tuset_arXiv_2017}. 
These two datasets introduce various challenges, containing videos with single or multiple salient objects, crowded scenes, similar looking instances, occlusions, camera view changes, fast motion, etc. 

\begin{figure} [t]

\begin{centering}
\hfill{}
\begin{tabular}{c@{\hskip 0.01in}c@{\hskip 0.01in}c@{\hskip 0.01in}c@{\hskip 0.01in}c@{\hskip 0.01in}c}
\includegraphics[width=0.16\linewidth]{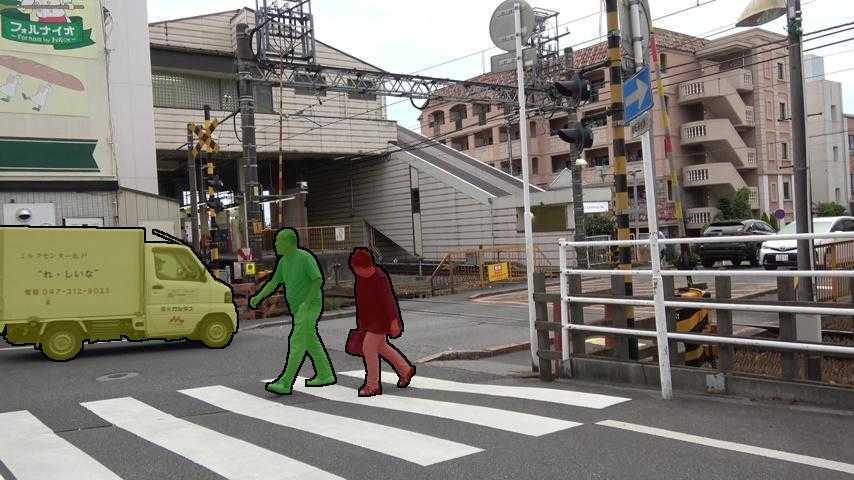} & {\footnotesize{}}
\includegraphics[width=0.16\linewidth]{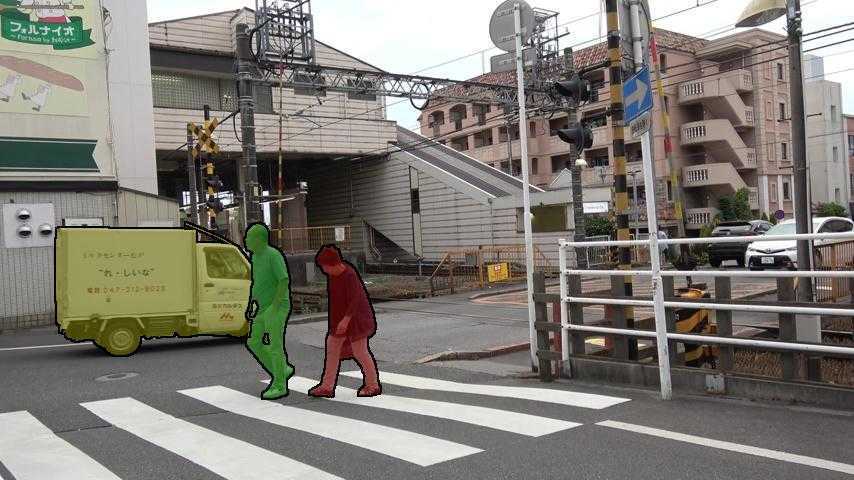} & {\footnotesize{}}
\includegraphics[width=0.16\linewidth]{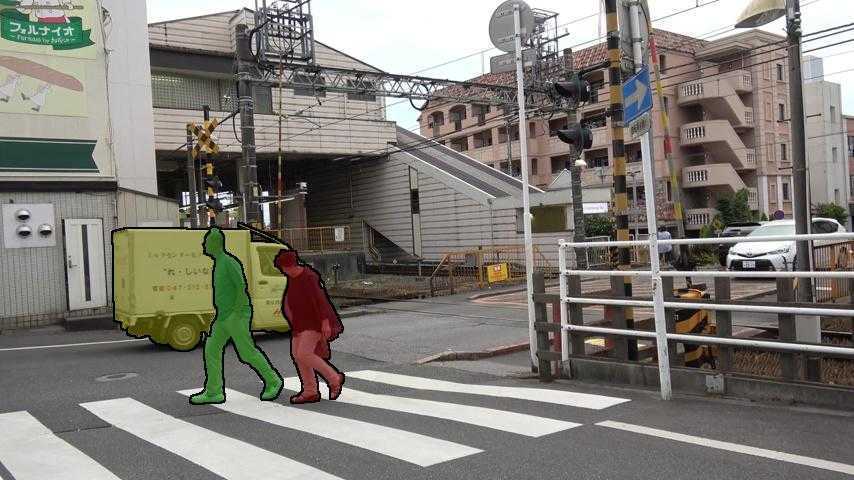} & {\footnotesize{}}
\includegraphics[width=0.16\linewidth]{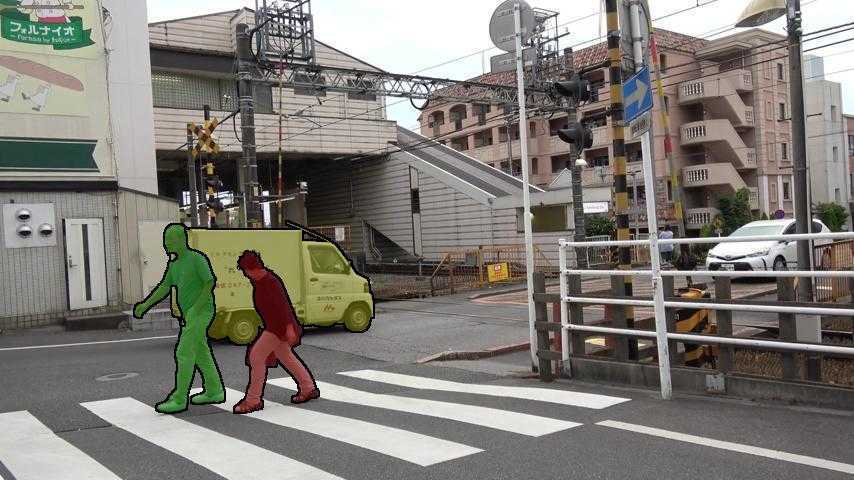} & {\footnotesize{}}
\includegraphics[width=0.16\linewidth]{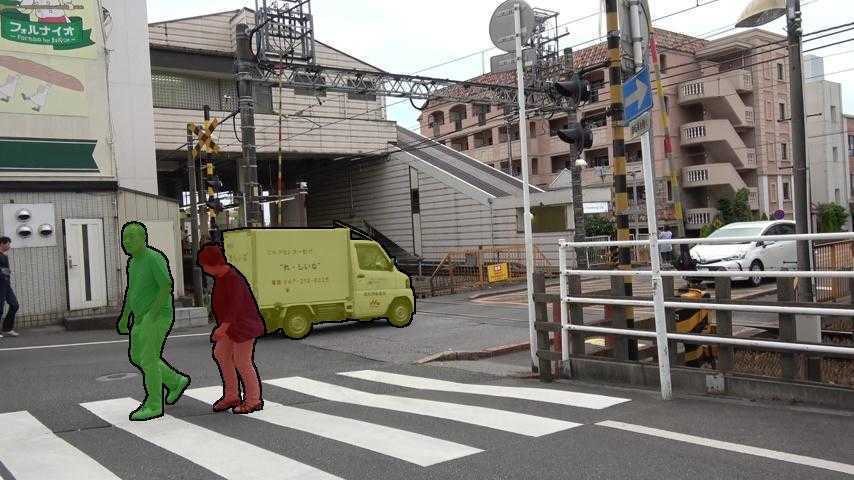} & {\footnotesize{}}
\includegraphics[width=0.16\linewidth]{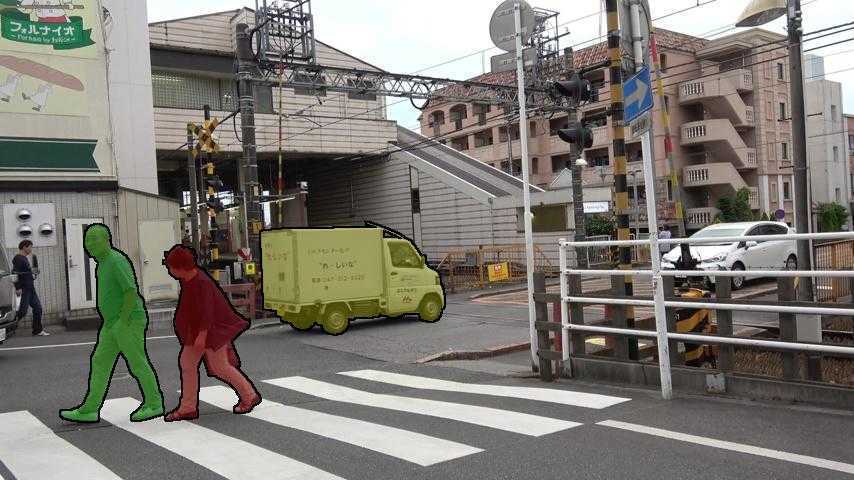}\tabularnewline
\multicolumn{3}{c}{\scriptsize{\textit{\textcolor{mygreen}{ID 1}: "A man in a grey t-shirt and yellow trousers"}} } & \multicolumn{3}{c}{\scriptsize{\textit{\textcolor{mygreen}{ID 1}: "A man in a grey shirt walking through the crossing"}} } \tabularnewline
\multicolumn{3}{c}{\scriptsize{\textit{\textcolor{myred}{ID 2}: "A woman in a black shirt"}} } & \multicolumn{3}{c}{\scriptsize{\textit{ \textcolor{myred}{ID 2}: "A woman walking through the crossing"}} } \tabularnewline
\multicolumn{3}{c}{\scriptsize{\textit{\textcolor{myyellow}{ID 3}: "A white truck on the road"}} } & \multicolumn{3}{c}{\scriptsize{\textit{\textcolor{myyellow}{ID 3}: "A white truck moving from the left to right"}} } \tabularnewline

\multicolumn{3}{c}{\footnotesize{First frame annotation}} & \multicolumn{3}{c}{\footnotesize{Full video annotation}} \tabularnewline
\end{tabular}\hfill{}
\par\end{centering}

\caption{\label{fig:Lang_annot} Example of annotations provided for the 1st frame vs. the full video.
Full video annotations include descriptions of activities and overall are more complex.
}

\end{figure}

$\text{DAVIS}_{\text{16}}$ \cite{Perazzi2016Cvpr} consists of 30 training and 20 test videos of diverse object categories
with all frames annotated with pixel-level accuracy. Note that in this dataset only a single object is annotated per video.
For the multiple object video segmentation task we consider $\text{DAVIS}_{\text{17}}$. Compared
to $\text{DAVIS}_{\text{16}}$, this is a more challenging dataset, with multiple objects annotated per video and more complex scenes with more distractors,
occlusions, smaller objects, and fine structures.
Overall, $\text{DAVIS}_{\text{17}}$ consists of a training set with $60$ videos, and a validation/test-dev/test-challenge
set with $30$ sequences each. 

As our goal is to segment objects in videos using language specifications, we augment all objects annotated with mask labels in $\text{DAVIS}_{\text{16}}$ and $\text{DAVIS}_{\text{17}}$
with non-ambigu{-}ous referring expressions.
We follow the work of \cite{mao2016generation} and ask the annotator to provide a language description of the object, which has a mask annotation, by looking only at the first frame of the video.
Then another annotator is given the first frame and the corresponding description, and asked to identify the referred object. If the annotator is unable to correctly identify the object, 
the description is corrected to remove ambiguity and to specify the object uniquely.
We have collected two referring expressions per target object annotated by non-computer vision experts (Annotator 1, 2).

However, by looking only at the 1st frame, the obtained referring expressions may potentially be invalid for an entire video. (We actually quantified that only$\sim 15\%$ of the collected descriptions become invalid over time and it does not affect strongly segmentation results as temporal consistency step helps to disambiguate some of such cases, see the supp. material for details.) 
Besides, in many applications, such as video editing or video-based advertisement, the user has access to a full video. Providing a language query which is valid for all frames might decrease the editing time and result in more coherent predictions. 
Thus, on $\text{DAVIS}_{\text{17}}$ we asked the workers to provide a description of the object by looking at the full video. 
We have collected one expression of the full video type per target object. Future work may choose to use either setting.

The average length for the first frame/full video expressions is $5.5/6.3$ words. 
For $\text{DAVIS}_{\text{17}}$ first frame annotations we notice that descriptions given by Annotator 1 are longer than the ones by Annotator 2 ($6.4$ vs. $4.6$ words). 
We evaluate the effect of description length on the grounding performance in \S\ref{subsec:Grounding-results}. 
Besides, the expressions relevant to a full video mention verbs more often than the first frame descriptions 
($44\%$ vs. $25\%$). This is intuitive, as referring to an object which changes its appearance and position over time may require mentioning its actions. 
Adjectives are present in over $50\%$ for all annotations. Most of them refer to colors (over $70\%$), shapes and sizes ($7\%$) and spatial/ordering words ($6\%$ first frame vs. $13\%$ full video expressions). 
The  full video expressions also have a higher number of adverbs and prepositions, and overall are more complex than the ones provided for the first frame, see Figure \ref{fig:Lang_annot} for examples.

Overall augmented $\text{DAVIS}_{\text{16/17}}$ contains $\sim1.2$k referring expressions for more than $400$ objects on $150$ videos with $\sim10$k frames. We believe the collected data will be of interest to segmentation as well as vision and language communities, providing an opportunity to explore language as alternative input for video object segmentation.

\begin{figure*}[t]
\begin{centering}
\setlength{\tabcolsep}{0.1em}
\renewcommand{\arraystretch}{0}
\par\end{centering}
\begin{centering}

\hfill{}%
\begin{tabular}{c@{\hskip 0.05in}c@{\hskip 0.05in}c@{\hskip 0.05in}c@{\hskip 0.05in}c@{\hskip 0.05in}c}

\multicolumn{6}{c}{\textit{\textcolor{mygreen}{ID 1}: "A girl with blonde hair dressed in blue".} } \tabularnewline
\includegraphics[width=0.15\linewidth]{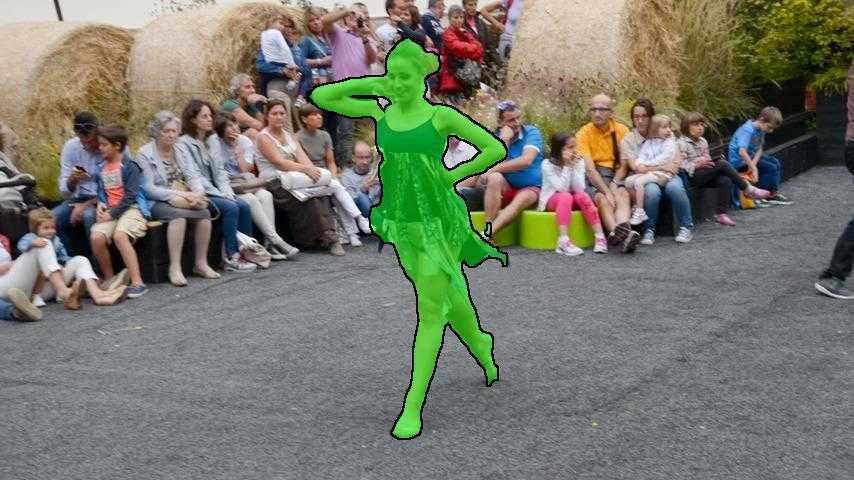} & {\footnotesize{}}
\includegraphics[width=0.15\linewidth]{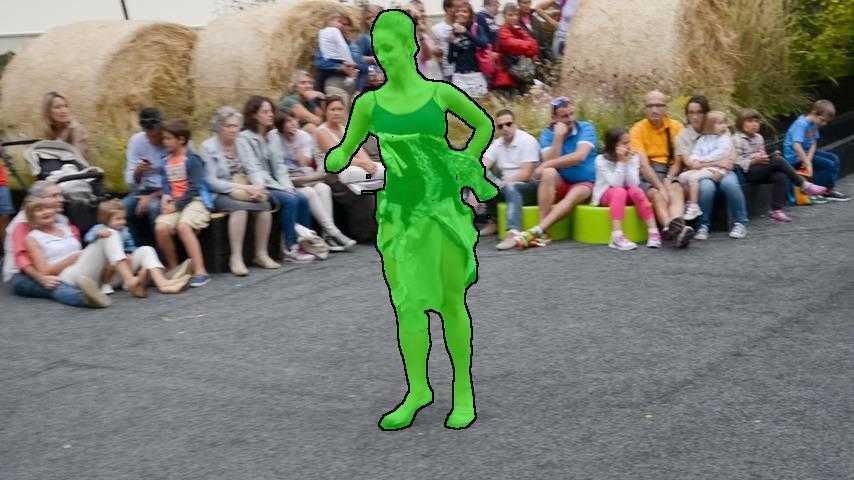} & {\footnotesize{}}
\includegraphics[width=0.15\linewidth]{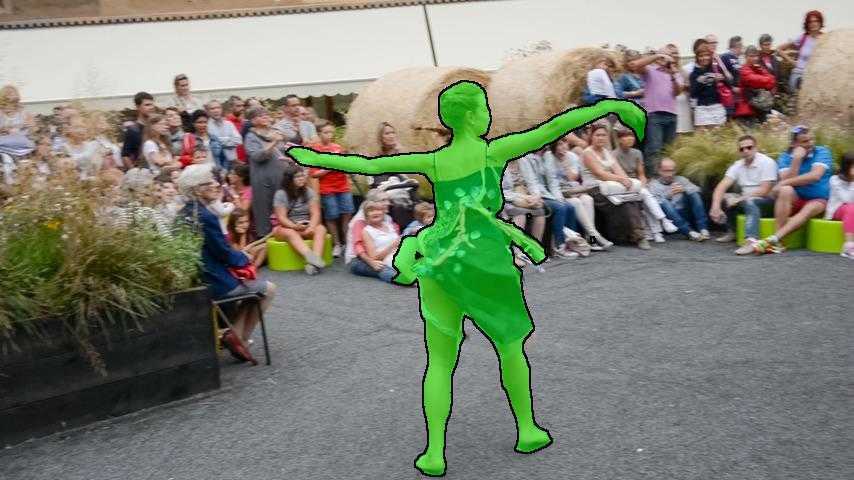} & {\footnotesize{}}
\includegraphics[width=0.15\linewidth]{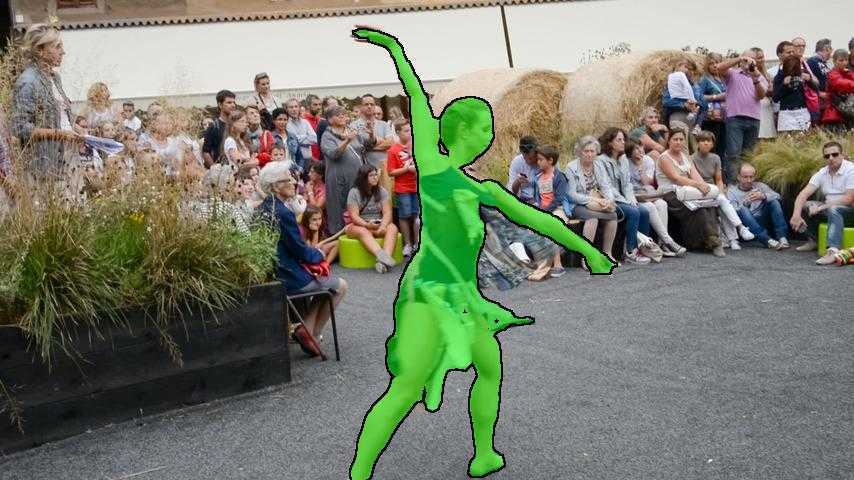} & {\footnotesize{}}
\includegraphics[width=0.15\linewidth]{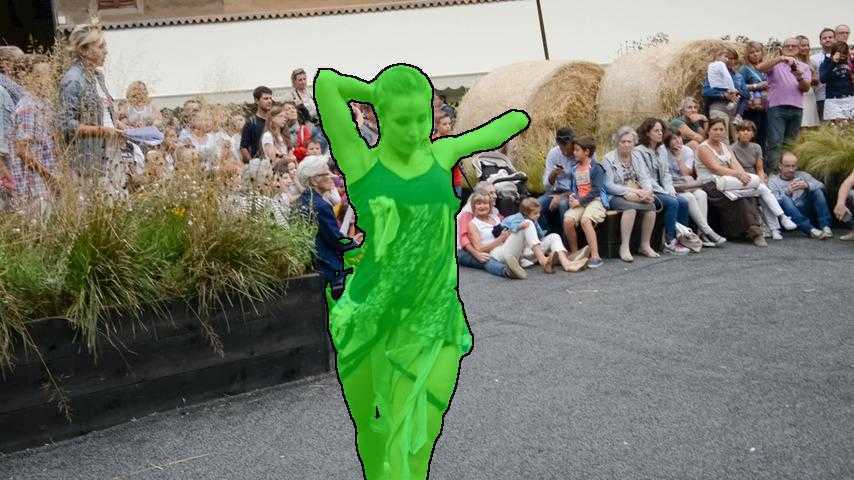} & {\footnotesize{}}
\includegraphics[width=0.15\linewidth]{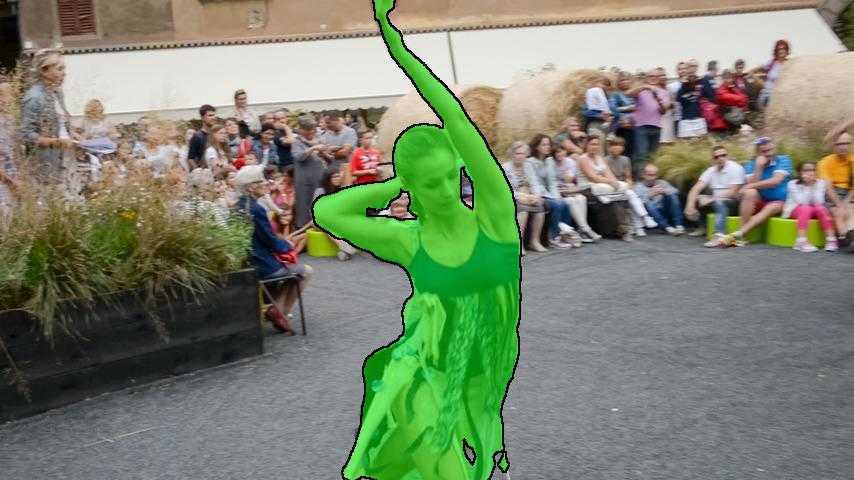}\tabularnewline

\multicolumn{6}{c}{\textit{\textcolor{mygreen}{ID 1}: "A brown camel in the front".} } \tabularnewline
\includegraphics[width=0.15\linewidth]{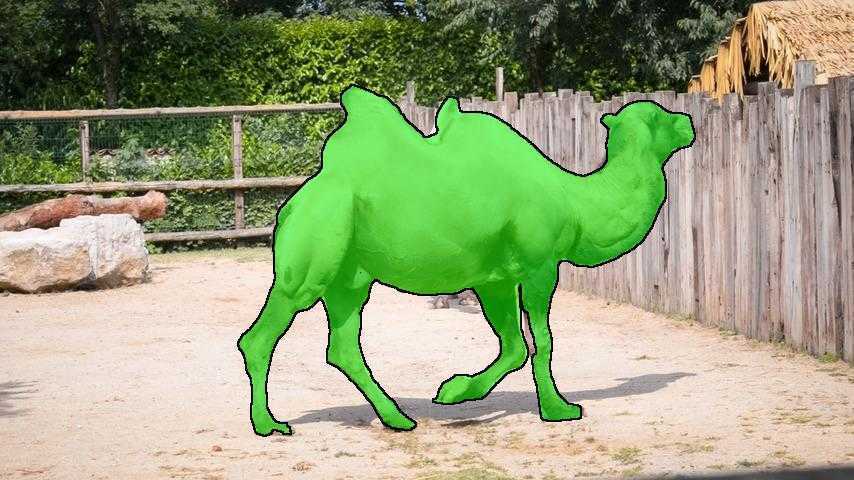} & {\footnotesize{}}
\includegraphics[width=0.15\linewidth]{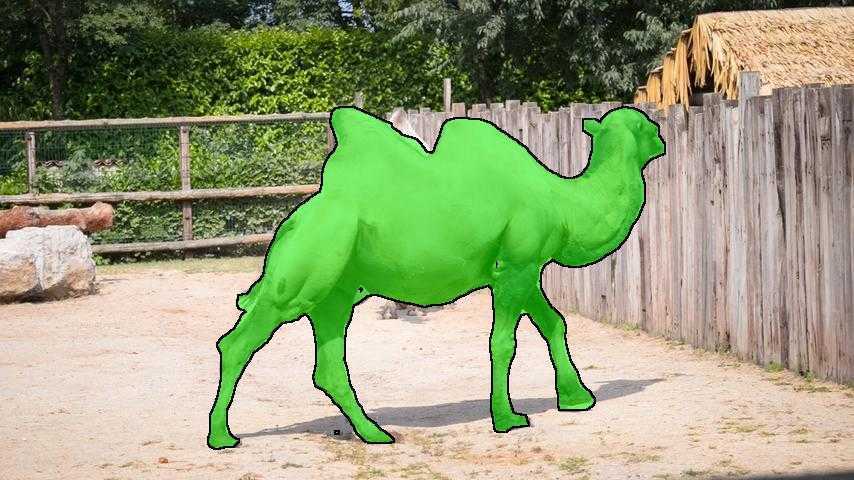} & {\footnotesize{}}
\includegraphics[width=0.15\linewidth]{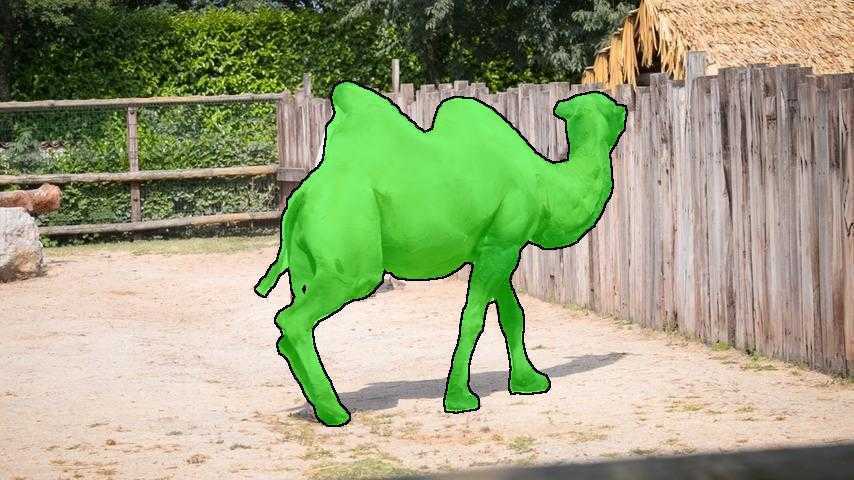} & {\footnotesize{}}
\includegraphics[width=0.15\linewidth]{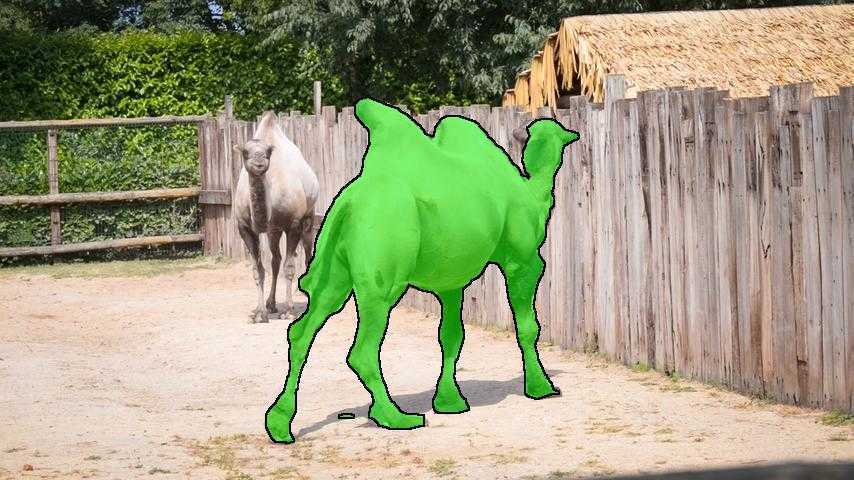} & {\footnotesize{}}
\includegraphics[width=0.15\linewidth]{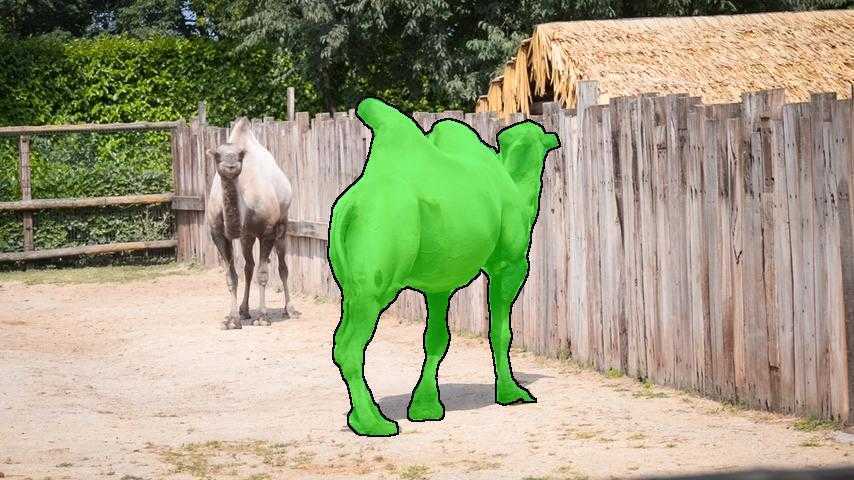} & {\footnotesize{}}
\includegraphics[width=0.15\linewidth]{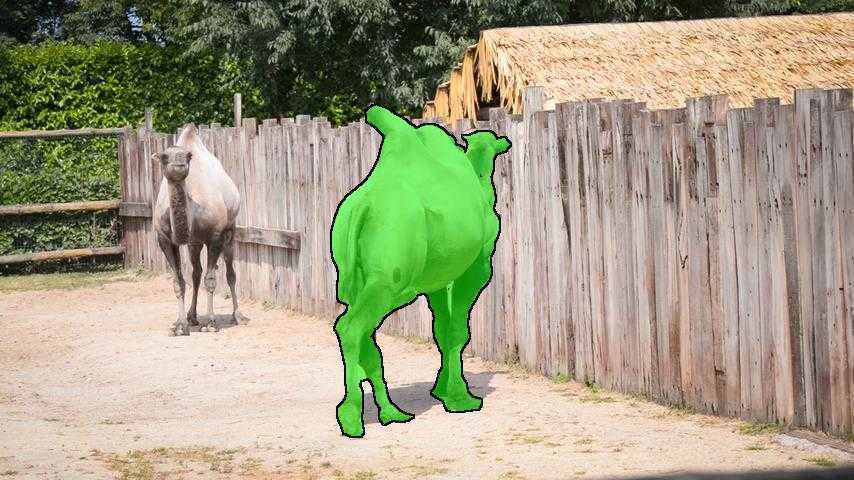}\tabularnewline

\multicolumn{6}{c}{\textit{ \textcolor{mygreen}{ID 1}: "A black scooter ridden by a man".  \textcolor{myred}{  ID 2}: "A man in a suit riding a scooter". }} \tabularnewline
\includegraphics[width=0.15\linewidth]{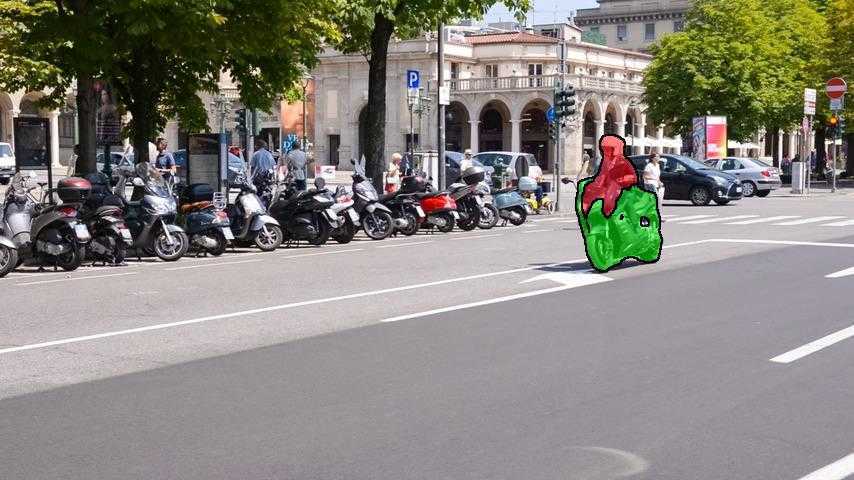} & {\footnotesize{}}
\includegraphics[width=0.15\linewidth]{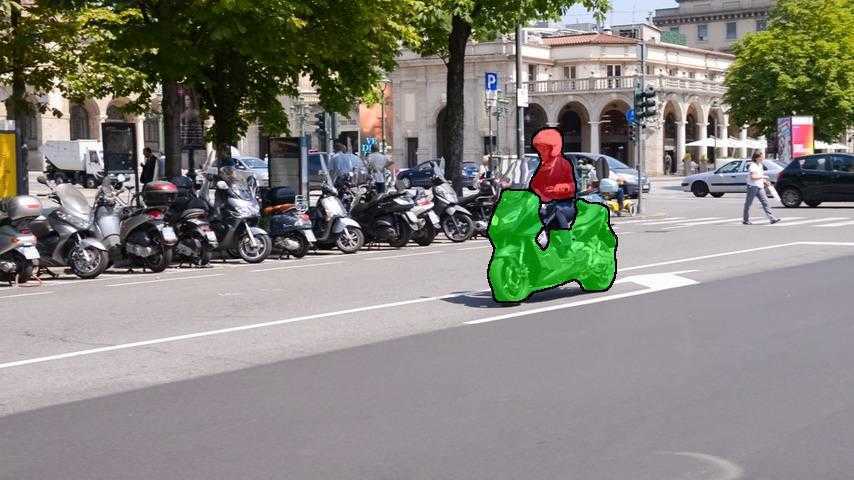} & {\footnotesize{}}
\includegraphics[width=0.15\linewidth]{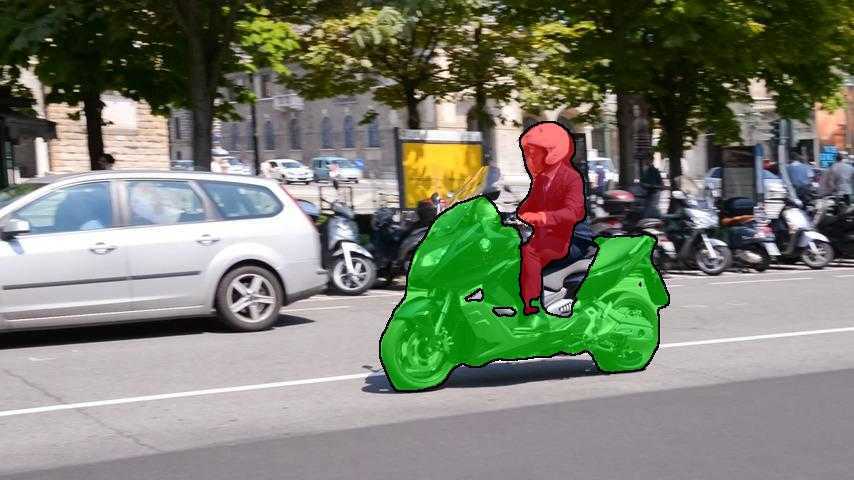} & {\footnotesize{}}
\includegraphics[width=0.15\linewidth]{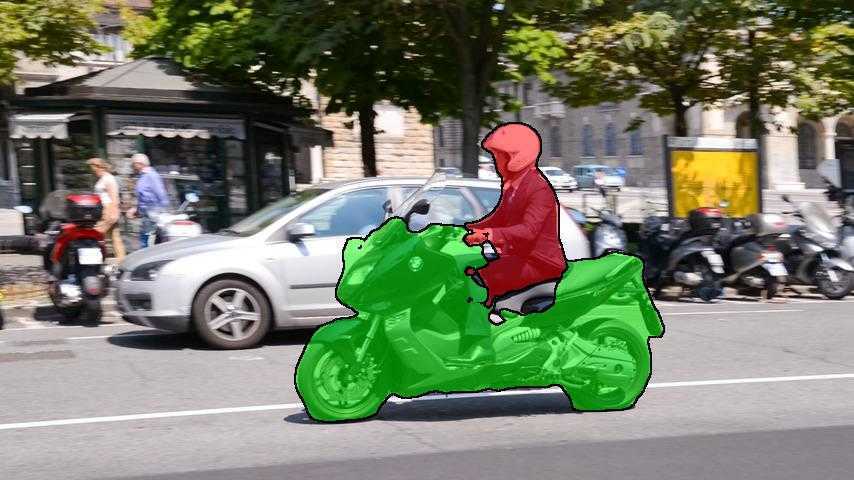} & {\footnotesize{}}
\includegraphics[width=0.15\linewidth]{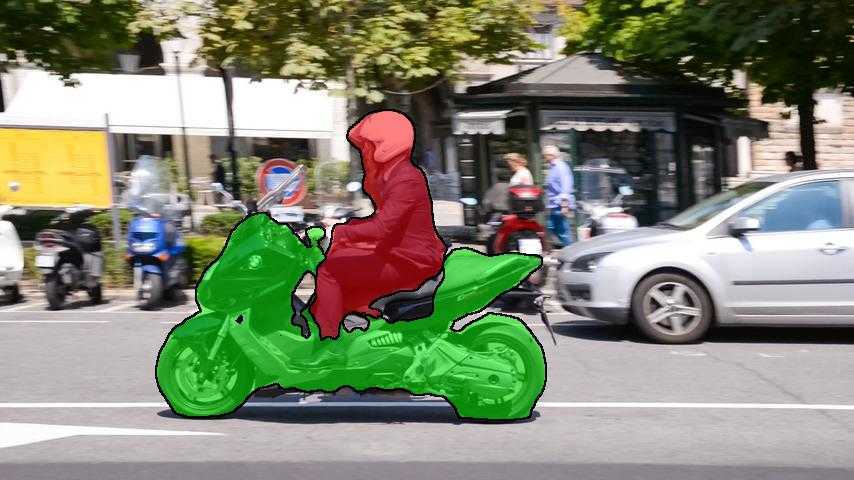} & {\footnotesize{}}
\includegraphics[width=0.15\linewidth]{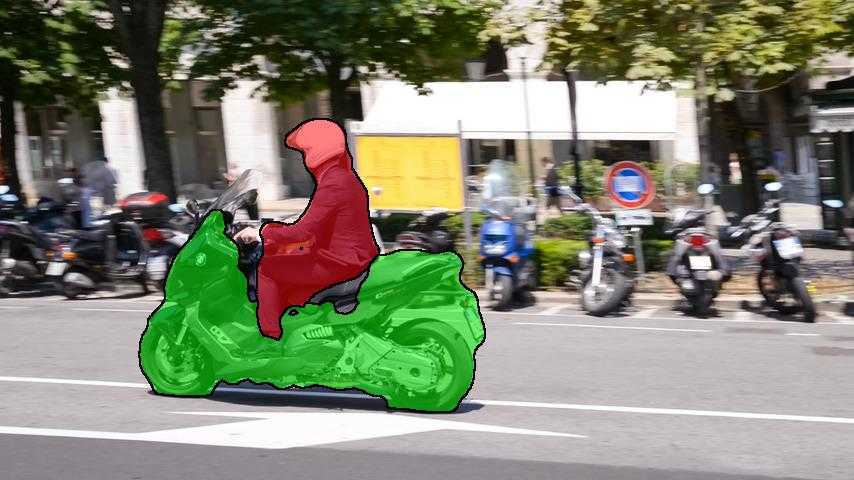}\tabularnewline

\end{tabular}\hfill{}
\par\end{centering}

\caption{\label{fig:qualitative-results-vos}Video object segmentation qualitative results using only referring expressions as supervision
on $\text{DAVIS}_{\text{16}}$ and $\text{DAVIS}_{\text{17}}$, val sets. Frames sampled along
the video.}

\end{figure*}

\section{\label{subsec:Grounding-results}Evaluation of natural language grounding in video}

In this section we discuss the performance of natural language grounding models on video data.
We experiment with DBNet \cite{2017-cvpr-dbnet} and MattNet \cite{yu2016modeling}. 
DBNet is trained on Visual Genome \cite{krishnavisualgenome} which contains images from MS~COCO \cite{lin2014eccvcoco} and YFCC100M \cite{thomee2016yfcc100m}, 
and spans thousands of object categories. 
MattNet is trained on referring expressions for MS~COCO images \cite{lin2014eccvcoco}, specifically RefCOCO and RefCOCO+ \cite{yu2016modeling}. 
Unlike RefCOCO which has no restrictions on the expressions, RefCOCO+ contains no spatial words and rather focuses on object appearance. 
Both aforementioned models rely on external bounding box proposals, such as EdgeBox \cite{Dollar2015Pami} or Mask R-CNN \cite{He_2017_ICCV}.

We carry out most of our evaluation on $\text{DAVIS}_{\text{16}}$ and $\text{DAVIS}_{\text{17}}$ with the referring expressions introduced in \S\ref{sec:Datasets}. 
To evaluate the localization quality we employ the intersection-over-union overlap (IoU) of the top scored box proposal with the ground truth bounding box, averaged across all queries. 

\subsection{$\text{DAVIS}_{\text{16}}$/$\text{DAVIS}_{\text{17}}$ referring expression grounding}

Table \ref{tab:comparative-result-grounding-davis16-17} reports performance of the grounding models on $\text{DAVIS}_{\text{16}}$ and $\text{DAVIS}_{\text{17}}$ referring expressions. 
In the following we summarize our key observations.

(1) We see the effect of replacing EdgeBox with Mask R-CNN object proposals for DBNet model ($54.1$ to $64.9$). Employing better proposals significantly improves the quality of this grounding method, thus we rely 
on Mask R-CNN proposals in all the following experiments. 
(2) We note the stability of grounding performance across two annotations (see $\Delta$(A1,A2)), showing that the grounding methods are quite robust to variations in language expressions. % ($\Delta$(A1,A2) varies from $0.2$ to $3.2$). 
(3) The grounding models trained on images are not stable across frames, even when small changes in appearance occur (e.g. see Figure \ref{fig:qualitative-results-grounding}). 
We see that our proposed temporal consistency technique benefits both methods (e.g. DBNet: $64.9$ vs. $68.8$ on $\text{DAVIS}_{\text{16}}$, MattNet $51.6$ vs. $52.8$ on $\text{DAVIS}_{\text{17}}$). 
(4) On both datasets MattNet performs better than DBNet. The gap is particularly large on $\text{DAVIS}_{\text{16}}$ ($72.5$ vs. $68.8$), as $\text{DAVIS}_{\text{16}}$ contains videos of a single foreground moving object, 
while DBNet is trained on a densely labeled Visual Genome dataset with many foreground and background objects.
(5) On $\text{DAVIS}_{\text{16}}$ MattNet trained on RefCOCO+ outperforms MattNet trained on RefCOCO ($72.5$ vs. $71.4$), 
while both perform similar on $\text{DAVIS}_{\text{17}}$. As RefCOCO+ contains no spatial words, MattNet trained on this dataset is more accurate in localizing queries mentioning object appearance. 
(6) Compared to $\text{DAVIS}_{\text{16}}$, $\text{DAVIS}_{\text{17}}$ is significantly more challenging, as it contains cluttered scenes with multiple moving objects (e.g. for MattNet $71.4$ vs. $52.8$). 
(7) When comparing results on expressions provided for the first frame versus expressions provided for the full video, we observe diverging trends. 
While DBNet is able to improve its performance ($48.4$ vs. $49.6$), MattNet performance decreases ($52.8$ vs. $51.3$). We attribute this to the fact that DBNet is trained on the more diverse Visual Genome descriptions.

\begin{table*}[t]

\setlength{\tabcolsep}{0.25em} 
\renewcommand{\arraystretch}{0.95}
\begin{centering}
\begin{tabular}{c|c|c|c|cc||cc|c}
\multirow{3}{*}{{\footnotesize{}{Method}}} & \multirow{3}{*}{\begin{tabular}{c}{\footnotesize{}{Object}}\tabularnewline {\footnotesize{}{proposals}}\tabularnewline \end{tabular}} 
& \multirow{3}{*}{\begin{tabular}{c}{\footnotesize{}{Train.}}\tabularnewline {\footnotesize{}{data}}\tabularnewline \end{tabular}} 
& \multirow{3}{*}{\begin{tabular}{c}{\footnotesize{}{Temp.}}\tabularnewline {\footnotesize{}{cons.}}\tabularnewline \end{tabular}} & \multicolumn{2}{c||}{ \footnotesize{}{$\text{DAVIS}_{\text{16}}$}}
& \multicolumn{3}{c}{ \footnotesize{}{$\text{DAVIS}_{\text{17}}$}} \tabularnewline
& & & & \multicolumn{2}{c||} {{\footnotesize{}1st frame}} &   \multicolumn{2}{c|} {{\footnotesize{}1st frame}} 
& {\footnotesize{}{Full video}} \tabularnewline
& & & & {\footnotesize{}mIoU}  & {\footnotesize{}$\Delta$(A1,A2)} &{\footnotesize{}mIoU}  &{\footnotesize{}$\Delta$(A1,A2)}  & {\footnotesize{}mIoU} \tabularnewline
\hline 
\hline 
\multirow{2}{*}{{\footnotesize{}{DBNet}}} & {EdgeBox}& \multirow{2}{*}{{\footnotesize{}{\small{}Vis.Gen.}}} & \textcolor{black}{\scriptsize{}-} & {\small{}54.1}& {\small{}1.0} & \textcolor{black}{\scriptsize{}-}  & \textcolor{black}{\scriptsize{}-} & \textcolor{black}{\scriptsize{}-} \tabularnewline
 & {Mask R-CNN} &  & \textcolor{black}{\scriptsize{}-} & {\small{}64.9}& {\small{}2.1} & {\small{}48.4} & {\small{}1.3} & {\small{}49.6}\tabularnewline
 \arrayrulecolor{lightgray} \hline \arrayrulecolor{black}
\multirow{2}{*}{{\footnotesize{}{MattNet}}} & \multirow{2}{*}{{\footnotesize{}{Mask R-CNN}}} 

 & {\footnotesize{}RefCOCO} & \textcolor{black}{\scriptsize{}-} & {\small{}67.1}& {\small{}2.2} & {\small{}51.6} & {\small{}1.6}  & {\small{}50.3}\tabularnewline
 &  & {\footnotesize{}RefCOCO+} & \textcolor{black}{\scriptsize{}-}& {\small{}69.1}& {\small{}3.2} & {\small{}50.8} & {\small{}1.2} & {\small{}50.1}\tabularnewline
\hline
\hline
{\small{}DBNet} & {Mask R-CNN} & {\small{}Vis.Gen.} & \textcolor{black}{\scriptsize{}\Checkmark{}} & {\small{}68.8}& {\small{}0.6} & {\small{}49.6} & {\small{}1.6} & {\small{}50.2}\tabularnewline
 \arrayrulecolor{lightgray} \hline \arrayrulecolor{black}
\multirow{2}{*}{{\footnotesize{}{MattNet}}}  & \multirow{2}{*}{{\footnotesize{}{Mask R-CNN}}} 
  
 & {\footnotesize{}RefCOCO} & \textcolor{black}{\scriptsize{}\Checkmark{}}& {\small{}71.4}& {\small{}0.2} & {\small{}52.8} & {\small{}0.5} & {\small{}51.3}\tabularnewline
 &   & {\footnotesize{}RefCOCO+} & \textcolor{black}{\scriptsize{}\Checkmark{}} & {\small{}72.5}& {\small{}0.3}  & {\small{}52.3}  & {\small{}0.0} & {\small{}51.2}\tabularnewline

\end{tabular}

\par
\end{centering}

\caption{\label{tab:comparative-result-grounding-davis16-17} Comparison of the DBNet\cite{2017-cvpr-dbnet} and MattNet \cite{yu2018mattnet} models on $\text{DAVIS}_{\text{16}}$ training set 
and $\text{DAVIS}_{\text{17}}$ val set. $\Delta$(A1,A2) denotes the difference between Annotator 1 and 2.  }

\end{table*}

\subsubsection{Attribute-based analysis.}
Next we perform a more detailed analysis of the grounding models on $\text{DAVIS}_{\text{17}}$. We split the textual queries/videos into subsets where a certain attribute is present and report the averaged results for the subsets. 
Table \ref{tab:comparative-result-grounding-davis17-attr1} presents attribute-based grounding performance on first-frame based expressions averaged across annotators. 
To estimate the upper bound performance and the impact of imperfect bounding box proposals we add an Oracle comparison, where performance is reported on the ground-truth object boxes. We summarize our findings in the following.

(1) As MattNet is trained on MS~COCO images and both models rely on MS~COCO-based Mask R-CNN proposals, we compare performance for expressions which include COCO versus non-COCO objects. 
Both models drop in performance on non-COCO expressions, showing the impact of the domain shift to $\text{DAVIS}_{\text{17}}$ (e.g. for MattNet $59.6$ vs. $36.9$). 
Even DBNet which is trained on a larger training corpus suffers from the same effect ($55.5$ vs. $37.3$). 
(2) We label the $\text{DAVIS}_{\text{17}}$ expressions 
as ``spatial'' if they include some of the spatial words (e.g. left, right). Such queries are significantly harder for all models (e.g. for MattNet $33.8$ vs. $58.5$). 
(3) Verbs are important as they allow to disambiguate an object in a video based on its actions. Presence of verbs in expressions is a challenging factor for DBNet trained on Visual Genome, 
while MattNet does significantly better ($37.4$ vs. $55.8$). 
(4) Expression length is also an important factor. We quantize our expressions into Short (<4 words), 
Medium (4--6 words) and Long (>6 words). All models demonstrate similar drop in performance as expression length increases (e.g. for MattNet $63.9\rightarrow 50.2 \rightarrow 49.1$). 
(5) Videos with more objects are more difficult, as these objects also tend to be very similar, such as e.g. fish in a tank (e.g. for MattNet $86.1\rightarrow 51.2 \rightarrow 16.1$). 
(6) From the Oracle performance on COCO versus non-COCO expressions, we see that all models are able to significantly improve their performance even for non-COCO objects (e.g. for DBNet $37.3$ to $59.0$). 
DBNet benefits more than MattNet from Oracle boxes, showing its higher potential to generalize to a new domain given better proposals. 

\begin{table*}[t]
	\setlength{\tabcolsep}{0.1em} 
	\renewcommand{\arraystretch}{1}
	\begin{centering}

	\begin{tabular}{@{}l@{\ }|@{\ }c@{\ }|@{}c@{}|@{\ }c@{\ }c@{\ }|@{\ }c@{\ \ }c@{\ }|@{\ }c@{\ \ }c@{\ }|@{\ }c@{\ \ }c@{\ \ }c@{\ }|@{\ }c@{\ }c@{\ }c@{}}
			\multirow{3}{*}{{\footnotesize{}{Method}}} & \multirow{3}{*}{\begin{tabular}{c}{\footnotesize{}{Train.}}\tabularnewline {\footnotesize{}{data}}\tabularnewline \end{tabular}} & \multirow{3}{*}{\begin{tabular}{c}{\footnotesize{}{Obj.}}\tabularnewline {\footnotesize{}{prop.}} \end{tabular} }
			& \multicolumn{12}{c} {{\footnotesize{} mIoU}} \tabularnewline
			& & & \multirow{2}{*}{ {\footnotesize{}CO.}} &  \multirow{2}{*}{{\footnotesize{}\~{}CO.}} & \multirow{2}{*}{ {\footnotesize{}Sp.}} &  \multirow{2}{*}{{\footnotesize{}\~{}Sp.}} & \multirow{2}{*}{ {\footnotesize{}Ve.}} &  \multirow{2}{*}{{\footnotesize{}\~{}Ve.}} & \multicolumn{3}{c}{ {\footnotesize{} Expr. length}} & \multicolumn{3}{c}{ {\footnotesize{} Num. obj.}} \tabularnewline
			& & &   && & & & & {\footnotesize{}S} & {\footnotesize{}M} & {\footnotesize{}L} &{\footnotesize{}1} & {\footnotesize{}2-3} & {\footnotesize{}>3}\tabularnewline
			\hline 
			\hline 
			\footnotesize{}{DBNet} &  \footnotesize{}{Vis.Gen.} & \multirow{2}{*}{\footnotesize{} {\begin{tabular}{c}{\footnotesize{}{Mask}}\tabularnewline {\footnotesize{}{R-CNN}}\tabularnewline \end{tabular}}} 
			& \footnotesize{}{55.5} & \footnotesize{}{37.3}  &  \footnotesize{}{\textbf{36.5}} & \footnotesize{}{55.7}  &   \footnotesize{}{37.4} & \footnotesize{}\textbf{52.0}  & \footnotesize{}{61.8} & \footnotesize{}{49.2} & \footnotesize{}{33.6} & \footnotesize{}{79.5} & \footnotesize{}{49.3} & \footnotesize{}\textbf{22.6} \tabularnewline
			{{\footnotesize{}{MattNet}}} &  \footnotesize{}{RefCOCO} &  
			& \textbf{\footnotesize{}{59.6}} &\footnotesize{}{36.9}   & \footnotesize{}{33.8} & \footnotesize{}\textbf{58.5}  & \footnotesize{}\textbf{55.8} & \footnotesize{}{51.7}  & \footnotesize{}\textbf{63.9} & \footnotesize{}{\textbf{50.2}} & \footnotesize{}\textbf{49.1} & \footnotesize{}\textbf{86.1} & \footnotesize{}\textbf{51.2} & \footnotesize{}{16.1} \tabularnewline
			\arrayrulecolor{lightgray} \hline \arrayrulecolor{black}
			\footnotesize{DBNet} &  \footnotesize{}{Vis.Gen.} & \multirow{2}{*}{\footnotesize{}{Oracle}} & \textbf{\footnotesize{}{79.3}} & \footnotesize{}\textbf{59.0}   & \footnotesize{}{\textbf{47.7}} & \footnotesize{}\textbf{81.7}  & {\footnotesize{}{70.3}} & \footnotesize{}\textbf{77.6}   & \footnotesize{}\textbf{84.8} & \footnotesize{}\textbf{69.9} & \footnotesize{}\textbf{67.9} & \footnotesize{}{100} & \footnotesize{}\textbf{73.8} & \footnotesize{}{\textbf{37.2}} \tabularnewline
			{{\footnotesize{}{MattNet}}} &  \footnotesize{}{RefCOCO} &  & \footnotesize{}{73.2} & \footnotesize{}{46.6}   & \footnotesize{}{42.2} & \footnotesize{}{72.5}  & \footnotesize{}\textbf{74.7} & \footnotesize{}{62.9}   & \footnotesize{}{79.0} & \footnotesize{}{61.1}   & \footnotesize{}{59.0}  & \footnotesize{}{100} & \footnotesize{}{64.5} & \footnotesize{}{23.2}\tabularnewline
		\end{tabular}
		\par\end{centering}

	\caption{\label{tab:comparative-result-grounding-davis17-attr1}Grounding performance breakdown for different attributes
		on $\text{DAVIS}_{\text{17}}$, val set. 
		Results obtained after the temporal consistency, using average between two annotators (1st frame based). 
		Attributes: COCO/non-COCO, Spatial/non-Spatial, Verbs/no Verbs, Expression length (Short, Medium, Long) and Number of objects.}

\end{table*}

\section{\label{subsec:VOS-results}Video object segmentation results}

In this section we present single and multiple video object segmentation results using natural language referring expressions on two datasets: 
$\text{DAVIS}_{\text{16}}$ \cite{Perazzi2016Cvpr} and $\text{DAVIS}_{\text{17}}$ \cite{Pont-Tuset_arXiv_2017}. In addition, we experiment
with fusing two complementary sources of information, employing both the pixel-level mask and language supervision on the first frame. All results here are obtained using the bounding boxes 
given by the MattNet model \cite{yu2018mattnet} trained on RefCOCO \cite{yu2016modeling} after the temporal consistency step (see \S\ref{subsec:Grounding}).

%\paragraph{Evaluation criteria.}

For evaluation we use the IoU measure
(also called Jaccard index - $J$) between the ground truth and
the predicted segmentation, averaged across all video sequences and all frames.  
For $\text{DAVIS}_{\text{17}}$ we also employ the $J\&F$ measure proposed in \cite{Pont-Tuset_arXiv_2017}.
% Please refer to \cite{Pont-Tuset_arXiv_2017} for more details.

\subsection{\label{davis16-vos}$\text{DAVIS}_{\text{16}}$ single object segmentation}

Table \ref{tab:comparative-result-VOS-davis16} compares our results to previous work on $\text{DAVIS}_{\text{16}}$ \cite{Perazzi2016Cvpr}. 
As we employ MattNet \cite{yu2018mattnet}, which exploits Mask R-CNN \cite{He_2017_ICCV} box proposals, we also would like to compare to its segments.
We report the oracle Mask R-CNN results, where on each frame the segment with the highest ground truth overlap was chosen. Even with the oracle assignment of segments, \cite{He_2017_ICCV} under-performs compared to our segmentation model ($71.5$ vs. $83.1$).
This shows that for very detailed mask annotations (as in $\text{DAVIS}_{\text{16/17}}$) a more complex segmentation module than the Mask R-CNN segmentation head is required (which itself is a shallow FCN with reduced output resolution, resulting in coarse masks).

Our method, while only exploiting language, shows competitive performance, on par with techniques which use 
a pixel-level mask on the first frame ($82.8$ vs. $81.7$ for OnAVOS \cite{Voigtlaender2017OnlineAO}).
This shows that high quality results can be obtained via a more natural way of human-computer interaction -- referring to an object via language, making video segmentation techniques more applicable in practice.
Compared to mask supervision employing language results in a runtime speed up: it is $\sim\!\!15$ times faster to specify the object with language ($79$s \cite{lin2014eccvcoco} vs. $5$s) plus online tuning is not needed for good performance (\cite{Caelles2017SemanticallyGuidedVO} reports $10$min for online tuning with $80.2$ vs. our $82.8$). 
Note that \cite{Caelles2017SemanticallyGuidedVO,Voigtlaender2017OnlineAO} show superior results to our approach ($\sim86$ mIoU). However, they employ additional cues
by incorporating semantic information \cite{Caelles2017SemanticallyGuidedVO} or doing online adaptation \cite{Voigtlaender2017OnlineAO}. %, which brings a further increase in performance ($\sim 4$ points).
Potentially, these techniques can also be applied to our method, though it is out of scope of this paper.

\begin{wraptable}{r}{0.5\linewidth}
	\setlength{\tabcolsep}{0.1em} 
	\renewcommand{\arraystretch}{0.98}
	\begin{centering}
		
		\begin{tabular}{c cc|c}
			\multicolumn{2}{c}{\multirow{1}{*}{{\footnotesize{}{Supervision}}}} & \multirow{1}{*}{{\footnotesize{}Method}} & {\footnotesize{}mIoU}\tabularnewline
			\hline 
			\hline 
			\multicolumn{2}{c}{
				\multirow{1}{*}{ 
					{\footnotesize{Oracle}}%
					\begin{tabular}{c}
						{\footnotesize{}}\tabularnewline
					\end{tabular}{\footnotesize{} }
			}} 
			& {\small{}Mask R-CNN \cite{He_2017_ICCV}}  & {\small{}71.5} \tabularnewline
			\hline 
			
			\multicolumn{2}{c}{
				\multirow{3}{*}{ 
					{\footnotesize{}}%
					\begin{tabular}{c}
						{\footnotesize{}Unsupervised}\tabularnewline
					\end{tabular}{\footnotesize{} }
			}} 
			& {\small{}FusionSeg \cite{Jain2017ArxivFusionSeg}} & {\small{}70.7}\tabularnewline
			\multicolumn{2}{c}{} & {\small{}LVO \cite{TokmakovAS17}}  & {\small{}75.9} \tabularnewline
			\multicolumn{2}{c}{} & {\small{}ARP \cite{Koh_CVPR_2017}} & {\small{}76.2}\tabularnewline
			\hline 
			
			\multirow{11}{*}{\begin{turn}{90}
					{\footnotesize{}}%
					\begin{tabular}{c}
						{\footnotesize{}Semi-supervised}\tabularnewline
					\end{tabular}{\footnotesize{} }
				\end{turn}
			} 
			&
			\multirow{6}{*}{
				{\footnotesize{}}%
				\begin{tabular}{c}
					{\footnotesize{}1st frame}\tabularnewline
					{\footnotesize{}mask}\tabularnewline
				\end{tabular}{\footnotesize{} }
			} 
			& {\small{}SegFlow \cite{Cheng_ICCV_2017}} & {\small{}76.1} \tabularnewline
			& & {\small{}MaskTrack \cite{Khoreva2017CvprMaskTrack}} & {\small{}79.7} \tabularnewline
			& & {\small{}OSVOS$^1$ \cite{Caelles2017SemanticallyGuidedVO} } & {\textcolor{black}{\small{}80.2}}\tabularnewline
			& & {\small{}MaskRNN \cite{Hu2017MaskRNNIL}} & {\textcolor{black}{\small{}80.4}}\tabularnewline
			& & {\small{}OnAVOS$^2$ \cite{Voigtlaender2017OnlineAO}} & {\textcolor{black}{\small{}81.7}}\tabularnewline
			& & {\small{}Our } & {\small{}83.1} \tabularnewline
			\cline{2-4}
			
			&
			\multirow{2}{*}{
				{\footnotesize{}}%
				\begin{tabular}{c}
					{\footnotesize{}Clicks}\tabularnewline
				\end{tabular}{\footnotesize{} }
			} 
			
			& {\small{}iVOS\cite{Benard_arxiv17}}  & {\small{}80.6} \tabularnewline
			& & {\small{}DEXTR \cite{Maninis_arxiv17}}  & {\small{}80.9} \tabularnewline
			
			\cline{2-4}
			
			& \multirow{1}{*}{
				{\footnotesize{}}%
				\begin{tabular}{c}
					{\footnotesize{}Language}\tabularnewline
				\end{tabular}{\footnotesize{} }
			} 
			& {\small{}\arrayrulecolor{black}}$\text{Our}$  & {\small{}82.8} \tabularnewline
			\cline{2-4}
			
			& \multirow{1}{*}{
				{\footnotesize{}}%
				\begin{tabular}{c}
					{\footnotesize{}Mask + Lang.}\tabularnewline
				\end{tabular}{\footnotesize{} }
			} 
			& {\small{}\arrayrulecolor{black}}$\text{Our}$  & \textbf{\small{}84.5} \tabularnewline
			
		\end{tabular}
		\par\end{centering}
	\caption{\label{tab:comparative-result-VOS-davis16}Comparison of video object segmentation results
		on $\text{DAVIS}_{\text{16}}$, val set.}
\end{wraptable}

\footnotetext[1]{OSVOS$^{S}$ reports 86.0 mIoU by employing semantic segmentation as additional supervision.}
\footnotetext[2]{OnAVOS gives 86.1 mIoU by exploiting online adaptation on successive frames.}

Compared to the approaches which use point click supervision \cite{Benard_arxiv17,Maninis_arxiv17}, our method shows superior performance ($82.8$ vs. $80.6$ and $80.9$). 
This indicates that language can be successfully utilized as an alternative and cheaper form of supervision for video object segmentation, on par with clicks and scribbles.

\subsubsection{Maks and language.}

\begin{table}[t]
	\setlength{\tabcolsep}{0.5em} 
	\renewcommand{\arraystretch}{1}
	\begin{centering}
		\begin{tabular}{c|ccccccccccc}
			\multirow{1}{*}{{\footnotesize{}{Supervision}}} & {\footnotesize{}AC} & {\footnotesize{}LR} & {\footnotesize{}SV} & {\footnotesize{}SC} & {\footnotesize{}CS} & {\footnotesize{}DB} & {\footnotesize{}BC} & {\footnotesize{}FM} & {\footnotesize{}MB} & {\footnotesize{}DEF} & {\footnotesize{}OCC} \tabularnewline
			\hline 
			\hline 
			
			{\small{}Language}  & {\small{}80.1} & \textbf{\small{}79.0} & {\small{}74.4} & {\small{}77.6} & {\small{}85.7} & {\small{}66.4} & \textbf{\small{}85.0} & {\small{}77.7} & {\small{}78.1} & {\small{}84.3} & {\small{}80.1}  \tabularnewline
			{\small{}Mask }  & \textbf{\small{}81.2} & {\small{}78.1} & {\small{}75.9} & {\small{}79.0} & {\small{}85.6} & {\small{}68.0} & {\small{}82.8} & {\small{}79.0} & {\small{}79.9} & {\small{}85.6} & {\small{}80.5} \tabularnewline
			\arrayrulecolor{lightgray} \hline \arrayrulecolor{black}
			{\small{}Mask + Lang.} & {\small{}81.0} &\textbf{\small{}79.0} & \textbf{\small{}76.8} & \textbf{\small{}80.4} & \textbf{\small{}86.8} & \textbf{\small{}72.2} & {\small{}84.4} & \textbf{\small{}79.5} & \textbf{\small{}80.4} & \textbf{\small{}85.9} & \textbf{\small{}82.3} \tabularnewline
		\end{tabular}
		\par\end{centering}
	\caption{\label{tab:comparative-result-VOS-mask-language} Attribute-based results with different forms of supervision on $\text{DAVIS}_{\text{16}}$, val set.
		AC: appearance change, LR: low resolution, SV: scale variation, SC: shape complexity, CS: camera shake, DB: dynamic background, BC: background clutter,
		FM: fast motion, MB: motion blur, DEF: deformation, OCC: occlusions. See \S\ref{davis16-vos} for more details.}
\end{table}

In Table \ref{tab:comparative-result-VOS-davis16} we also report the results for variants using only mask supervision on the the first frame or 
combining both mask and language (see \S\ref{subsec:Segmentation} for details).
Notice that employing either mask or language results in comparable performance ($82.8$ vs. $83.1$), while fusing both modalities leads to a further improvement ($82.8$ vs. $84.5$). This shows that referring expressions are complementary to visual forms of supervision and can be exploited as an additional source of guidance for segmentation, on top of not only pixel-level masks, but potentially scribbles and point clicks.

Table \ref{tab:comparative-result-VOS-mask-language} presents a more detailed evaluation using video attributes. We report the averaged results on a subset of sequences where
a certain challenging attribute is present. Note that using language alone leads to more robust performance for videos with low resolution, camera shake and background clutter 
without the need for an expensive pixel-level mask. When utilizing both mask and language we observe that the system becomes consistently more robust to various video challenges (e.g. fast motion, occlusions, motion blur, etc.) and compares favorably to mask only on all attributes, 
except appearance change. Overall, employing language can help the model to better handle occlusions, avoid drift and better adapt to complex dynamics inherent to video.

\subsubsection{Ablation study.}

\begin{wraptable}{r}{0.45\linewidth}
	\setlength{\tabcolsep}{0em}
	\renewcommand{\arraystretch}{1}
	\begin{centering}
		\begin{tabular}{lcc}
			\footnotesize{} Variant  & \footnotesize{} mIoU  & \footnotesize{} $\Delta$ \tabularnewline
			\hline
			\hline
			\footnotesize{} Full system  & \small{} \textbf{$82.5$} & \small - \tabularnewline
			\hline
			\footnotesize{} No box jittering & \small{} $80.6$ & \small{} $-1.9$\tabularnewline
			\arrayrulecolor{lightgray} \hline \arrayrulecolor{black}
			\footnotesize{} No optical flow magnitude  & \small{} $75.9$ & \small{} $-4.7$\tabularnewline
			\arrayrulecolor{lightgray} \hline \arrayrulecolor{black}
			\footnotesize{} No temporal consistency  & \small{} $72.5$ & \small{} $-3.4$\tabularnewline
			
			\footnotesize{} Backbone architecture of \cite{Khoreva2017CvprMaskTrack} & \small{} $72.2$ & \small{} $-3.7$\tabularnewline
		\end{tabular}
		\par\end{centering}
	\caption{\label{tab:ablation-study}Ablation study on $\text{DAVIS}_{\text{16}}$.}
\end{wraptable}

We validate the contributions of the components in our method (see \S\ref{sec:Method}) by presenting an ablation study in Table \ref{tab:ablation-study} on $\text{DAVIS}_{\text{16}}$, training set. 
Augmenting the ground truth boxes by random jittering makes the system more robust to sloppy boxes at test time ($82.5$ vs. $80.6$), while 
employing motion cues allows to better handle moving objects ($80.6$ vs. $75.9$).
Temporal consistency step helps to provide more temporally coherent boxes (4.3 mIoU point boost for grounding, see Table \ref{tab:comparative-result-grounding-davis16-17}) and hence improve the final segmentation quality ($75.9$ vs. $72.5$).
Exploiting the proposed network architecture versus using the network proposed in \cite{Khoreva2017CvprMaskTrack} results in $3.7$ point boost ($75.9$ vs. $72.2$), providing more detailed object masks.
Overall, all components introduced in our approach lead to the state-of-the-art results on $\text{DAVIS}_{\text{16}}$.

\subsection{$\text{DAVIS}_{\text{17}}$ multiple object segmentation}

Table \ref{tab:comparative-result-VOS-davis17} presents results on $\text{DAVIS}_{\text{17}}$ \cite{Pont-Tuset_arXiv_2017}.
The lower numbers in comparison with Table \ref{tab:comparative-result-VOS-davis16} indicate that $\text{DAVIS}_{\text{17}}$ is significantly 
more difficult than $\text{DAVIS}_{\text{16}}$. Even when employing mask supervision on the first frame the dataset presents a challenging task and there is much room for improvement.
The semi-supervised methods perform well on foreground-background segmentation, 
but have problems separating multiple foreground objects, handling small objects and preserving the correct object identities \cite{Pont-Tuset_arXiv_2017}.

Compared to mask supervision using language descriptions significantly under-performs. We believe that one of the main problems is a relatively unstable behavior of the underlying grounding model.
There are a lot of identity switches, that are heavily penalized by the evaluation metric as every pixel should be assigned to one instance. We conducted an oracle experiment assigning
Mask R-CNN box proposals to the correct object ids and then performing segmentation (denoted ``Oracle - Grounding''). 
We observe a significant increase in performance ($37.3$ to $54.9$), making the results competitive to mask supervision. If we utilize Mask R-CNN segment proposals for oracle case, the result 
is $2.1$ points lower than using our segmentation model on top.
The underlying choice of proposals for the grounding model could also have its effect. If the object is not detected by Mask R-CNN, the grounding model has no chances to recover the correct instance.
To evaluate the influence of proposals we conduct an oracle experiment where the ground truth boxes are exploited in the grounding model (denoted ``Oracle - Box proposals''). With oracle boxes we observe
an increase in performance ($37.3$ to $42.1$), however, recovering the correct identities still poses a problem for grounding.

Another factor influencing the results is the domain shift between the training and test data. Both Mask R-CNN and MattNet are trained on MS~COCO \cite{lin2014eccvcoco}, 
and have troubles recovering instances not belonging to $80$ COCO categories.
We split the $\text{DAVIS}_{\text{17}}$ validation set into COCO and non-COCO objects/language queries ($43$ vs. $18$) and evaluate separately on two subsets. 
As in \S\ref{subsec:Grounding-results}, we observe much higher results for COCO queries ($45$ to $27.5$), indicating the problem of generalization from training to test data.

The method which exploits scribble supervision \cite{Pont-Tuset_arXiv_2018} performs on par with our approach. Note that even for scribble supervision the task remains difficult.

\subsubsection{Mask and language.}

\begin{wraptable}{r}{0.5\linewidth}
	\setlength{\tabcolsep}{0em} 
	\renewcommand{\arraystretch}{0.95}
	\begin{centering}
		
		\begin{tabular}{cc|cc}
			\multirow{1}{*}{{\footnotesize{}{Supervision}}} & \multirow{1}{*}{{\footnotesize{}Method}} & {\footnotesize{} mIoU} & {\footnotesize{}$J\& F$} \tabularnewline
			\hline 
			\hline 
			\multicolumn{1}{c}{
				\multirow{3}{*}{ 
					{\footnotesize{Oracle}}%
					\begin{tabular}{c}
						{\footnotesize{}}\tabularnewline
					\end{tabular}{\footnotesize{} }
			}} 
			& {\small{}Mask R-CNN \cite{He_2017_ICCV}}  & {\small{}52.8}  & {\small{}53.3}  \tabularnewline
			\arrayrulecolor{lightgray} \cline{2-4} \arrayrulecolor{black}
			& {\small{}\arrayrulecolor{black}}$\text{Grounding}$  & {\small{54.9}} & {\small{57.4}}\tabularnewline
			& {\small{}\arrayrulecolor{black}}$\text{Box proposals}$  & {\small{42.1}} & {\small{45.3}}\tabularnewline
			\hline 
			
			\multirow{3}{*}{
				{\footnotesize{}}%
				\begin{tabular}{c}
					{\footnotesize{}1st frame}\tabularnewline
					{\footnotesize{}mask}\tabularnewline
				\end{tabular}{\footnotesize{} }
			} 
		    & {\small{}OSVOS \cite{Caelles2017Cvpr}} & {\textcolor{black}{\small{}52.1}} & \small{}57.0\tabularnewline
			& {\small{}OnAVOS$^3$ \cite{DAVIS2017-5th}} & {\textcolor{black}{\small{}57.0}} & {\textcolor{black}{\small{}59.4}}\tabularnewline
			& {\small{}MaskRNN \cite{Hu2017MaskRNNIL}} & {\textcolor{black}{\small{}60.5}} & -\tabularnewline
			& {\small{}Our } & {\small{}58.0}& {\small{}60.8} \tabularnewline
			\hline
			
			\multirow{2}{*}{
				{\footnotesize{}}%
				\begin{tabular}{c}
					{\footnotesize{}Scribbles}\tabularnewline
				\end{tabular}{\footnotesize{} }
			}
			& {\small{}CNN lin. class. \cite{Pont-Tuset_arXiv_2018}} & {\textcolor{black}{\small{}-}} & {\textcolor{black}{\small{}39.3}}\tabularnewline
			& {\small{}Scribble-OSVOS \cite{Pont-Tuset_arXiv_2018}} & {\textcolor{black}{\small{}-}} & {\textcolor{black}{\small{}39.9}}\tabularnewline
			
			\hline

			\multirow{3}{*}{
				{\footnotesize{}}%
				\begin{tabular}{c}
					{\footnotesize{}Language}\tabularnewline
				\end{tabular}{\footnotesize{} }
			} 
			& {\small{}\arrayrulecolor{black}}$\text{Our}$  & {\small{}37.3} & {\small{}39.3}\tabularnewline
			\arrayrulecolor{lightgray} \cline{2-4} \arrayrulecolor{black}
			& {\small{}\arrayrulecolor{black}}$\text{Our, COCO}$  & \textit{\small{}45.0} & \textit{\small{}47.5} \tabularnewline
			& {\small{}\arrayrulecolor{black}}$\text{Our, non-C.}$  & \textit{\small{}27.5} & \textit{\small{}29.4} \tabularnewline
			
			\hline

			\multirow{1}{*}{
				\footnotesize{} Mask+Lang.
			} 
			& \multirow{1}{*}{{\small{}\arrayrulecolor{black}}$\text{Our}$ } & \multirow{1}{*}{{\small{} 59.0}} & \multirow{1}{*} {\small{} 62.2} \tabularnewline
		\end{tabular}
		\par\end{centering}

	\caption{\label{tab:comparative-result-VOS-davis17}Comparison of semi-supervised video object segmentation methods
		on $\text{DAVIS}_{\text{17}}$, val set. Numbers
		in italic are reported on subsets of $\text{DAVIS}_{\text{17}}$ containing/non-containing COCO objects.}

\end{wraptable}
\footnotetext[3]{OnAVOS reports 64.5 mIoU by performing online adaptation on successive frames.}

In Table \ref{tab:comparative-result-VOS-davis17} we also report the results for variants of our approach using only mask supervision or 
combining mask and language. Employing language on top of mask leads to an increase in performance over using mask only ($58$ to $59$), again showing complementarity of both sources of supervision.

Figure \ref{fig:qualitative-results-vos} provides qualitative results of our method using only language as supervision.
We observe successful handling of similar looking objects, fast motion, deformations and partial occlusions.

\subsubsection{Discussion.}

Our results indicate that language alone can be successfully used as an alternative and a more natural form of supervision. Particularly, high quality results can be achieved for videos
with the salient target object. Videos with multiple similar looking objects pose a challenge for grounding models, as they have problems preserving object identities across frames. 
Experimentally we show that better proposals, grounding and proximity of training and test data can further boost the performance for videos with multiple objects.
Language is complementary to mask supervision and can be exploited as an additional source of guidance for segmentation.

\section{Conclusion}

In this work we propose the task of video object segmentation using language referring expressions. We propose an approach to address this new task as well as extend two well-known video object segmentation benchmarks with textual descriptions of target objects.
Our experiments indicate that language alone can be successfully exploited to obtain high quality segmentations of objects in videos. While allowing a more natural human-computer interaction, using guidance from
language descriptions can also make video segmentation more robust to occlusions, complex dynamics and cluttered backgrounds.
We show that classical semi-supervised video object segmentation which uses the mask annotation on the first frame can be further improved by the use of language descriptions.
We believe there is a lot of potential in fusing lingual (referring expressions) and visual (clicks, scribbles or masks) forms of supervision for object segmentation in video. We hope that our results encourage more research on video object segmentation with referring expressions and foster discovery of new techniques applicable in realistic settings, which discard tedious pixel-level annotations.

%\clearpage

\bibliographystyle{splncs04}
\bibliography{egbib}

\clearpage
\newpage

\appendix
\vspace{0.5em}
\section*{\large{Supplementary Material}}
\vspace{0.5em}
\setcounter{table}{0}
\renewcommand{\thetable}{\Alph{section}\arabic{table}}
\setcounter{figure}{0}
\renewcommand{\thefigure}{\Alph{section}\arabic{figure}}

This supplementary material provides additional quantitative and qualitative
results and is structured as follows.

Section \ref{sec:sup_datasets} discusses two types of referring expressions - 1st frame vs. full video - and the effect of 1st frame annotations being invalid for the whole video.  It also provides additional examples of the collected referring expressions for video object segmentation task (see Figure \ref{fig:sup_Lang_annot}). 

Section \ref{sec:sup_ground} provides additional evaluation of natural language grounding models on the Lingual ImageNet Videos \cite{LiCVPR2017} and compares results with the work of \cite{LiCVPR2017} (Table \ref{tab:comparative-result-LingualImageNet}).

Section \ref{sec:sup_vos} provides additional evaluation metrics for $\text{DAVIS}_{\text{16}}$ (Table \ref{tab:comparative-result-davis-16}) 
and comparisons of different grounding models, effect of temporal consistency and annotation types on video object segmentation task (Table \ref{tab:comparative-result-VOS-davis17-grounding}). 
We also include more qualitative examples for Language, Mask and Mask + Language approaches (see Figures \ref{fig:sup_qualitative-results-vos}-\ref{fig:sup_qualitative-results-mask-vos}).

\section{\label{sec:sup_datasets} Referring expressions for video object segmentation}

As our goal is to segment objects in videos using language specifications, we augment all objects annotated with mask labels in $\text{DAVIS}_{\text{16}}$ \cite{Perazzi2016Cvpr} and $\text{DAVIS}_{\text{17}}$ \cite{Pont-Tuset_arXiv_2017}
with non-amb{-}iguous referring expressions.

\begin{wrapfigure}{r}{0.5\linewidth}
	\begin{centering}
		\setlength{\tabcolsep}{0em}
		\renewcommand{\arraystretch}{0}
		\par\end{centering}
	\begin{centering}
		\hfill{}%
		\begin{tabular}{@{}c@{}c@{}c@{}}
			\multicolumn{3}{c}{\fontsize{8}{7.2}\selectfont \textit{Original query: "A brown camel" vs.} } \tabularnewline
			\includegraphics[width=0.32\linewidth]{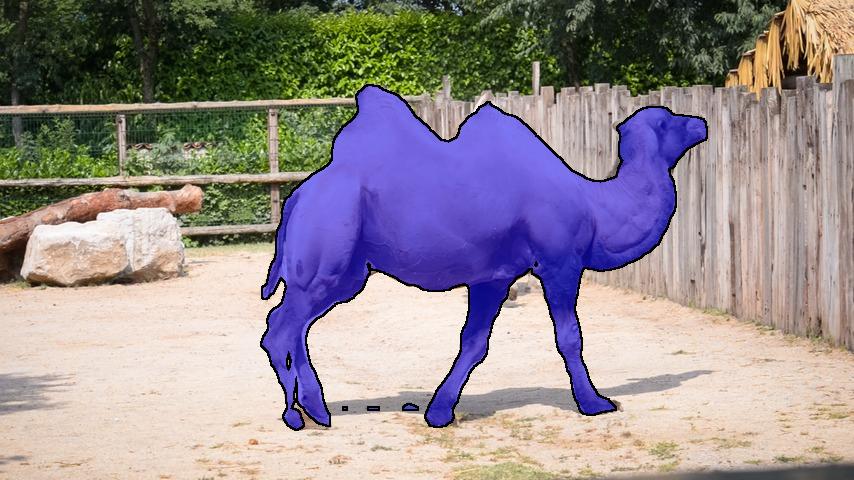} & {\footnotesize{}}
			\includegraphics[width=0.32\linewidth]{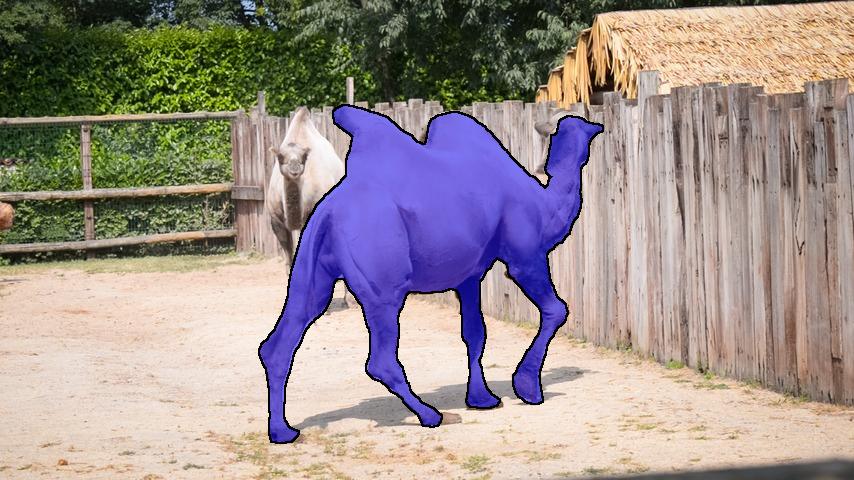} & {\footnotesize{}}
		   \includegraphics[width=0.32\linewidth]{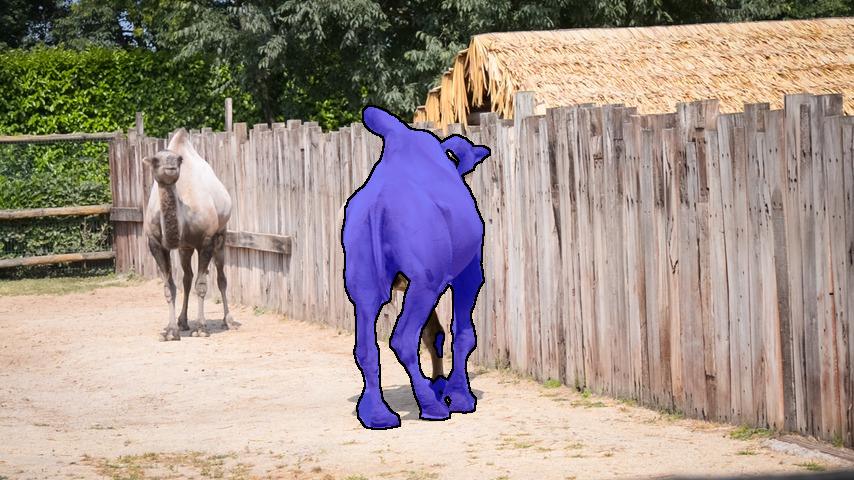}\tabularnewline
			\multicolumn{3}{c}{\fontsize{8}{7.2}\selectfont\textit{Corrected: "A brown camel in the front"} } \tabularnewline
			\includegraphics[width=0.32\linewidth]{figures/rebuttal/00000} & {\footnotesize{}}
			\includegraphics[width=0.32\linewidth]{figures/rebuttal/00039} & {\footnotesize{}}
			\includegraphics[width=0.32\linewidth]{figures/rebuttal/00089}\tabularnewline
		\end{tabular}\hfill{}
		\par\end{centering}
	\caption{\label{fig:qualitative-results-vos} Predictions for the ambiguous query and its correction.}
\end{wrapfigure}

We collected referring expression annotations using two different settings, asking the annotators to provide a description of the target object based on the first frame only as well as on the full video. Future work may choose which setting they prefer more.

We experiment with both annotation types. While the first type is more similar to image-based referring expressions, the second type has different trends, tending to be more complex/long due to increased complexity of the video. We report the grounding (Table 1 in the main paper) and VOS results (Table \ref{tab:comparative-result-VOS-davis17-grounding}) with both types, showing that DBNet \cite{2017-cvpr-dbnet} benefits from the "full video" descriptions, while MattNet \cite{yu2016modeling} has difficulties coping with more complex language.

Concerned that the referring expressions obtained by only looking at the 1st frame might be potentially invalid for the entire video, on $\text{\footnotesize DAVIS}_{\text{17}}$ we ask a user to mark which 1st frame expressions become ambiguous/invalid over time, and to correct them to be valid for the full video (e.g. Fig \ref{fig:qualitative-results-vos}). Only $\sim\!\!15\%$ of all descriptions were marked invalid.  
Though some descriptions become ambiguous/invalid over time, it does not impact strongly the results (original 36.9 vs. corrected 37.1 mIoU). One of the reasons is that \emph{temporal consistency} helps to disambiguate some of such cases (Fig \ref{fig:qualitative-results-vos}). Another reason is that invalid descriptions might still contain valid info (e.g. ``a boy in red on the left'', the boy is no longer on the left, but still in red).

\begin{figure}[t]
\begin{centering}
\hfill{}
\begin{tabular}{c@{\hskip 0.01in}c@{\hskip 0.01in}c@{\hskip 0.01in}c@{\hskip 0.01in}c@{\hskip 0.01in}c}
\includegraphics[width=0.15\linewidth]{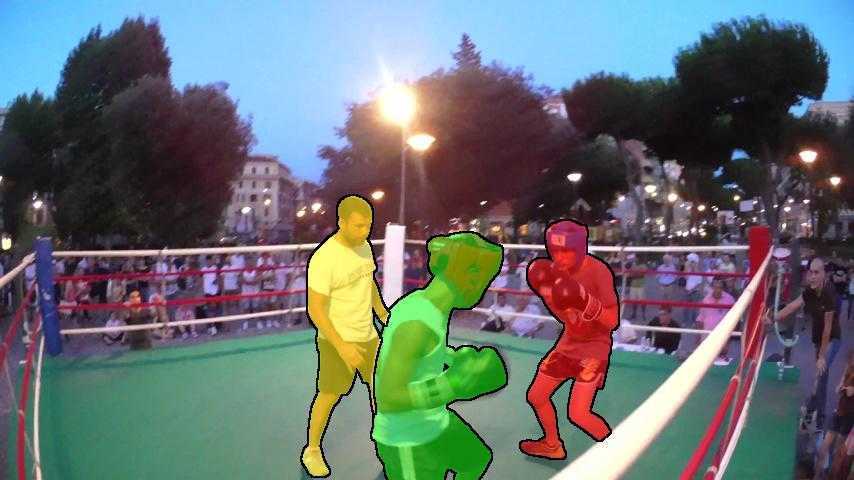} & {\footnotesize{}}
\includegraphics[width=0.15\linewidth]{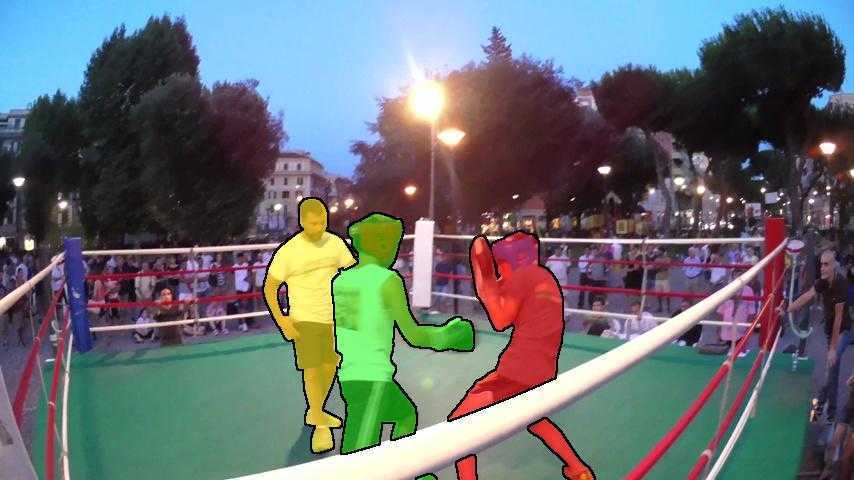} & {\footnotesize{}}
\includegraphics[width=0.15\linewidth]{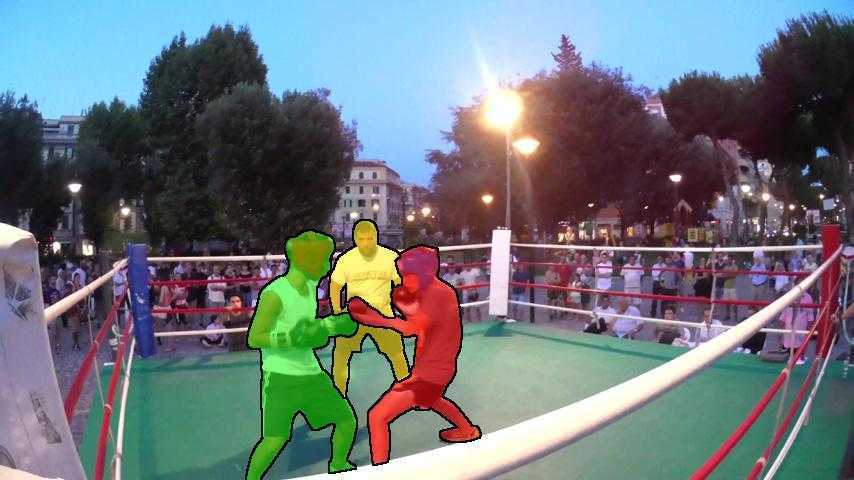} & {\footnotesize{}}
\includegraphics[width=0.15\linewidth]{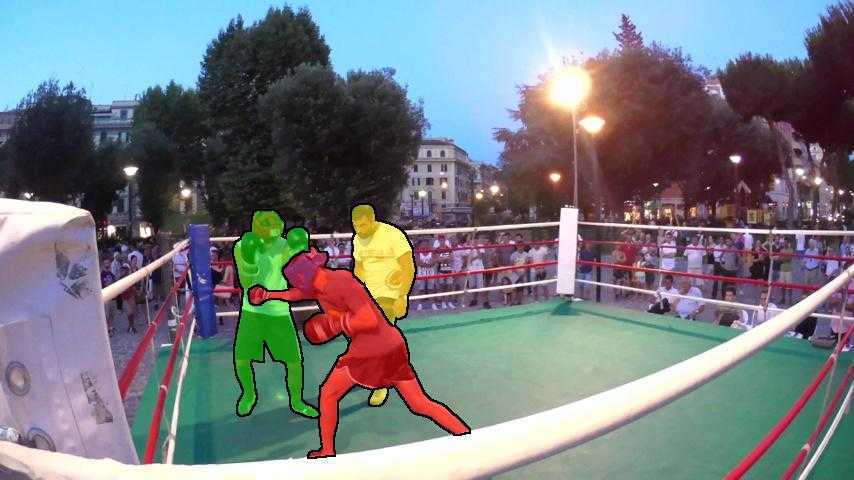} & {\footnotesize{}}
\includegraphics[width=0.15\linewidth]{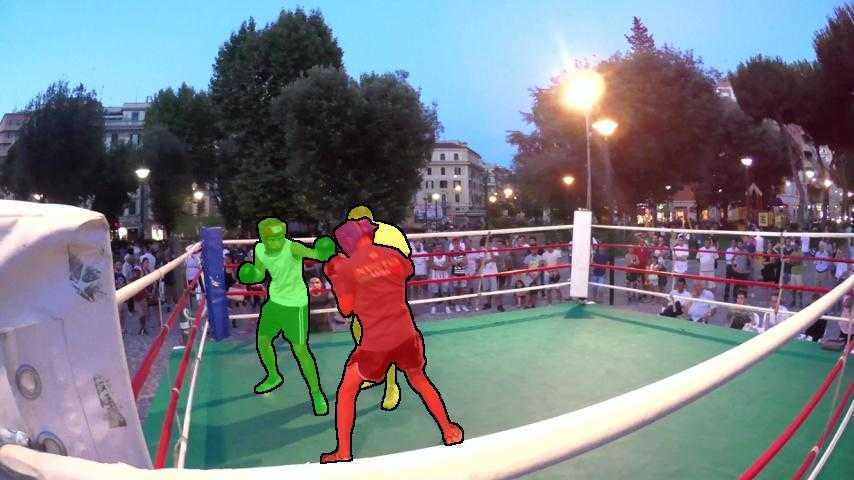} & {\footnotesize{}}
\includegraphics[width=0.15\linewidth]{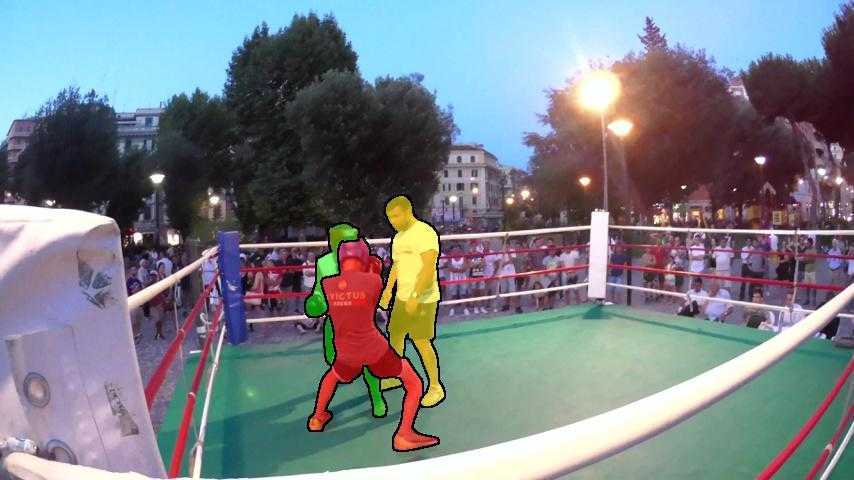}\tabularnewline
\multicolumn{3}{c}{\scriptsize{\textit{\textcolor{mygreen}{ID 1}: "A man on the left wearing blue"}} } & \multicolumn{3}{c}{\scriptsize{\textit{\textcolor{mygreen}{ID 1}: "A man in a blue dress on the left getting punched"}} } \tabularnewline
\multicolumn{3}{c}{\scriptsize{\textit{\textcolor{myred}{ID 2}: "A man on the right wearing red"}} } & \multicolumn{3}{c}{\scriptsize{\textit{ \textcolor{myred}{ID 2}: "A man in a red dress on the right punching"}} } \tabularnewline
\multicolumn{3}{c}{\scriptsize{\textit{\textcolor{myyellow}{ID 3}: "A referee in the middle in white"}} } & \multicolumn{3}{c}{\scriptsize{\textit{\textcolor{myyellow}{ID 3}: "A man in a white shirt and black shorts in the middle"}} } \tabularnewline

\multicolumn{6}{c}{\textit{\vspace{0.5em} }} \tabularnewline

\includegraphics[width=0.15\linewidth]{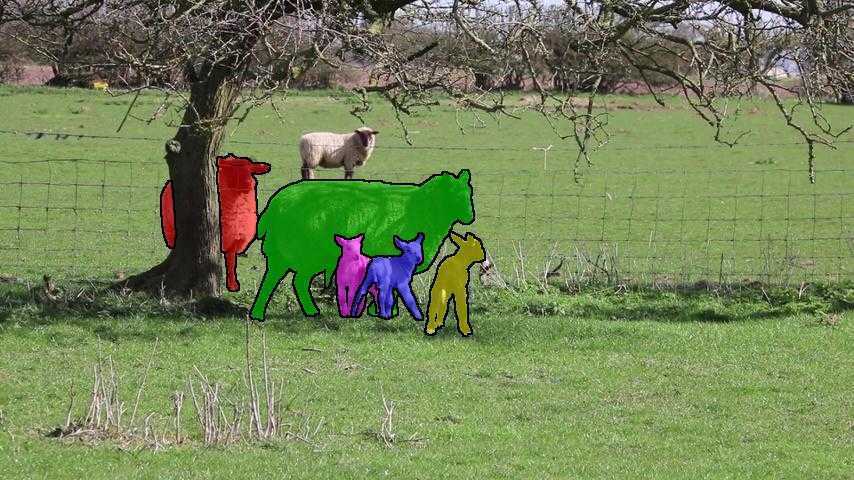} & {\footnotesize{}}
\includegraphics[width=0.15\linewidth]{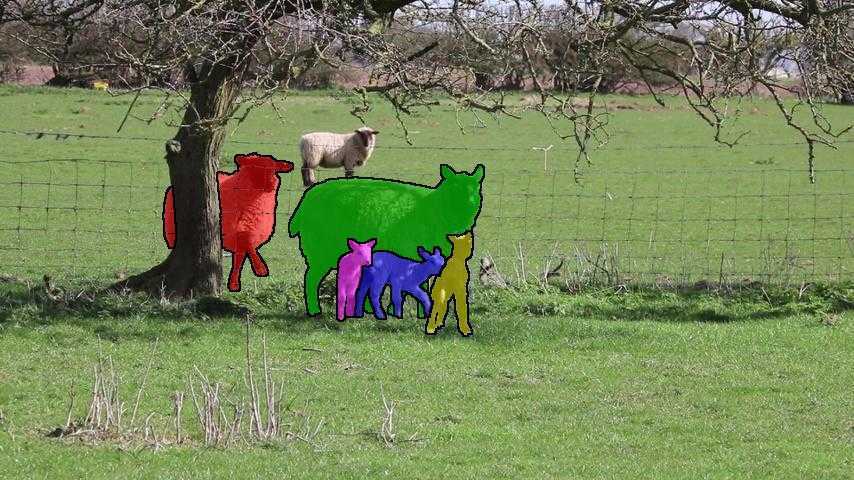} & {\footnotesize{}}
\includegraphics[width=0.15\linewidth]{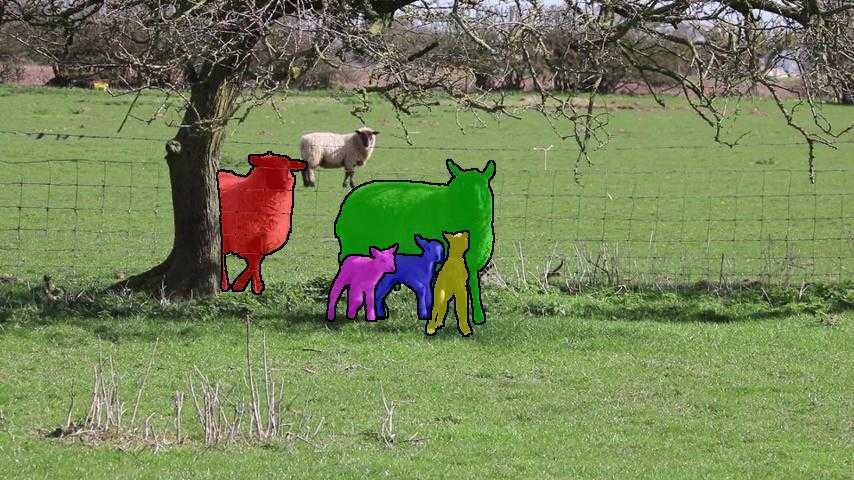} & {\footnotesize{}}
\includegraphics[width=0.15\linewidth]{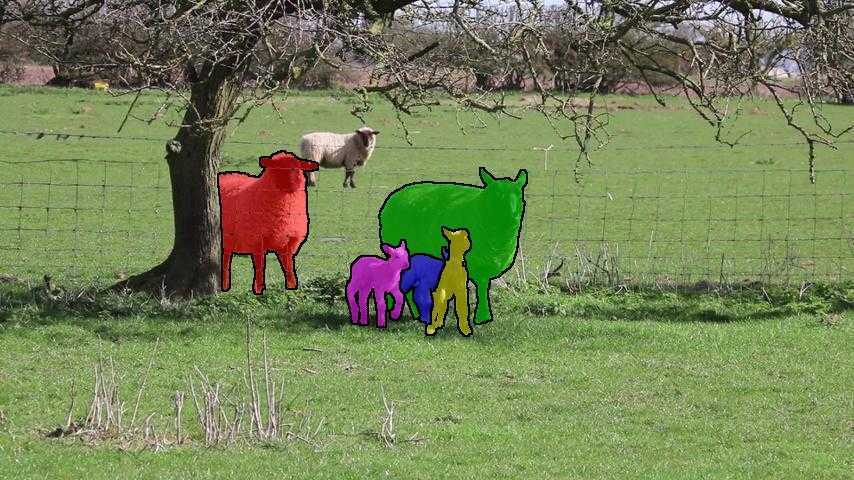} & {\footnotesize{}}
\includegraphics[width=0.15\linewidth]{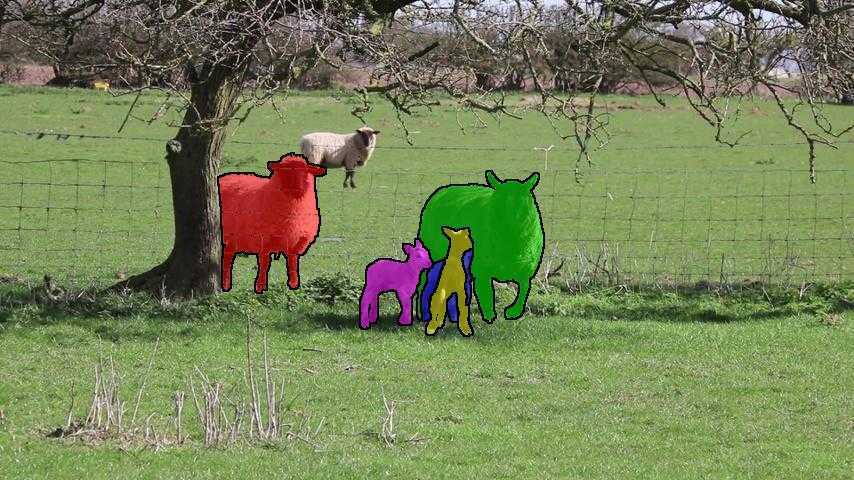} & {\footnotesize{}}
\includegraphics[width=0.15\linewidth]{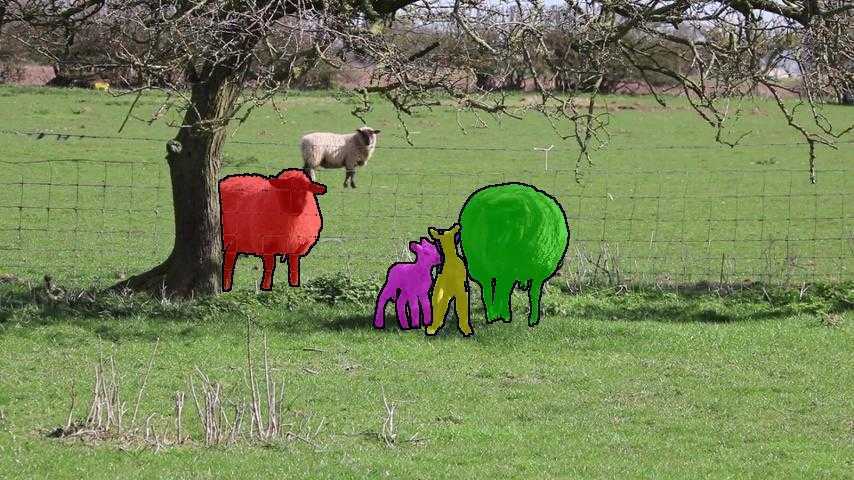}\tabularnewline
\multicolumn{3}{c}{\scriptsize{\textit{\textcolor{mygreen}{ID 1}: "A brown sheep in the middle"}} } & \multicolumn{3}{c}{\scriptsize{\textit{\textcolor{mygreen}{ID 1}: "A brown sheep in the front"}} } \tabularnewline
\multicolumn{3}{c}{\scriptsize{\textit{\textcolor{myred}{ID 2}: "A sheep on the left with a black face"}} } & \multicolumn{3}{c}{\scriptsize{\textit{ \textcolor{myred}{ID 2}: "A grey sheep with dark face moving behind fence"}} } \tabularnewline
\multicolumn{3}{c}{\scriptsize{\textit{\textcolor{myyellow}{ID 3}: "A black lamb with white nose"}} } & \multicolumn{3}{c}{\scriptsize{\textit{\textcolor{myyellow}{ID 3}: "A black baby sheep"}} } \tabularnewline
\multicolumn{3}{c}{\scriptsize{\textit{\textcolor{mypink}{ID 4}: "A white lamb next to a brown sheep"}} } & \multicolumn{3}{c}{\scriptsize{\textit{ \textcolor{mypink}{ID 4}: "A white baby sheep closer to a brown sheep"}} } \tabularnewline
\multicolumn{3}{c}{\scriptsize{\textit{\textcolor{myblue2}{ID 5}: "A white lamb in the middle next to a white sheep"}} } & \multicolumn{3}{c}{\scriptsize{\textit{\textcolor{myblue2}{ID 5}: "A white baby sheep farther from a brown sheep"}} } \tabularnewline

\multicolumn{6}{c}{\textit{\vspace{0.5em} }} \tabularnewline

\includegraphics[width=0.15\linewidth]{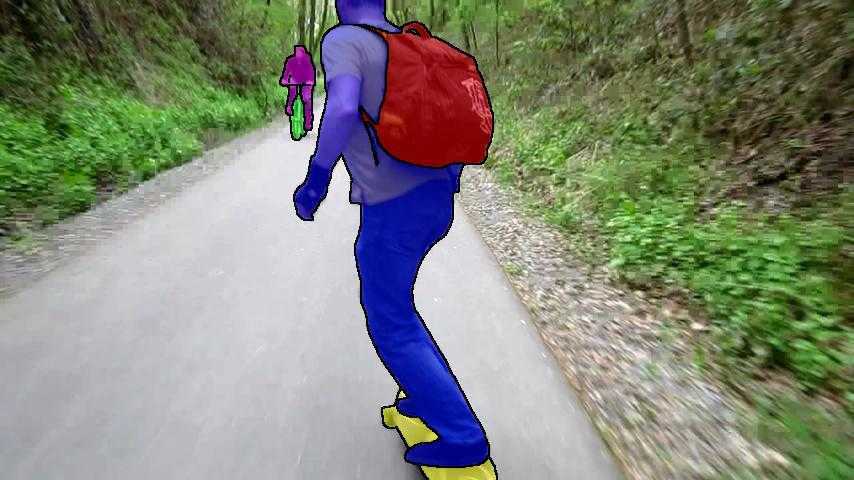} & {\footnotesize{}}
\includegraphics[width=0.15\linewidth]{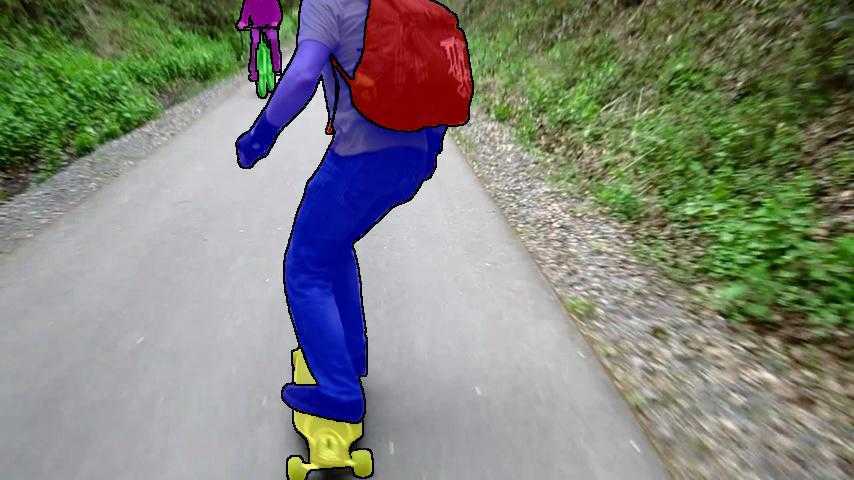} & {\footnotesize{}}
\includegraphics[width=0.15\linewidth]{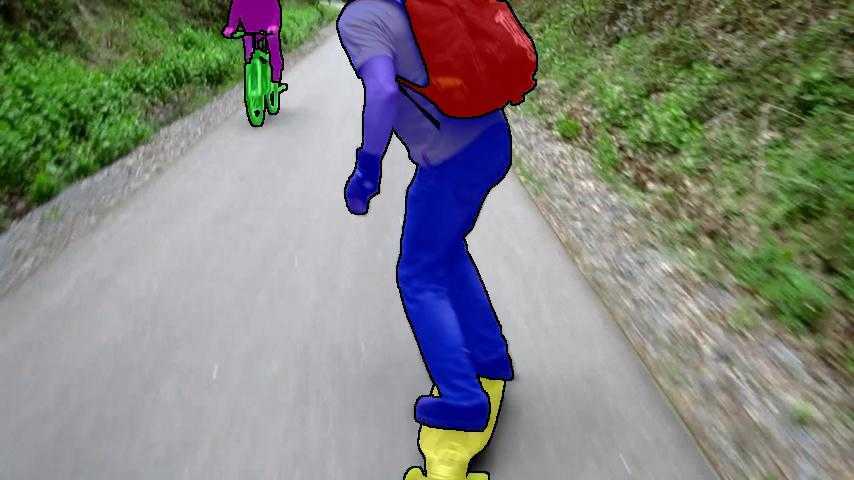} & {\footnotesize{}}
\includegraphics[width=0.15\linewidth]{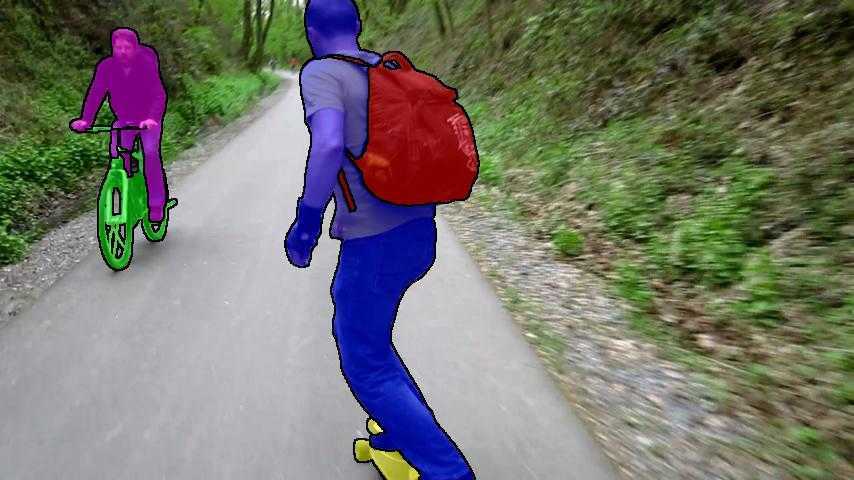} & {\footnotesize{}}
\includegraphics[width=0.15\linewidth]{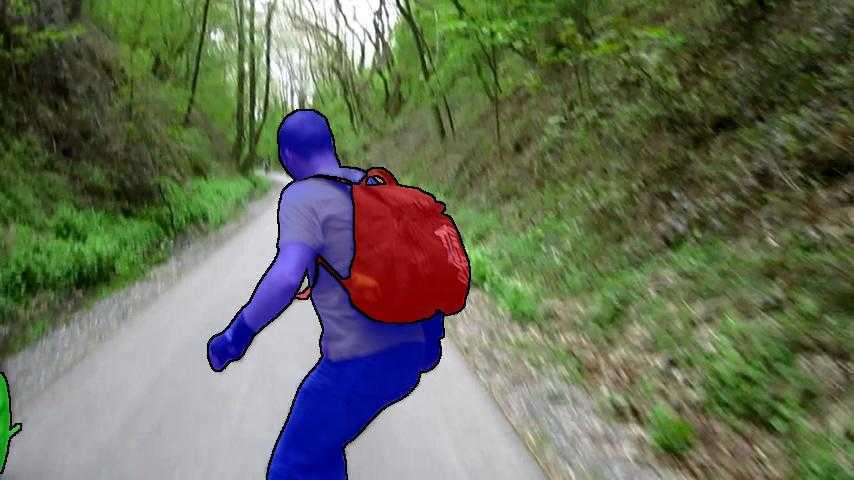} & {\footnotesize{}}
\includegraphics[width=0.15\linewidth]{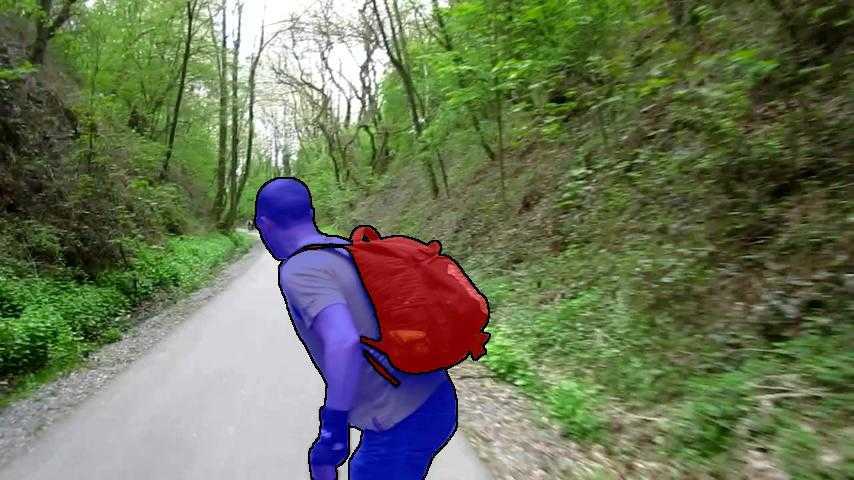}\tabularnewline
\multicolumn{3}{c}{\scriptsize{\textit{\textcolor{mygreen}{ID 1}: "A black bicycle"}} } & \multicolumn{3}{c}{\scriptsize{\textit{\textcolor{mygreen}{ID 1}: "A bicycle moving on the road"}} } \tabularnewline
\multicolumn{3}{c}{\scriptsize{\textit{\textcolor{myred}{ID 2}: "A backpack"}} } & \multicolumn{3}{c}{\scriptsize{\textit{ \textcolor{myred}{ID 2}: "A backpack worn by a guy"}} } \tabularnewline
\multicolumn{3}{c}{\scriptsize{\textit{\textcolor{myyellow}{ID 3}: "A black board"}} } & \multicolumn{3}{c}{\scriptsize{\textit{\textcolor{myyellow}{ID 3}: "A longboard"}} } \tabularnewline
\multicolumn{3}{c}{\scriptsize{\textit{\textcolor{mypink}{ID 4}: "A man on a bicycle in a black jacket"}} } & \multicolumn{3}{c}{\scriptsize{\textit{ \textcolor{mypink}{ID 4}: "A guy riding a bicycle"}} } \tabularnewline
\multicolumn{3}{c}{\scriptsize{\textit{\textcolor{myblue2}{ID 5}: "A man in a yellow t-shirt"}} } & \multicolumn{3}{c}{\scriptsize{\textit{\textcolor{myblue2}{ID 5}: "A person rolling over longboard"}} } \tabularnewline
\multicolumn{6}{c}{\textit{\vspace{0em} }} \tabularnewline
\multicolumn{3}{c}{\footnotesize{First frame annotation}} & \multicolumn{3}{c}{\footnotesize{Full video annotation}} \tabularnewline

\end{tabular}\hfill{}
\par\end{centering}
\vspace{0.5em}
\caption{\label{fig:sup_Lang_annot} Example of collected annotations provided for the first frame (left) vs. the full video (right).
Full video annotations include descriptions of activities and overall are more complex than the ones provided for the first frame.
}
\vspace{0em}
\end{figure}

We present additional examples of collected referring expressions in Figure \ref{fig:sup_Lang_annot}.

\section{\label{sec:sup_ground}Language grounding results on Lingual ImageNet Videos}

\begin{wraptable}{r}{0.5\linewidth}
	\setlength{\tabcolsep}{0.1em} 
	\renewcommand{\arraystretch}{0.95}
	\begin{centering}
		\begin{tabular}{c|c|c}
			\multirow{1}{*}{{\footnotesize{}{Method}}}& \multirow{1}{*}{{\footnotesize{}{Supervision}}} & \multicolumn{1}{c}{\footnotesize{}AUC score}\tabularnewline
			\hline 
			\hline 
			\multirow{3}{*}{{\footnotesize{}{\begin{tabular}{c}{\footnotesize{}{Tracking by}}\tabularnewline {\footnotesize{}{language \cite{LiCVPR2017}}}\tabularnewline \end{tabular}}}} &  {\small{}Language} & {\small{}26.3}  \tabularnewline
			& {\small{}Box } & {\small{}47.9} \tabularnewline
			& {\small{}Box + Language} & {{\small{} 49.4}} \tabularnewline
			\hline
			\footnotesize{}{DBNet} &  {\small{}Language} & {\small{}54.0}  \tabularnewline
			\footnotesize{}{MattNet} &  {\small{}Language} & \textbf{\small{}60.8}  \tabularnewline
		\end{tabular}
		\par\end{centering}
	\caption{\label{tab:comparative-result-LingualImageNet}Comparison of grounding models on Lingual ImageNet Videos, val set. 
	}
\end{wraptable}

For the natural language grounding task we additionally consider Lingual ImageNet Videos \cite{LiCVPR2017}, which provides referring expression annotations for a subset of 
the ImageNet Video Object Detection dataset \cite{russakovsky2015ijcv}.
The dataset is split into a training and a validation set, each consisting of $50$ videos.
The performance on Lingual ImageNet \cite{LiCVPR2017} is measured in terms of the AUC (area under the curve) score metric, following \cite{LiCVPR2017}.

Here we compare to \cite{LiCVPR2017}, who perform tracking of objects using language specifications. 
Table \ref{tab:comparative-result-LingualImageNet} presents grounding results reported by \cite{LiCVPR2017}, including tracking by language only, 
tracking given the ground-truth bounding box on the first frame, and the combined approach. 
Our method is based on language input only, specifically, we report the results after the temporal consistency step applied to DBNet and MattNet predictions. 
As we see both models significantly outperform \cite{LiCVPR2017}, 
even when \cite{LiCVPR2017} has access to the ground-truth bounding box on the first frame.

\section{\label{sec:sup_vos} Video object segmentation}

\subsection{\label{sec:voc-davis16_additional} Additional metrics for $\text{DAVIS}_{\text{16}}$}

We report video object segmentation results for the $\text{DAVIS}_{\text{16}}$ benchmark in Table \ref{tab:comparative-result-davis-16}, using evaluation metrics proposed in
\cite{Perazzi2016Cvpr}. Three measures are used: region similarity
in terms of intersection-over-union ($J$, higher is better), contour accuracy ($F$, higher
is better), and temporal instability of the masks ($T$, lower is better).
See \cite{Perazzi2016Cvpr} for more details.
Note that using only language supervision results in a smaller decay over time for $J$ and $F$ measures and a better overall temporal stability $T$ compared to employing pixel-level mask supervision on the first frame.

\begin{table*}[t]

\setlength{\tabcolsep}{0.1em} 
\renewcommand{\arraystretch}{1.2}
\begin{centering}
\hspace*{-1em}%
\begin{tabular}{c cc|ccc|ccc|c}
 \multicolumn{2}{c}{\multirow{3}{*}{{\footnotesize{}{Supervision}}}} & \multirow{3}{*}{{\footnotesize{}Method}} & \multicolumn{7}{c}{{\small{}$\text{DAVIS}_{\text{16}}$}}\tabularnewline
 
  &  &  & \multicolumn{3}{c|}{{\footnotesize{}Region,}{\small{} \ensuremath{J} }} & \multicolumn{3}{c|}{{\footnotesize{}Boundary,}{\small{} \ensuremath{F} }} & \begin{tabular}{c}
{\footnotesize{}Temp.}\tabularnewline
{\footnotesize{} stab.,}{\small{} \ensuremath{T}}\tabularnewline
\end{tabular}
\tabularnewline
 &  &   & {\footnotesize{}Mean $\uparrow$} & {\footnotesize{}Recall $\uparrow$} & {\footnotesize{}Decay $\downarrow$} & {\footnotesize{}Mean $\uparrow$} & {\footnotesize{}Recall $\uparrow$} & {\footnotesize{}Decay $\downarrow$} & {\footnotesize{}Mean $\downarrow$}\tabularnewline

\hline 
\hline 
\multicolumn{2}{c}{
\multirow{1}{*}{ 
{\footnotesize{Oracle}}%
\begin{tabular}{c}
{\footnotesize{}}\tabularnewline
\end{tabular}{\footnotesize{} }
}} 
 & {\small{}Mask R-CNN \cite{He_2017_ICCV}}  & {\small{}71.5} & 87.3 & 5.9 & 72.4 & 84.6 & 6.8 & 24.8 \tabularnewline
\hline 

\multicolumn{2}{c}{
\multirow{7}{*}{ 
{\footnotesize{}}%
\begin{tabular}{c}
{\footnotesize{}Unsupervised}\tabularnewline
\end{tabular}{\footnotesize{} }
}} 
 & {\small{}NLC \cite{Faktor2014Bmvc}}  & {\small{}55.1} & 55.8 & 12.6 & 52.3 & 51.9 & 11.4 &42.5\tabularnewline
\multicolumn{2}{c}{} & {\small{}FST\cite{papazoglou2013}}  & {\small{}55.8} & 64.9 & 0.0 & 51.1 & 51.6 &2.9 & 36.6\tabularnewline
 \multicolumn{2}{c}{} &{\small{}SegFlow\cite{Cheng_ICCV_2017}}  & {\small{}67.4} & 81.4 & 6.2 & 66.7 & 77.1 &5.1 &28.2 \tabularnewline
\multicolumn{2}{c}{} & {\small{}MP-Net \cite{Tokmakov2016Arxiv}}  & {\small{}70.0} & 85.0 & 1.3 & 65.9 & 79.2 & 2.5 & 57.2\tabularnewline
\multicolumn{2}{c}{} & {\small{}FusionSeg \cite{Jain2017ArxivFusionSeg}} & {\small{}70.7} & 83.5 & 1.5 &65.3 & 73.8 &1.8 & 32.8\tabularnewline
\multicolumn{2}{c}{} & {\small{}LVO \cite{TokmakovAS17}}  & {\small{}75.9} &89.1 &0.0 & 72.1 & 8.4 &1.3 &26.5\tabularnewline
\multicolumn{2}{c}{} & {\small{}ARP \cite{Koh_CVPR_2017}} & {\small{}76.2} &91.1 & 7.0 & 70.6 &83.5 &7.9 &39.3\tabularnewline
\hline 

\multirow{14}{*}{\begin{turn}{90}
{\footnotesize{}}%
\begin{tabular}{c}
{\footnotesize{}Semi-supervised}\tabularnewline
\end{tabular}{\footnotesize{} }
\end{turn}
} 
&
\multirow{12}{*}{
{\footnotesize{}}%
\begin{tabular}{c}
{\footnotesize{}1st frame}\tabularnewline
{\footnotesize{}mask}\tabularnewline
\end{tabular}{\footnotesize{} }
} 
 & {\small{}FCP \cite{Perazzi2015Iccv}}  & {\small{}58.4} & 71.5 & \textbf{-2.0} & 49.2 & 49.5 & \textbf{-1.1} & 30.6 \tabularnewline
& & {\small{}BVS \cite{Maerki2016Cvpr}}  & {\small{}60.0} & 66.9 & 28.9 & 58.8 & 67.9 & 21.3 & 34.7\tabularnewline
& & {\small{}ObjFlow \cite{Tsai2016Cvpr}}  & {\small{}68.0}  & 75.6 & 26.4 & 63.4 & 70.4 & 27.2 & 22.2\tabularnewline
&& {\small{}PLM \cite{Yoon_2017_ICCV}}  & {\small{}70.2}& 86.3 &11.2 & 62.5 &73.2 & 14.7 & 31.8 \tabularnewline
 & & {\small{}VPN \cite{Jampani2016Arxiv}} & {\textcolor{black}{\small{}70.2}} & 82.3 & 12.4 & 65.5 & 69.0 & 14.4  & 32.4\tabularnewline
& & {\small{}CTN \cite{Jang_CVPR_2017}} & {\textcolor{black}{\small{}73.5}} & 87.4 & 15.6 & 69.3 & 79.6 & 12.9 & 22.0\tabularnewline
& & {\small{}SegFlow \cite{Cheng_ICCV_2017}} & {\small{}76.1} & 90.6 & 12.1 &76.0 & 85.5 & 10.4 & \textbf{18.9}\tabularnewline
& & {\small{}MaskTrack \cite{Khoreva2017CvprMaskTrack}} & {\small{}79.7} & 93.1 & 8.9 & 75.4 & 87.1 & 9.0 & 21.8\tabularnewline
& & {\small{}OSVOS \cite{Caelles2017Cvpr} } & {\textcolor{black}{\small{}79.8}}& 93.6 & 14.9 & 80.6 & 92.6 & 15.0 &37.8\tabularnewline
& & {\small{}MaskRNN \cite{Hu2017MaskRNNIL}} & {\textcolor{black}{\small{}80.4}} & {96.0} & 4.4 & 82.3 & 93.2& 8.8 & 19.0\tabularnewline
& & {\small{}OnAVOS\tablefootnote{OnAVOS gives 86.1 mIoU by online adaptation on successive frames.} \cite{Voigtlaender2017OnlineAO}} & {\textcolor{black}{\small{}81.7}} & 92.2 & 11.9 & 81.1 & 88.2 & 11.2 & 27.3\tabularnewline
& & {\small{}Our } & {\small{}83.1} & 95.1 & 9.8 & 85.7 & 94.4 & 9.6 & 24.0 \tabularnewline
\cline{2-10}

& \multirow{1}{*}{
{\footnotesize{}}%
\begin{tabular}{c}
{\footnotesize{}Language}\tabularnewline
\end{tabular}{\footnotesize{} }
} 
& {\small{}\arrayrulecolor{black}}$\text{Our}$  & {\small{}82.8} & 94.1 & 3.2 & 85.4 & 94.7 & 3.4 & 22.6 \tabularnewline
\cline{2-10}

& \multirow{1}{*}{
{\footnotesize{}}%
\begin{tabular}{c}
{\footnotesize{}Mask + Lang.}\tabularnewline
\end{tabular}{\footnotesize{} }
} 
& {\small{}\arrayrulecolor{black}}$\text{Our}$  & \textbf{\small{}84.5} & \textbf{96.3} & 8.2 & \textbf{86.9} & \textbf{95.9} & 8.7 & 24.8  \tabularnewline

\end{tabular}

\par\end{centering}
\caption{\label{tab:comparative-result-davis-16}Comparison of video object segmentation results
on $\text{DAVIS}_{\text{16}}$, validation set.}
\end{table*}

\begin{table}
\setlength{\tabcolsep}{0.5em} 
\renewcommand{\arraystretch}{1.2}
\begin{centering}
\begin{tabular}{cc|c|cc}
 \multirow{1}{*}{{\footnotesize{}Annotation type}} & \multirow{1}{*}{{ \footnotesize{}{Grounding}}} & \multirow{1}{*}{{\footnotesize{}Temporal consistency}} & {\footnotesize{} mIoU} & {\footnotesize{}$J\& F$} \tabularnewline
\hline 
\hline 
 \multirow{4}{*}{1st frame }&  \multirow{2}{*}{\footnotesize{}{DBNet}} & - & 32.6 & 34.7 \tabularnewline
 &  & \textcolor{black}{\scriptsize{}\Checkmark{}} & 35.4 & 37.6\tabularnewline
 \arrayrulecolor{lightgray} \cline{2-5} \arrayrulecolor{black}
 & \multirow{2}{*}{ \footnotesize{}{MattNet}} & - & 35.4 & 38.5\tabularnewline
 &  & \textcolor{black}{\scriptsize{}\Checkmark{}} &\textbf{ 37.3} & \textbf{39.3}\tabularnewline
\hline 
 \multirow{2}{*}{Full video} & \footnotesize{}{DBNet} & \textcolor{black}{\scriptsize{}\Checkmark{}} & 35.5 & 37.7\tabularnewline
 \arrayrulecolor{lightgray} \cline{2-5} \arrayrulecolor{black}
 & \footnotesize{}{MattNet} & \textcolor{black}{\scriptsize{}\Checkmark{}} & 35.5 & 37.1\tabularnewline

\end{tabular}
\par\end{centering}
\caption{\label{tab:comparative-result-VOS-davis17-grounding} Effect of different grounding models, temporal consistency and annotation types on video object segmentation on $\text{DAVIS}_{\text{17}}$, validation set.}
\end{table}

\subsection{\label{sec:voc-davis17_additional} Effect of grounding models, temporal consistency and annotation types on video object segmentation}

Table \ref{tab:comparative-result-VOS-davis17-grounding} reports the effect of different grounding models, temporal consistency step for grounding and employing 
the first frame versus the full video descriptions on video object segmentation.

We compare DBNet versus MattNet (trained on RefCOCO \cite{yu2016modeling}) as a base grounding model for video object 
segmentation task. Exploiting MattNet grounding boxes results in a better performance compared to DBNet ($37.3$ vs. $35.4$).
Overall the temporal consistency step has a positive impact on video object segmentation performance across different grounding models (for MattNet $35.4 \rightarrow 37.3$ and for DBNet $32.6 \rightarrow 35.4$).

We also compare the segmentation performance from first frame versus full video descriptions in Table \ref{tab:comparative-result-VOS-davis17-grounding}.
Employing the full video versus the first frame descriptions results in a minor improvement for DBNet ($35.4$ vs. $35.5$), however has a negative effect for MattNet ($37.3$ vs. $35.5$).
The same diverging has been observed for language grounding results in the main paper
when comparing results on expressions provided for the first frame versus expressions provided for the full video in Table 2. 
We attribute this to the fact that DBNet is trained on the more diverse Visual Genome descriptions and can handle better more complex full video expressions.

% % >>>DBNet Vs. MattNet: 81.1 vs. 82.2

\begin{figure*}[t!]
\begin{centering}
\setlength{\tabcolsep}{0.1em}
\renewcommand{\arraystretch}{0}
\par\end{centering}
\begin{centering}
\hfill{}%
\begin{tabular}{c@{\hskip 0.01in}c@{\hskip 0.01in}c@{\hskip 0.01in}c@{\hskip 0.01in}c@{\hskip 0.01in}c}

\multicolumn{6}{c}{\textit{ \textcolor{mygreen}{ID 1}: "A red car".} } \tabularnewline
\includegraphics[width=0.16\linewidth]{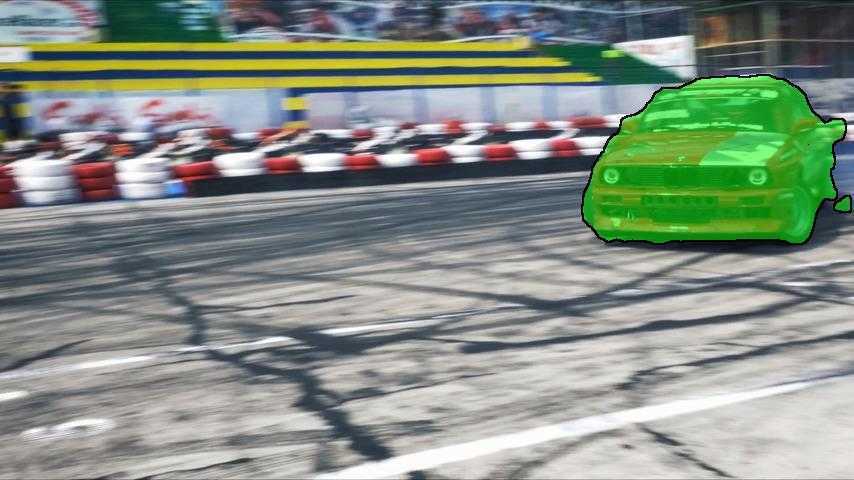} & {\footnotesize{}}
\includegraphics[width=0.16\linewidth]{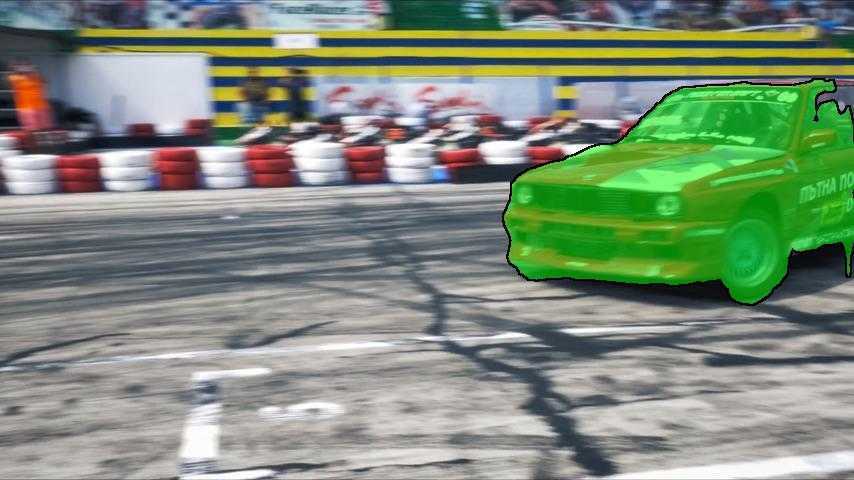} & {\footnotesize{}}
\includegraphics[width=0.16\linewidth]{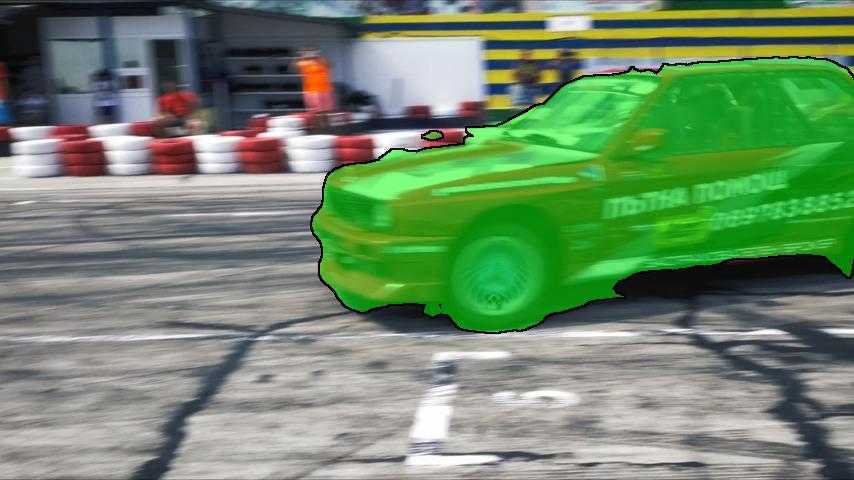} & {\footnotesize{}}
\includegraphics[width=0.16\linewidth]{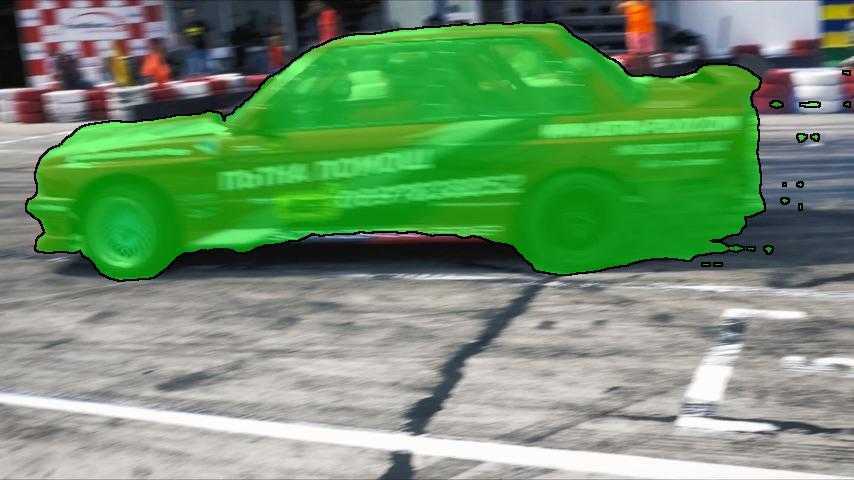} & {\footnotesize{}}
\includegraphics[width=0.16\linewidth]{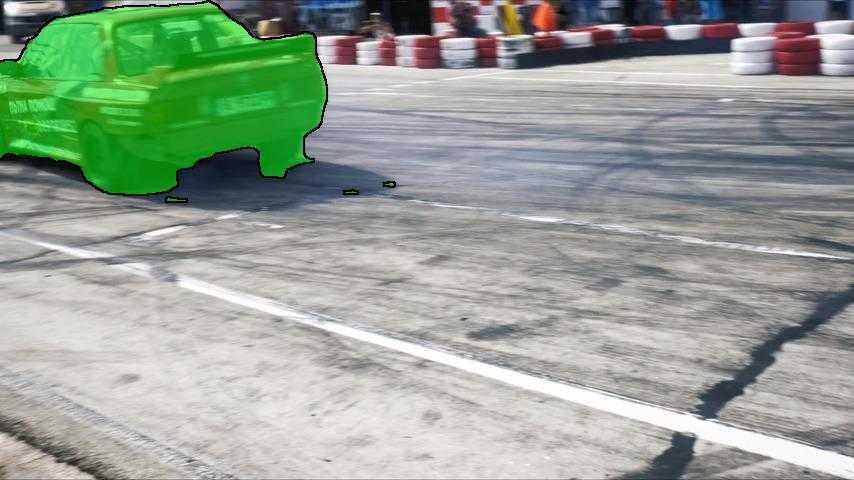} & {\footnotesize{}}
\includegraphics[width=0.16\linewidth]{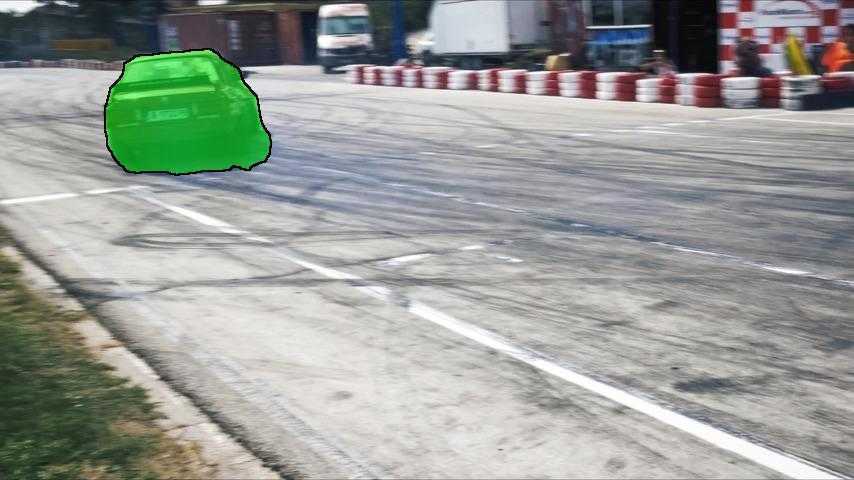}\tabularnewline
\multicolumn{6}{c}{\textit{\vspace{0.05em} }} \tabularnewline

\multicolumn{6}{c}{\textit{ \textcolor{mygreen}{ID 1}: "A man jumping across fences".} } \tabularnewline
\includegraphics[width=0.16\linewidth]{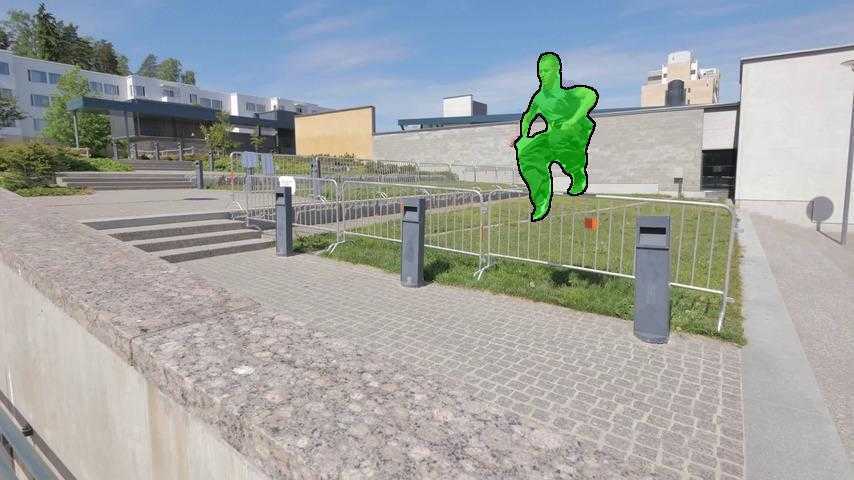} & {\footnotesize{}}
\includegraphics[width=0.16\linewidth]{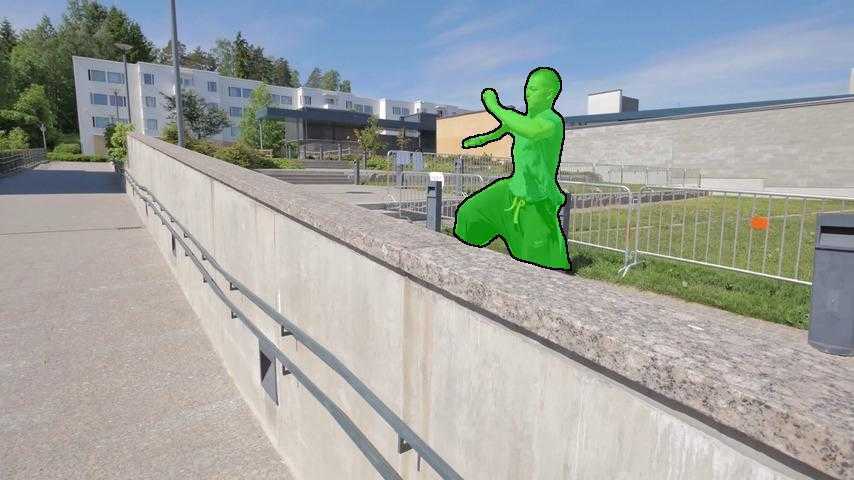} & {\footnotesize{}}
\includegraphics[width=0.16\linewidth]{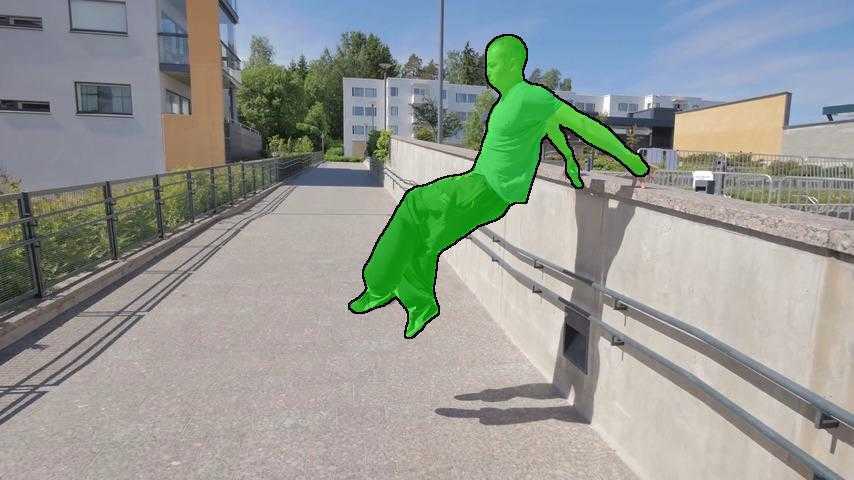} & {\footnotesize{}}
\includegraphics[width=0.16\linewidth]{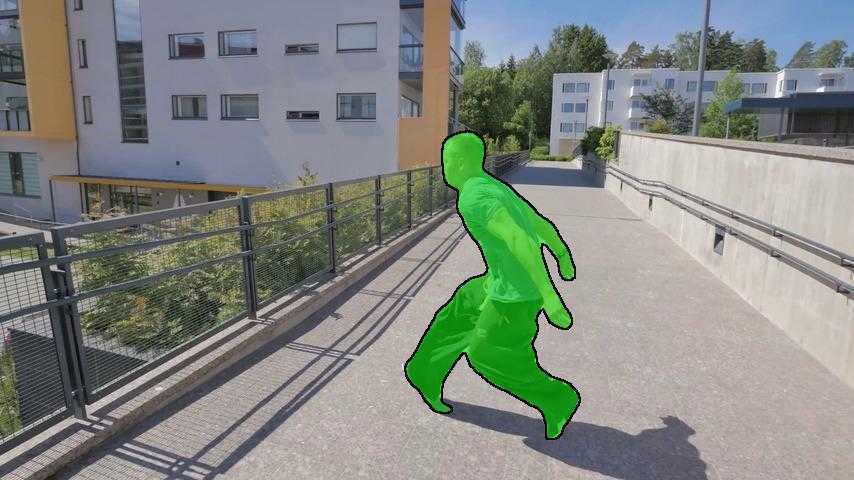} & {\footnotesize{}}
\includegraphics[width=0.16\linewidth]{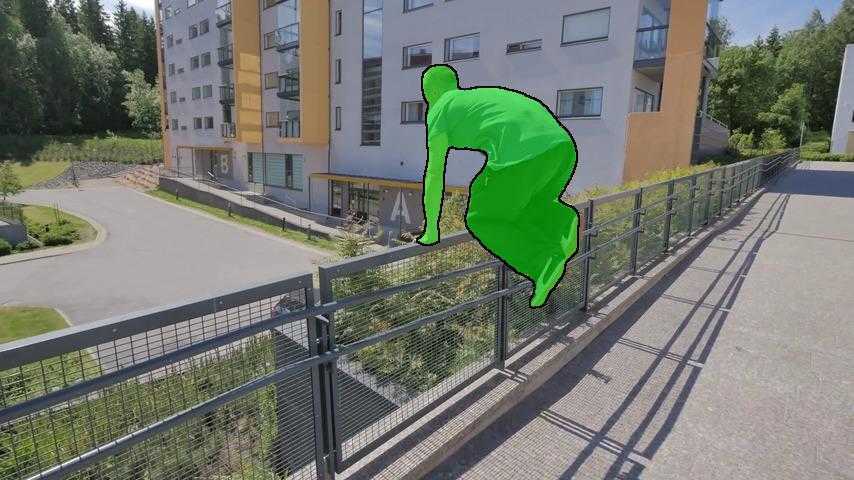} & {\footnotesize{}}
\includegraphics[width=0.16\linewidth]{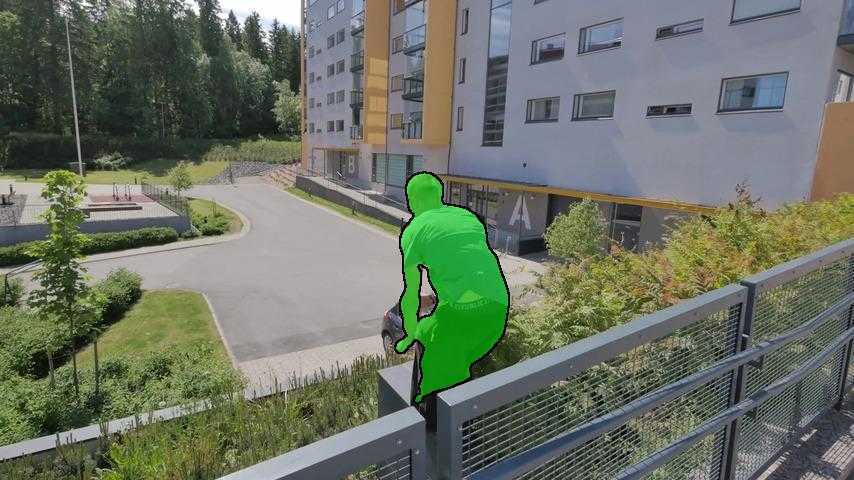}\tabularnewline
\multicolumn{6}{c}{\textit{\vspace{0.05em} }} \tabularnewline

\multicolumn{6}{c}{\textit{ \textcolor{mygreen}{ID 1}: "A dog running in the garden".} } \tabularnewline
\includegraphics[width=0.16\linewidth]{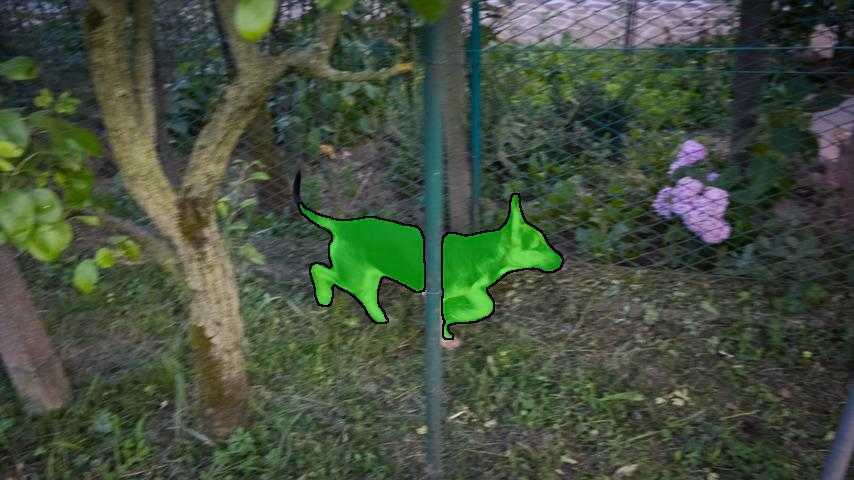} & {\footnotesize{}}
\includegraphics[width=0.16\linewidth]{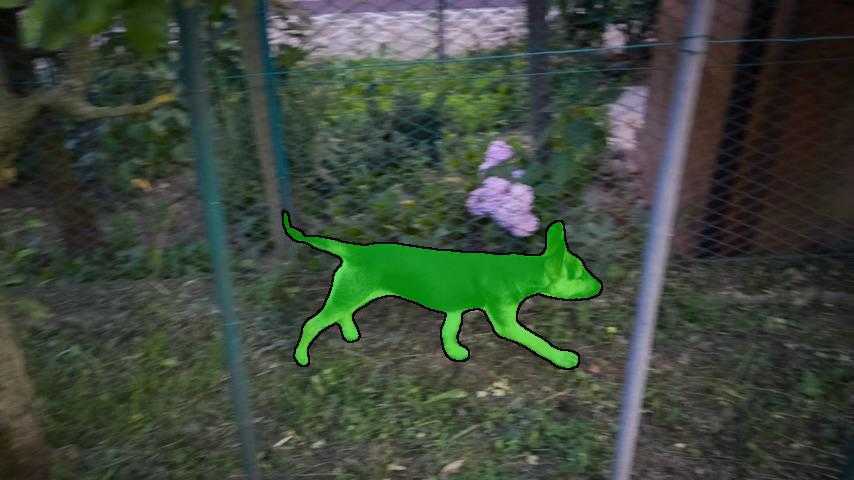} & {\footnotesize{}}
\includegraphics[width=0.16\linewidth]{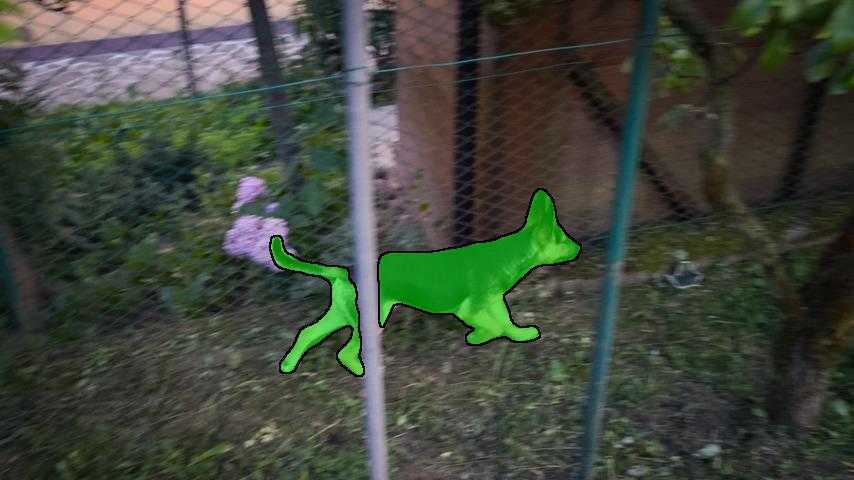} & {\footnotesize{}}
\includegraphics[width=0.16\linewidth]{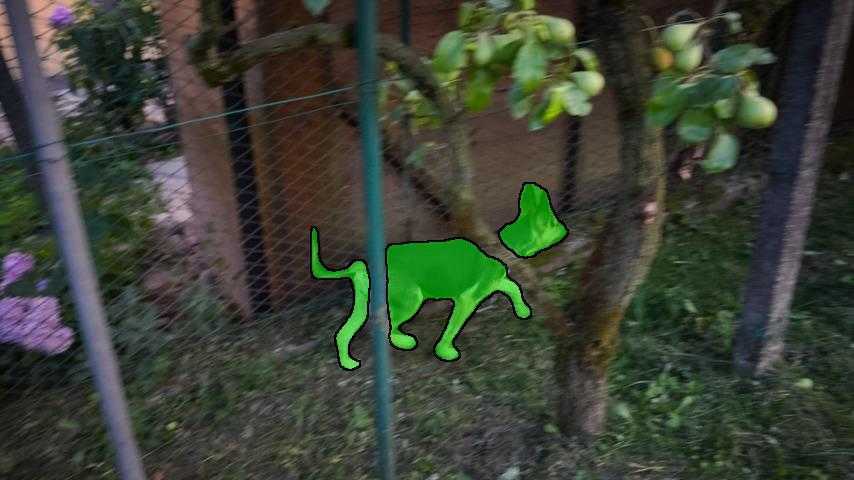} & {\footnotesize{}}
\includegraphics[width=0.16\linewidth]{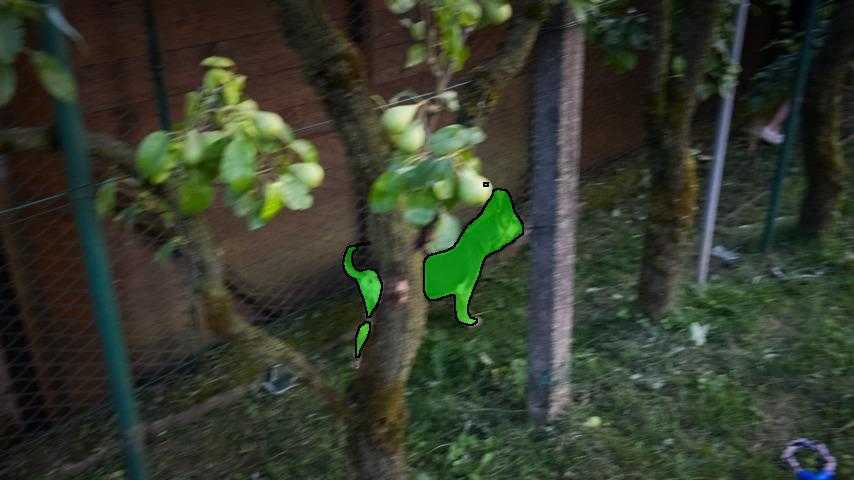} & {\footnotesize{}}
\includegraphics[width=0.16\linewidth]{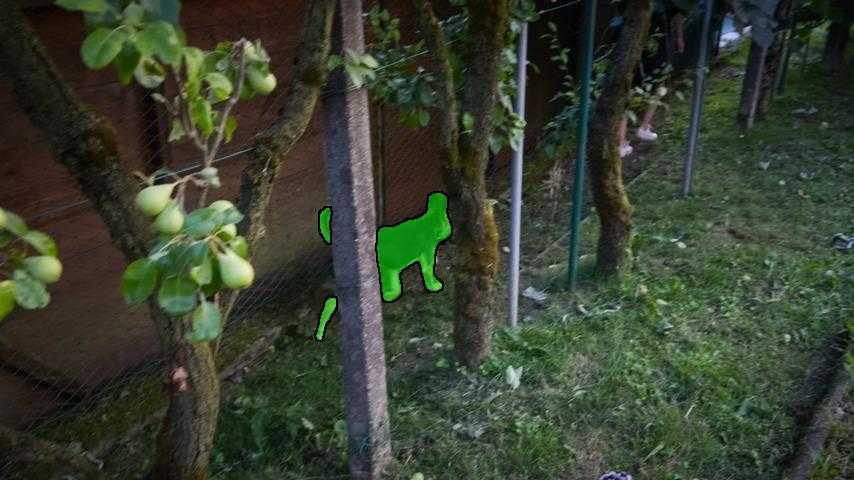}\tabularnewline
\multicolumn{6}{c}{\textit{\vspace{0.05em} }} \tabularnewline

\multicolumn{6}{c}{\textit{ \textcolor{mygreen}{ID 1}: "A goat walking on rocks".} } \tabularnewline
\includegraphics[width=0.16\linewidth]{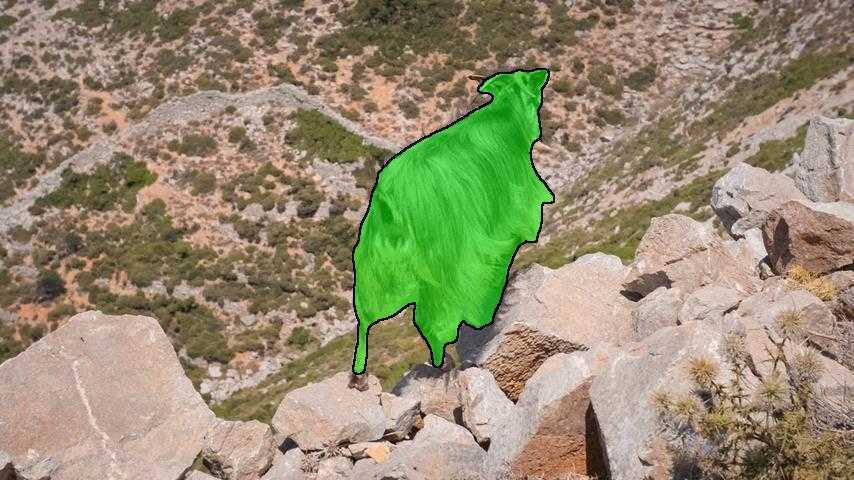} & {\footnotesize{}}
\includegraphics[width=0.16\linewidth]{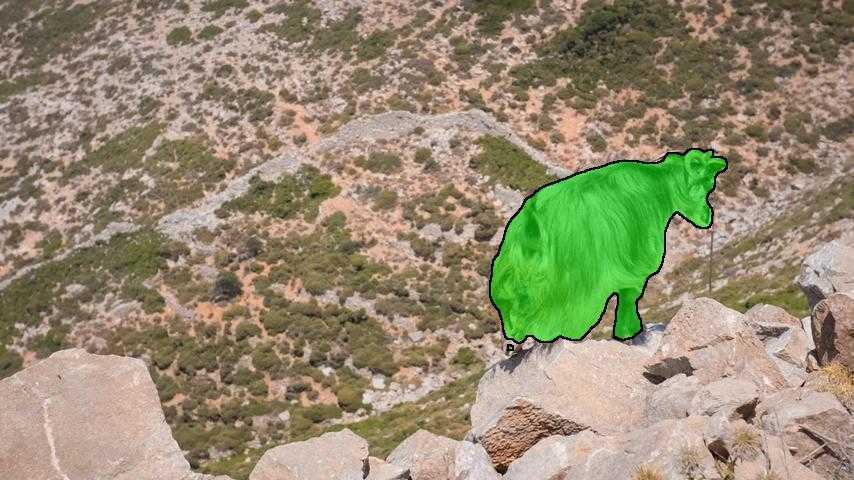} & {\footnotesize{}}
\includegraphics[width=0.16\linewidth]{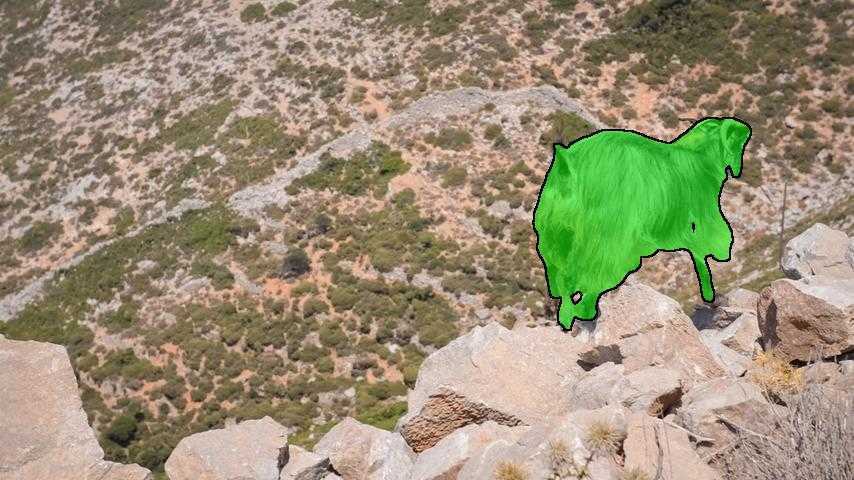} & {\footnotesize{}}
\includegraphics[width=0.16\linewidth]{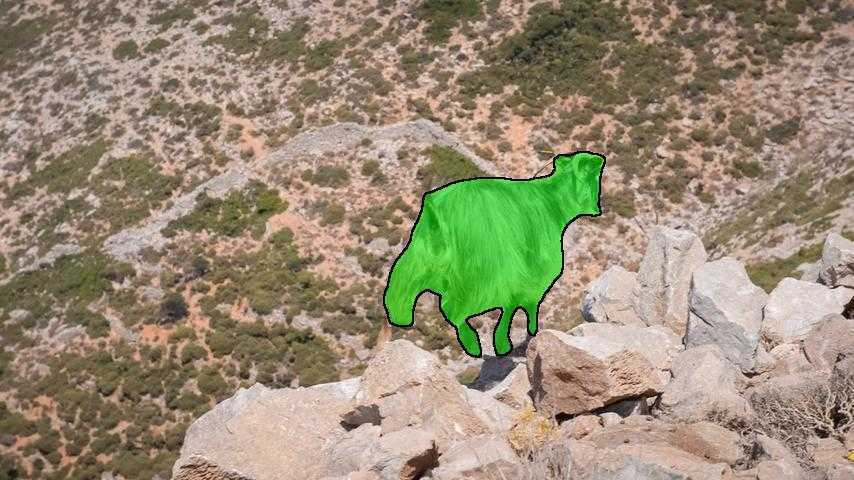} & {\footnotesize{}}
\includegraphics[width=0.16\linewidth]{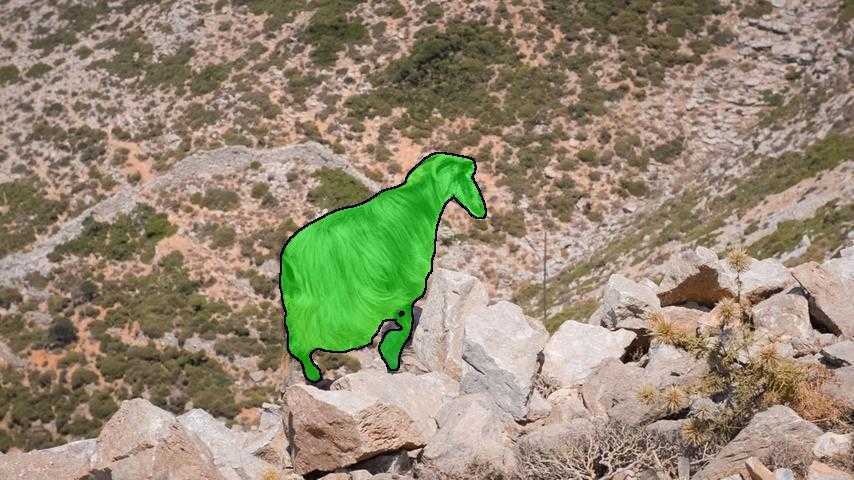} & {\footnotesize{}}
\includegraphics[width=0.16\linewidth]{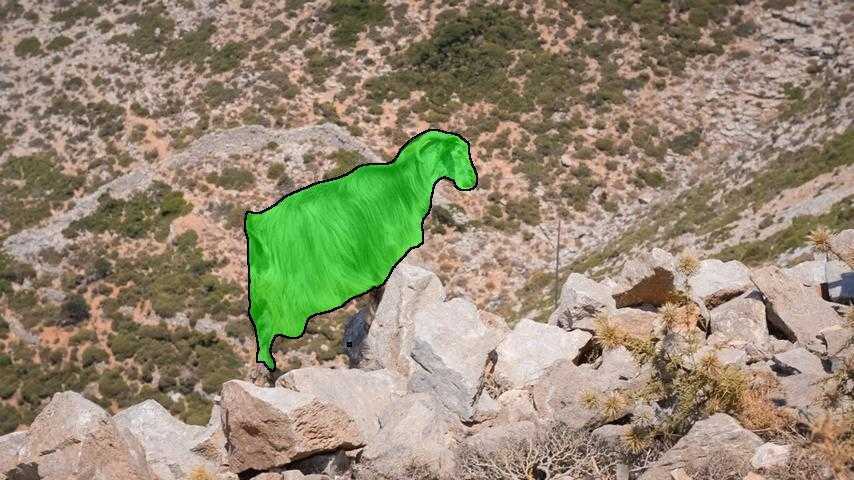}\tabularnewline
\multicolumn{6}{c}{\textit{\vspace{0.05em} }} \tabularnewline

\multicolumn{6}{c}{\textit{ \textcolor{mygreen}{ID 1}: "A red and white car".} } \tabularnewline
\includegraphics[width=0.16\linewidth]{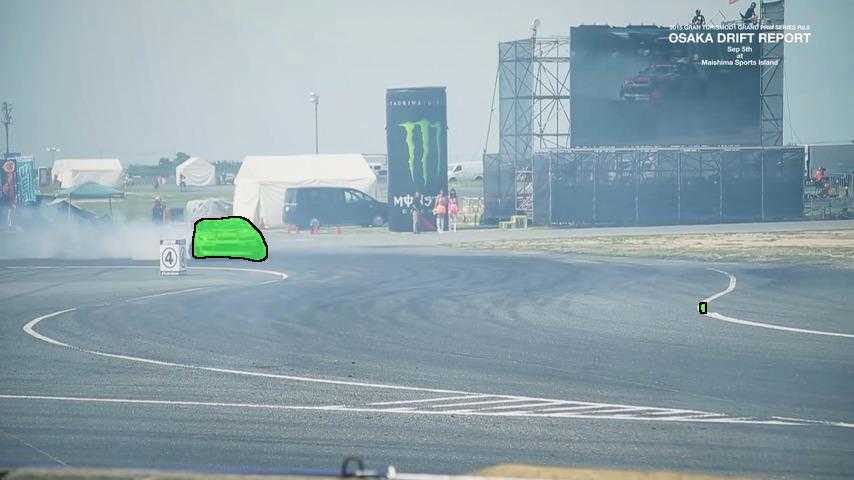} & {\footnotesize{}}
\includegraphics[width=0.16\linewidth]{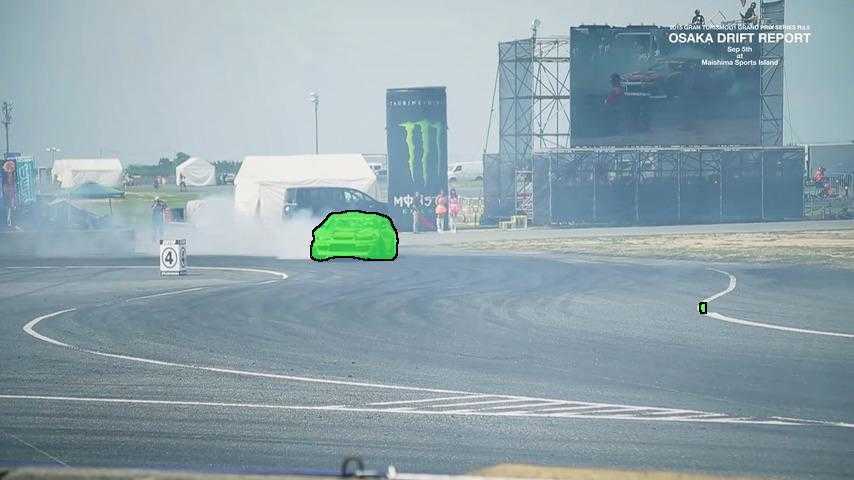} & {\footnotesize{}}
\includegraphics[width=0.16\linewidth]{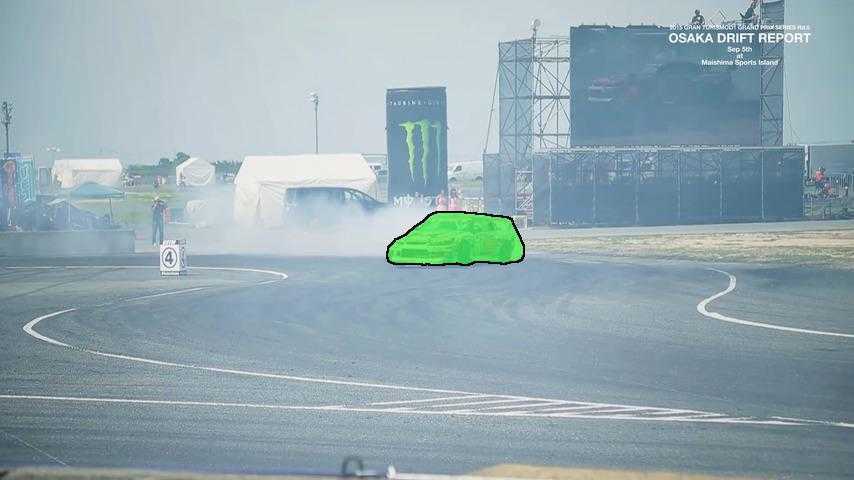} & {\footnotesize{}}
\includegraphics[width=0.16\linewidth]{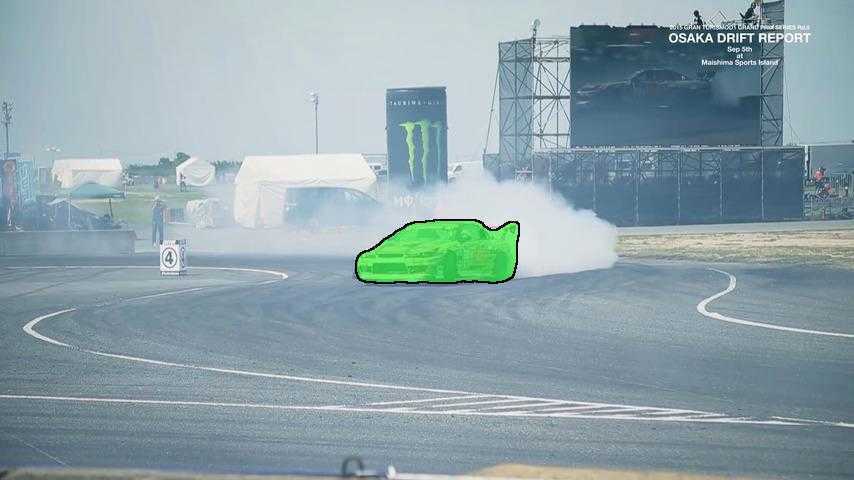} & {\footnotesize{}}
\includegraphics[width=0.16\linewidth]{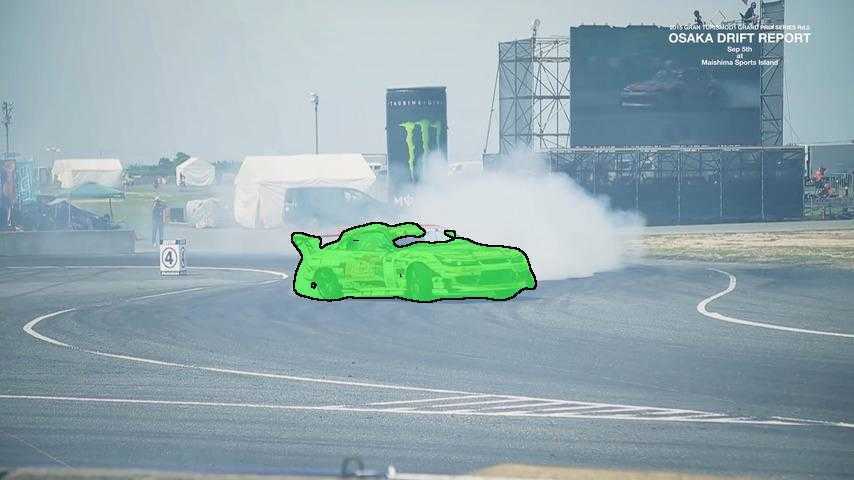} & {\footnotesize{}}
\includegraphics[width=0.16\linewidth]{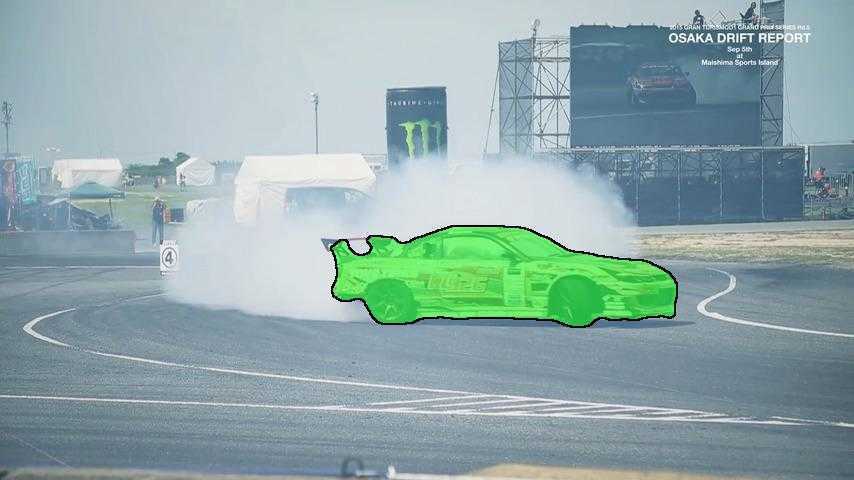}\tabularnewline
\multicolumn{6}{c}{\textit{\vspace{0.05em} }} \tabularnewline

\multicolumn{6}{c}{\textit{ \textcolor{mygreen}{ID 1}: "A woman riding a horse". \textcolor{myred}{ID 2}: "A horse doing high-jumps".} } \tabularnewline
\includegraphics[width=0.16\linewidth]{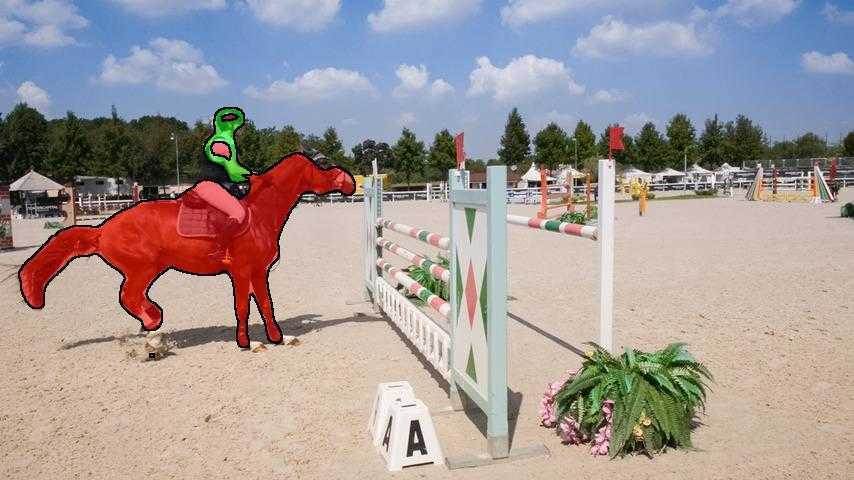} & {\footnotesize{}}
\includegraphics[width=0.16\linewidth]{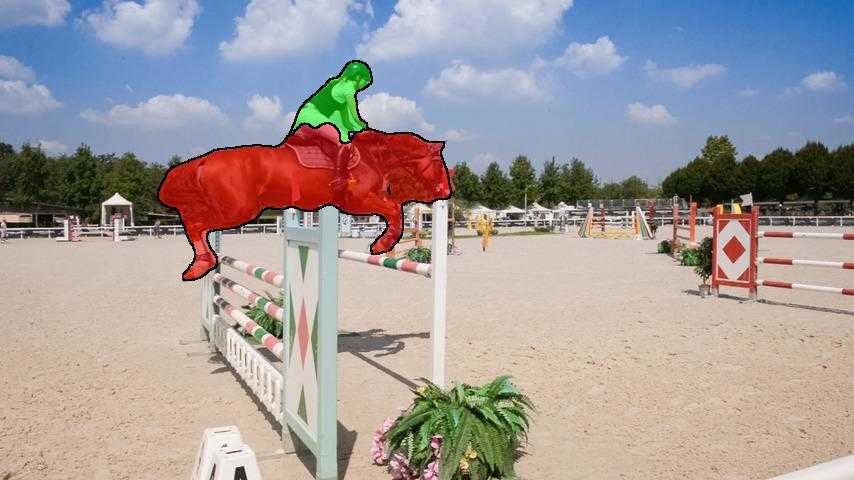} & {\footnotesize{}}
\includegraphics[width=0.16\linewidth]{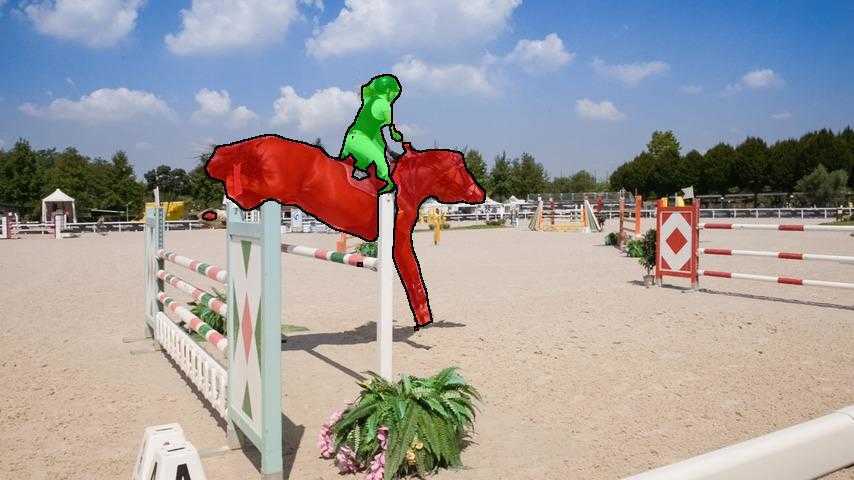} & {\footnotesize{}}
\includegraphics[width=0.16\linewidth]{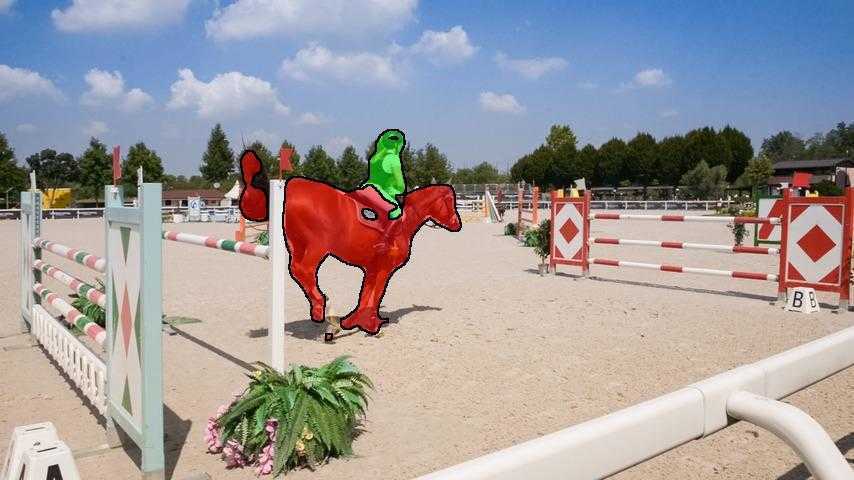} & {\footnotesize{}}
\includegraphics[width=0.16\linewidth]{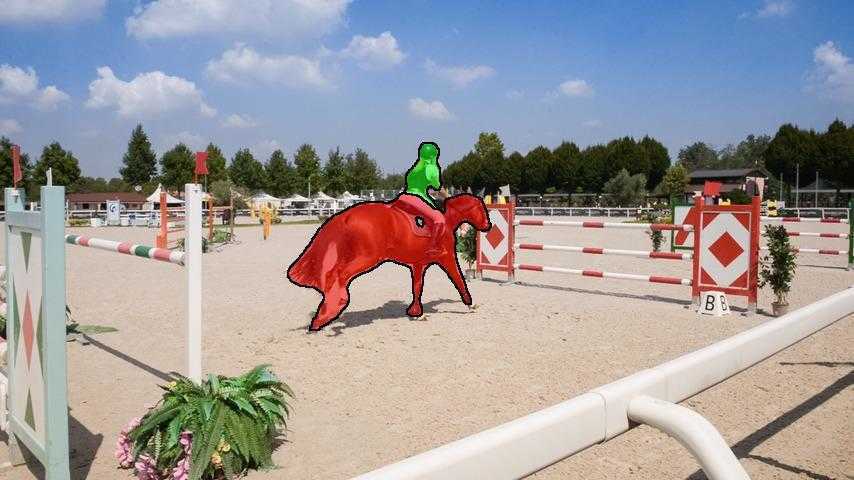} & {\footnotesize{}}
\includegraphics[width=0.16\linewidth]{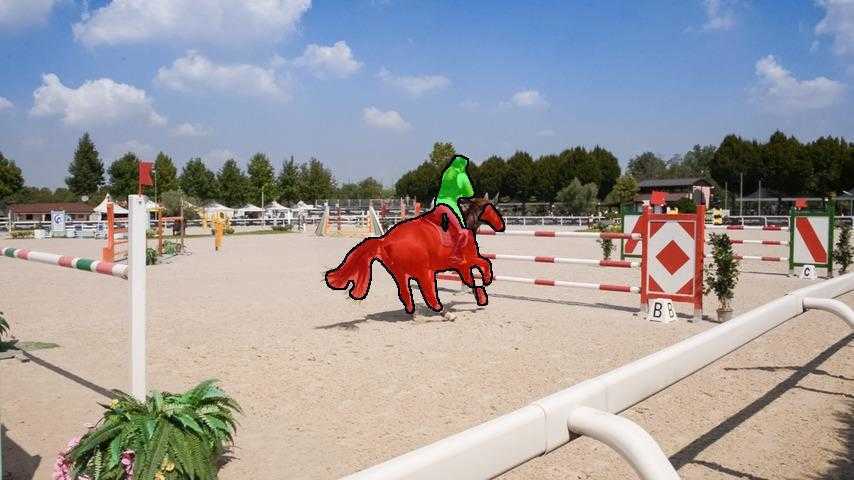}\tabularnewline
\multicolumn{6}{c}{\textit{\vspace{0.05em} }} \tabularnewline

\multicolumn{6}{c}{\textit{\textcolor{mygreen}{ID 1}: "A bald man with black belt in the center". \textcolor{myred}{ID 2}: "A man with blue belt on the right".} } \tabularnewline
\includegraphics[width=0.16\linewidth]{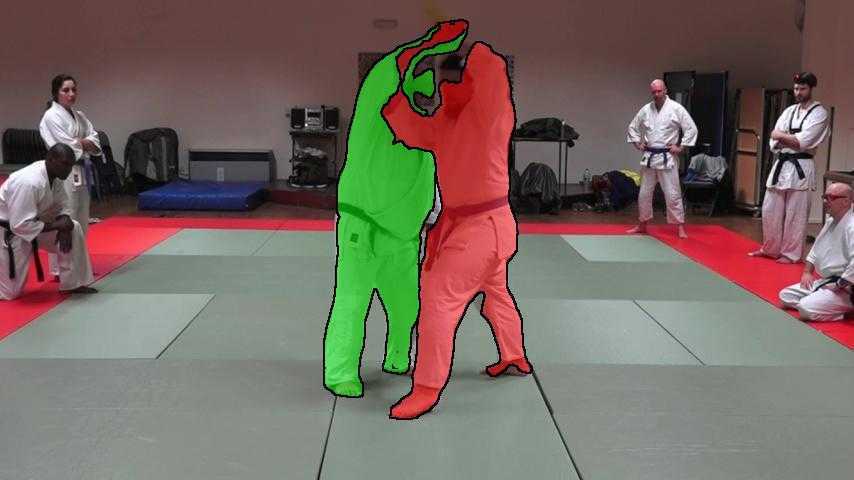} & {\footnotesize{}}
\includegraphics[width=0.16\linewidth]{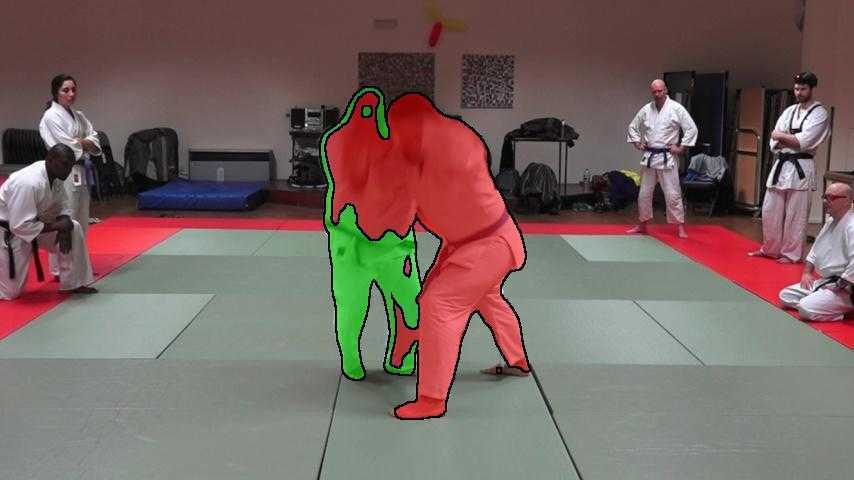} & {\footnotesize{}}
\includegraphics[width=0.16\linewidth]{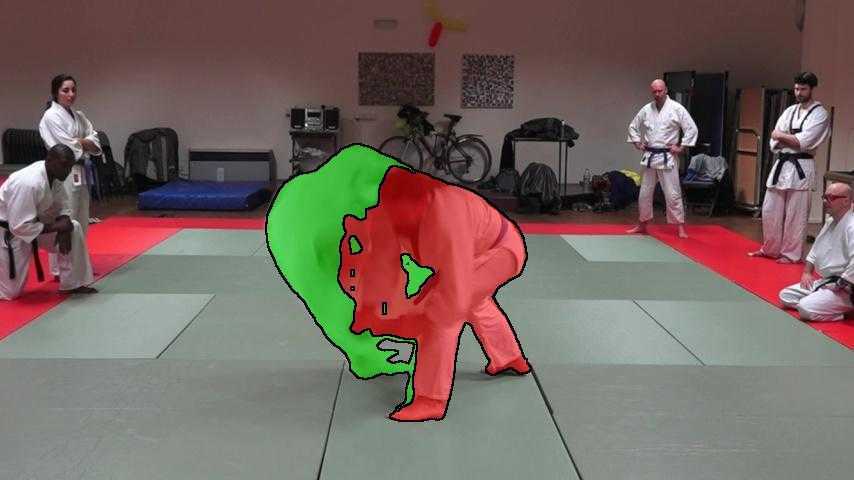} & {\footnotesize{}}
\includegraphics[width=0.16\linewidth]{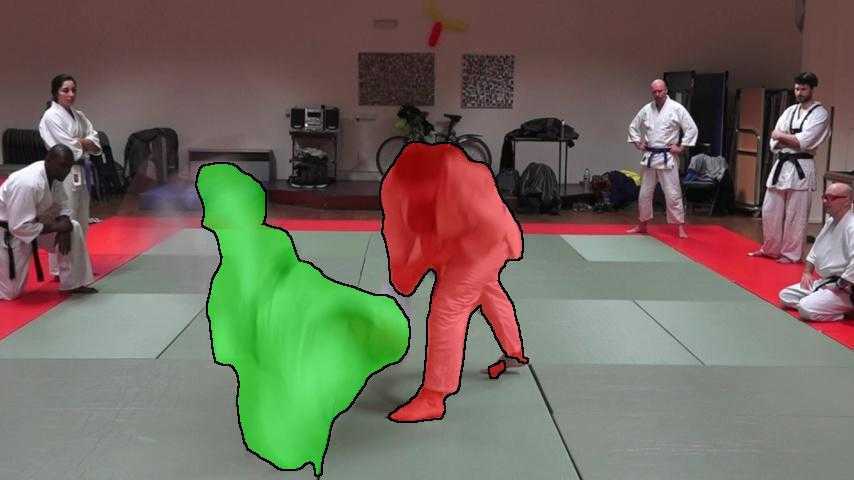} & {\footnotesize{}}
\includegraphics[width=0.16\linewidth]{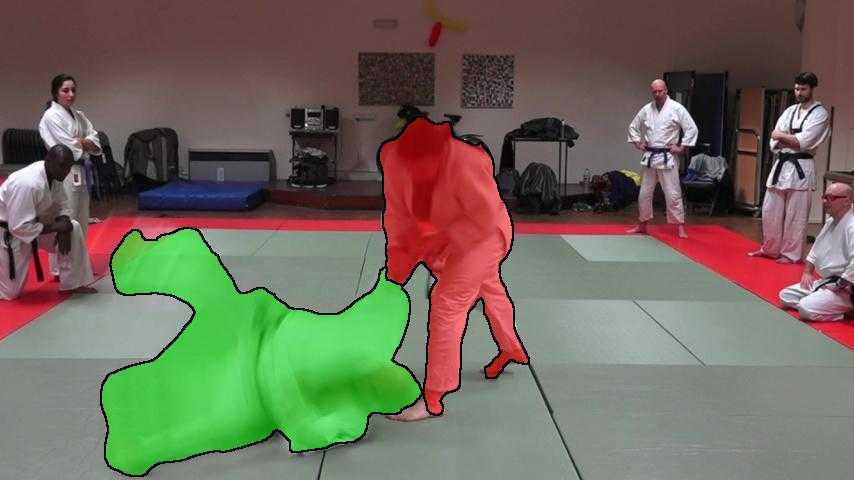} & {\footnotesize{}}
\includegraphics[width=0.16\linewidth]{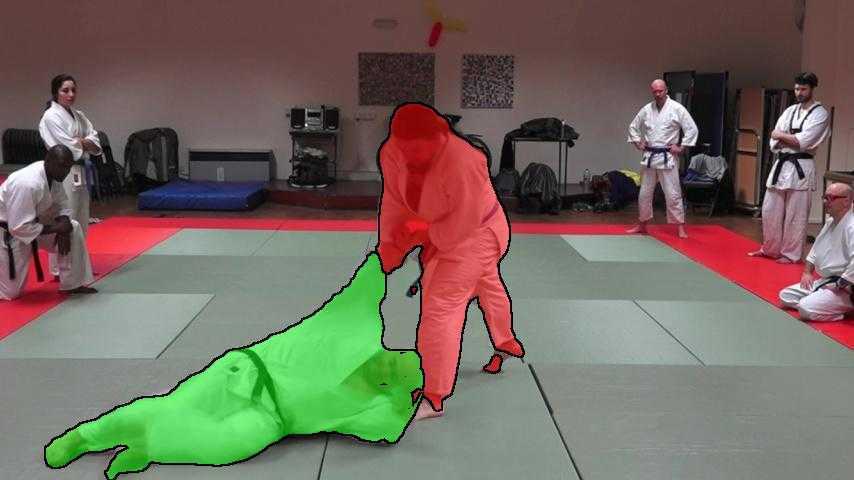} \tabularnewline 
\multicolumn{6}{c}{\textit{\vspace{0.05em} }} \tabularnewline

\multicolumn{6}{c}{\textit{\textcolor{mygreen}{ID 1}: "A boy wearing a white t-shirt". \textcolor{myred}{ID 2}: "A red bmx bike".} } \tabularnewline
\includegraphics[width=0.16\linewidth]{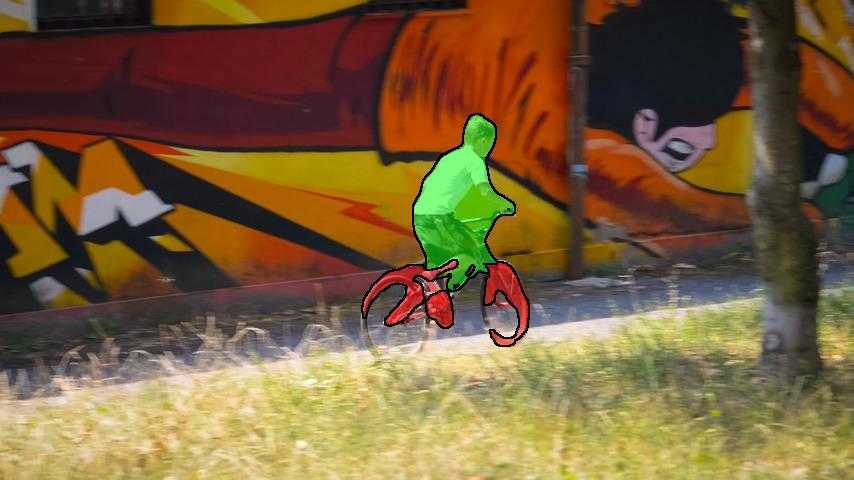} & {\footnotesize{}}
\includegraphics[width=0.16\linewidth]{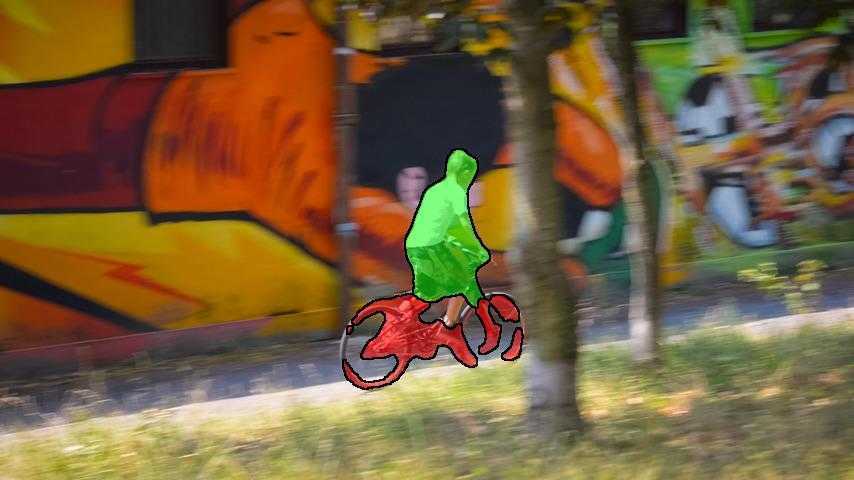} & {\footnotesize{}}
\includegraphics[width=0.16\linewidth]{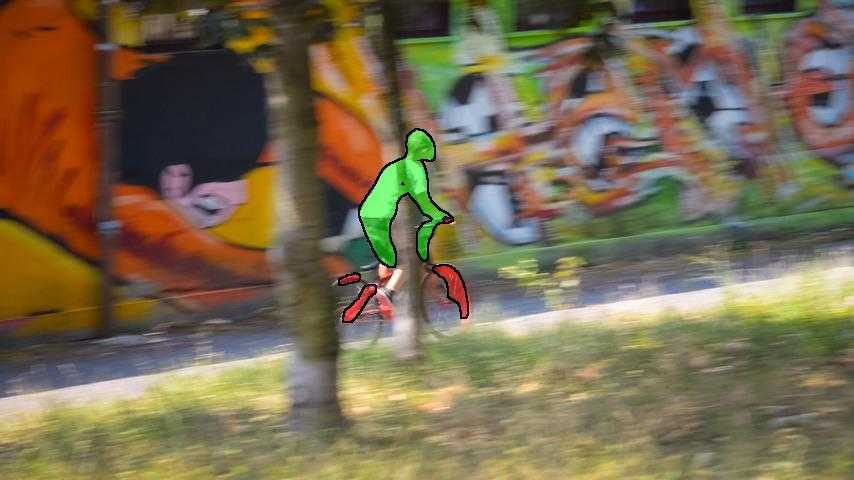} & {\footnotesize{}}
\includegraphics[width=0.16\linewidth]{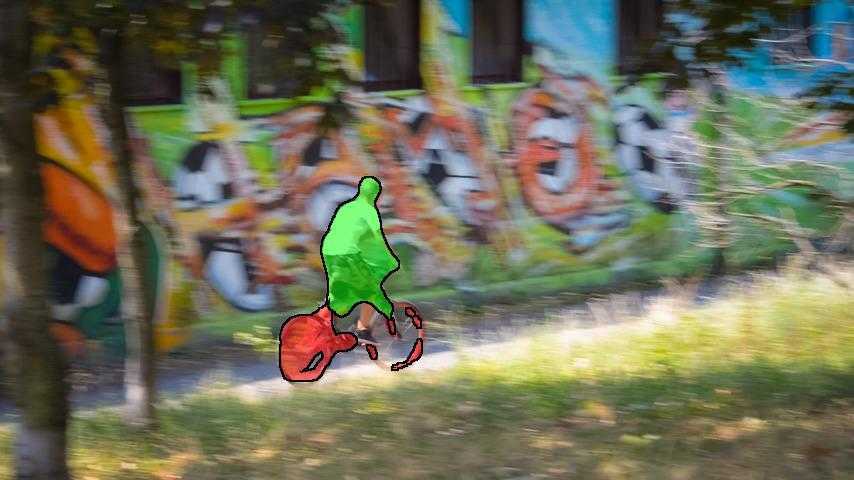} & {\footnotesize{}}
\includegraphics[width=0.16\linewidth]{figures/VOS_qual/bmx_trees_17/00047} & {\footnotesize{}}
\includegraphics[width=0.16\linewidth]{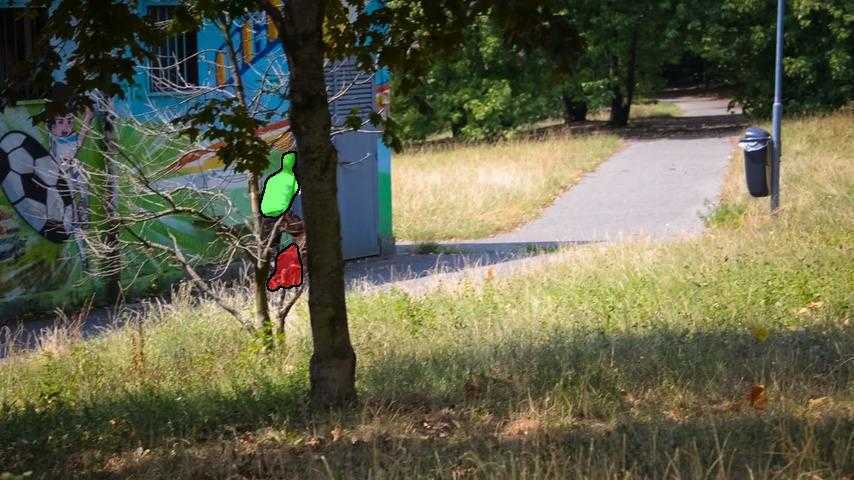}\tabularnewline
\multicolumn{6}{c}{\textit{\vspace{0.05em} }} \tabularnewline

\multicolumn{6}{c}{\textit{  \textcolor{mygreen}{ID 1}: "A green motorbike". \textcolor{myred}{  ID 2}: "A man riding a motorbike".}} \tabularnewline
\includegraphics[width=0.16\linewidth]{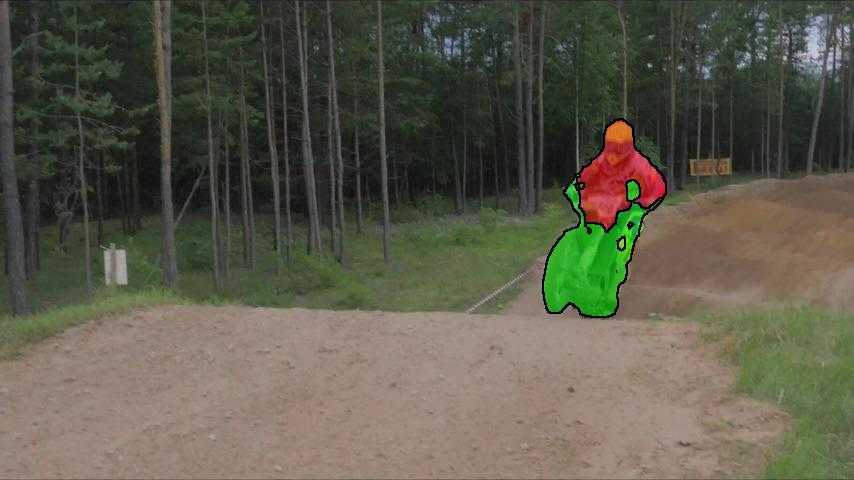} & {\footnotesize{}}
\includegraphics[width=0.16\linewidth]{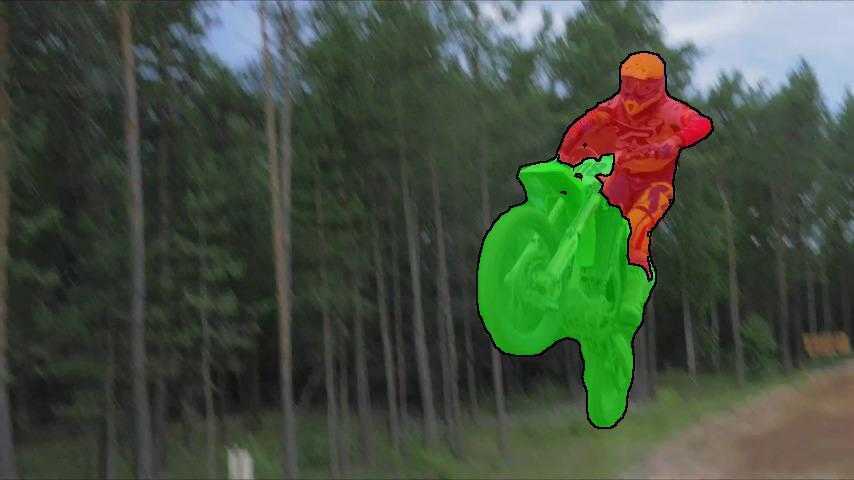} & {\footnotesize{}}
\includegraphics[width=0.16\linewidth]{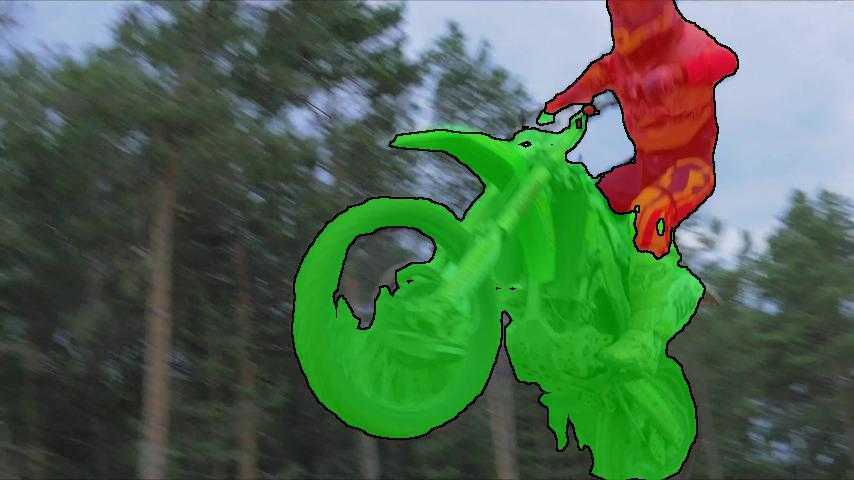} & {\footnotesize{}}
\includegraphics[width=0.16\linewidth]{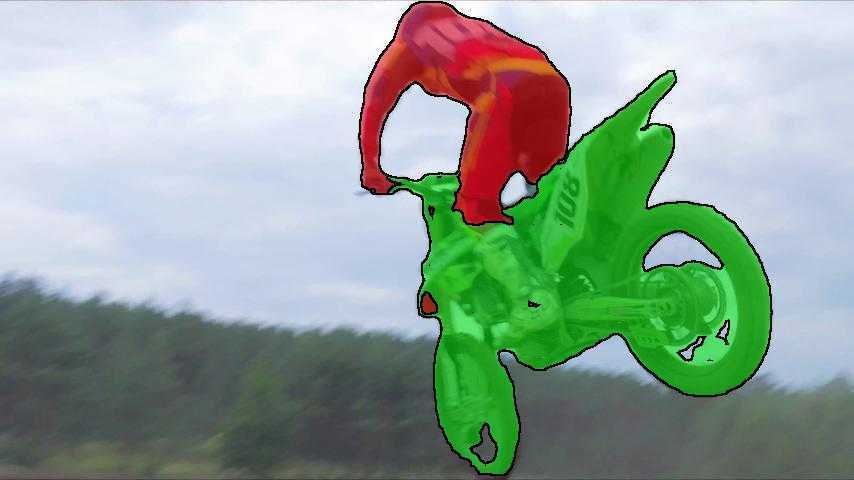} & {\footnotesize{}}
\includegraphics[width=0.16\linewidth]{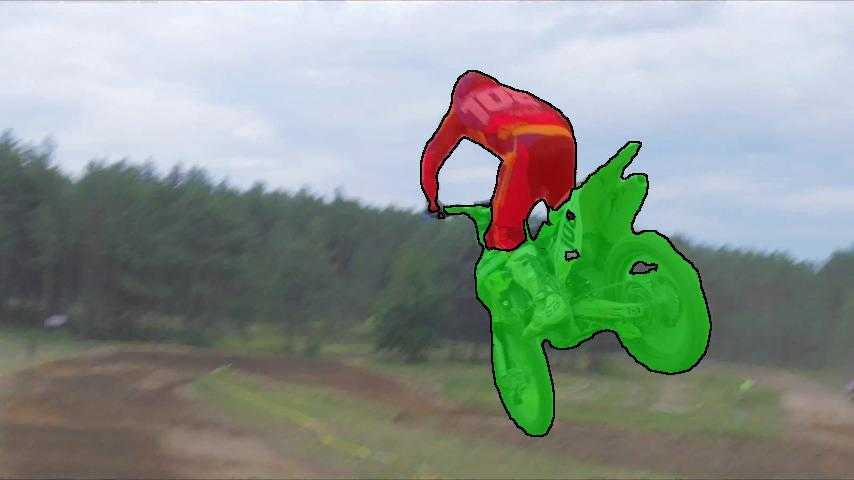} & {\footnotesize{}}
\includegraphics[width=0.16\linewidth]{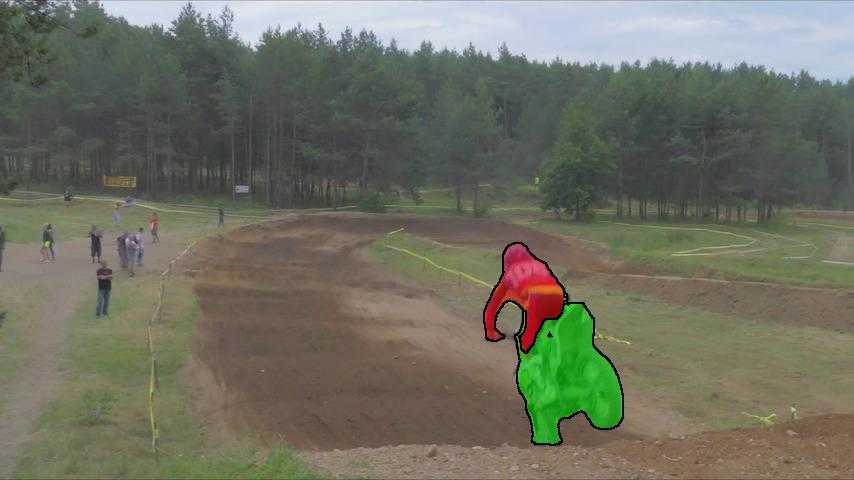}\tabularnewline

\end{tabular}\hfill{}
\par\end{centering}

\caption{\label{fig:sup_qualitative-results-vos}Video object segmentation qualitative results using only Language as supervision
on $\text{DAVIS}_{\text{16}}$ and $\text{DAVIS}_{\text{17}}$, val sets. Frames sampled along
the video duration. 
 }
\vspace{0em}
\end{figure*}

\subsection{\label{sec:sup_voc-qualitative} Qualitative results for video object segmentation}

Figure \ref{fig:sup_qualitative-results-vos} provides more qualitative examples of Language-only supervision for video object segmentation on $\text{DAVIS}_{\text{16}}$ and $\text{DAVIS}_{\text{17}}$, validation sets.
We observe successful handling of shape deformations, fast motion as well as partial and full occlusions.

\begin{figure*}[t!]
\begin{centering}
\setlength{\tabcolsep}{0.1em}
\renewcommand{\arraystretch}{0}
\par\end{centering}
\begin{centering}
\hfill{}%
\begin{tabular}{c@{\hskip 0.05in}c@{\hskip 0.05in}c@{\hskip 0.05in}c@{\hskip 0.05in}c@{\hskip 0.05in}c}

\multicolumn{6}{c}{\textit{ \textit{\textcolor{mygreen}{ID 1}: "A man wearing a cap". \textcolor{myred}{ID 2}: "A black bike".}}} \tabularnewline
\includegraphics[width=0.15\linewidth]{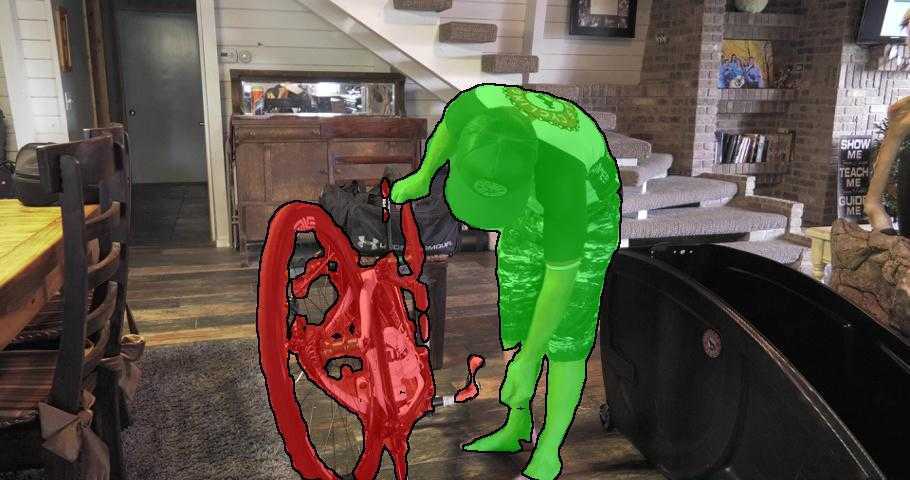} & {\footnotesize{}}
\includegraphics[width=0.15\linewidth]{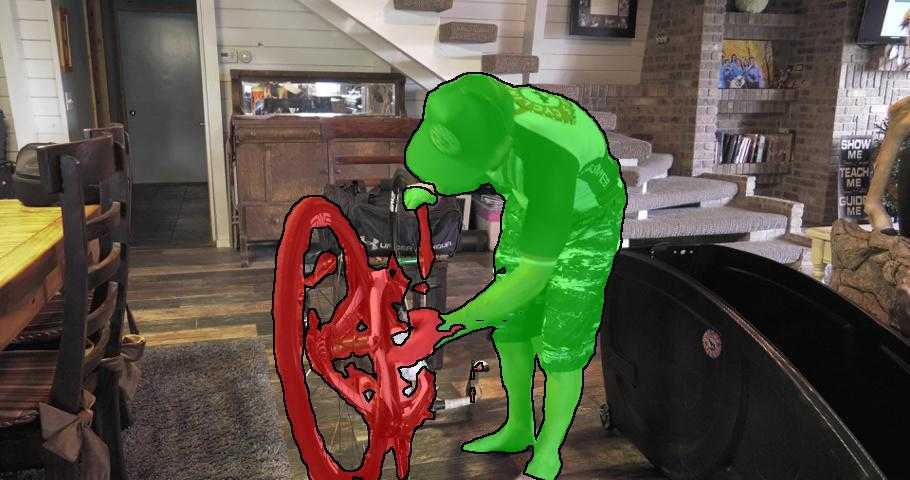} & {\footnotesize{}}
\includegraphics[width=0.15\linewidth]{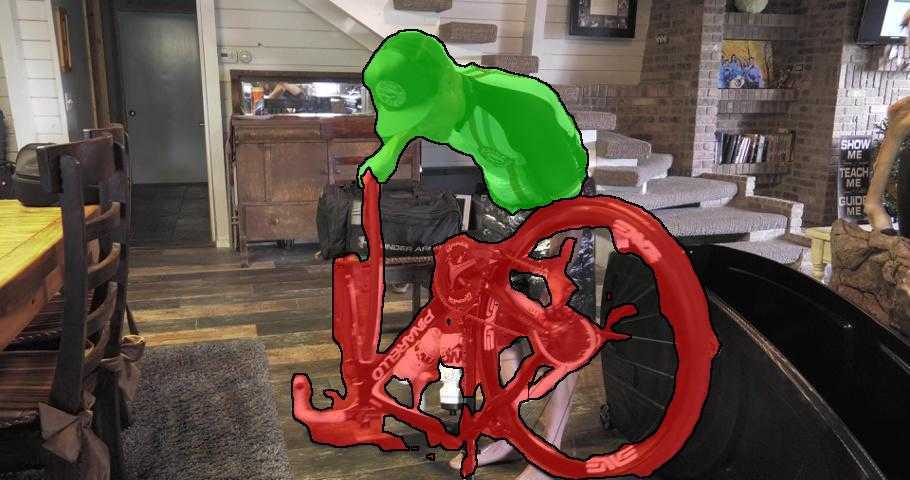} & {\footnotesize{}}
\includegraphics[width=0.15\linewidth]{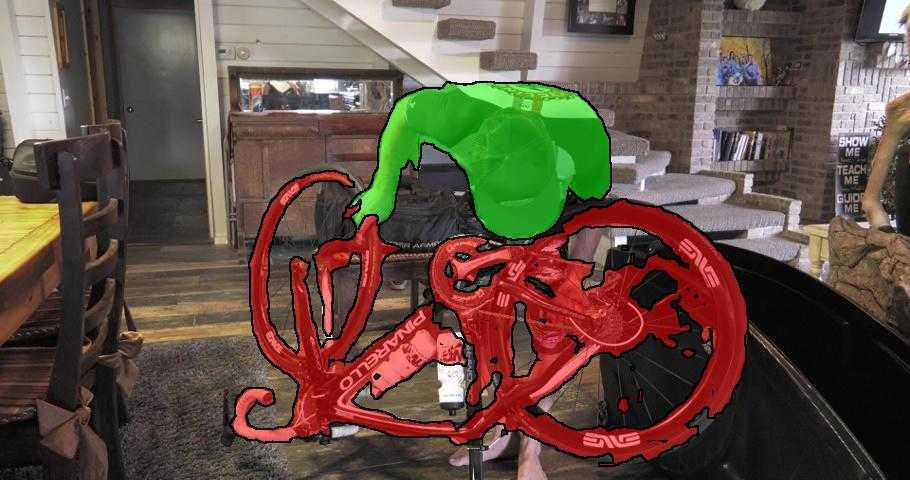} & {\footnotesize{}}
\includegraphics[width=0.15\linewidth]{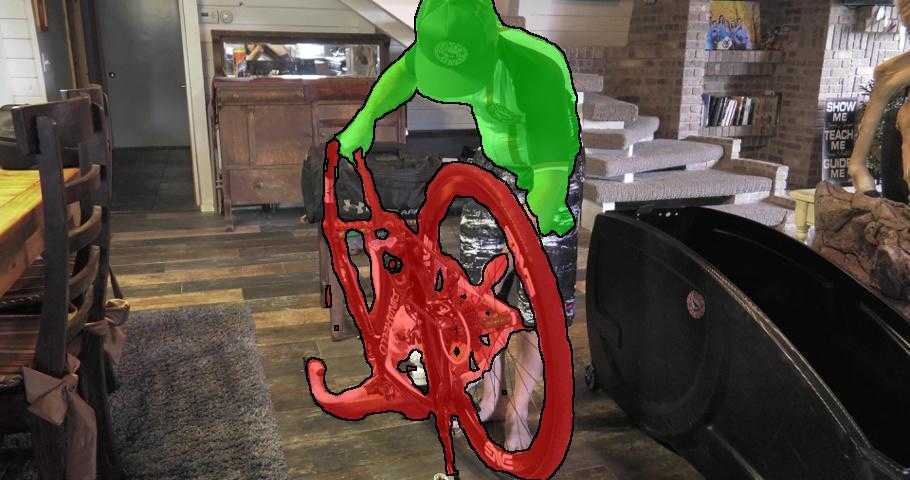} & {\footnotesize{}}
\includegraphics[width=0.15\linewidth]{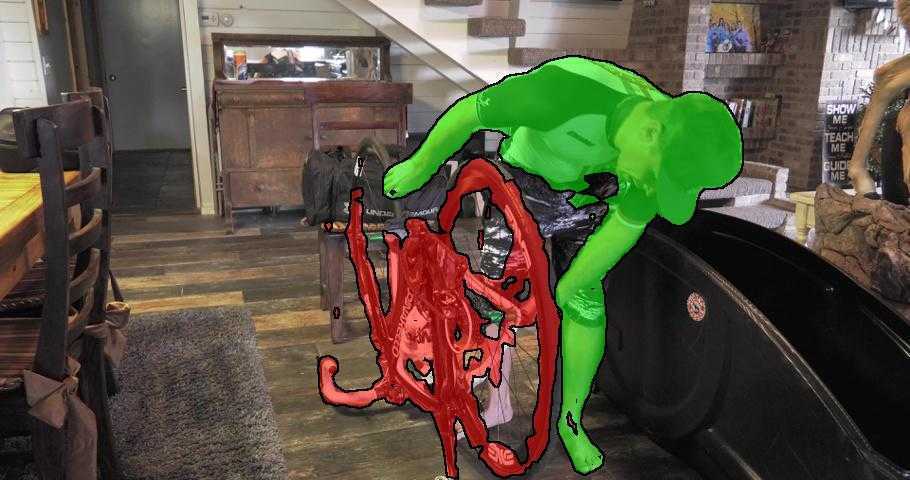}\tabularnewline
\multicolumn{6}{c}{\textit{\vspace{0.05em} }} \tabularnewline

\multicolumn{6}{@{}c@{}}{
\begin{tabular}{@{}c@{}}
 \textit{ \textcolor{mygreen}{ID 1}: "A brown piglet in the middle". \textcolor{myred}{ID 2}: "A brown and white colored piglet".}\tabularnewline
  \textit{  \textcolor{myyellow}{ID 3}: "An adult pig on the right".}\tabularnewline
\end{tabular}
} \tabularnewline
\includegraphics[width=0.15\linewidth]{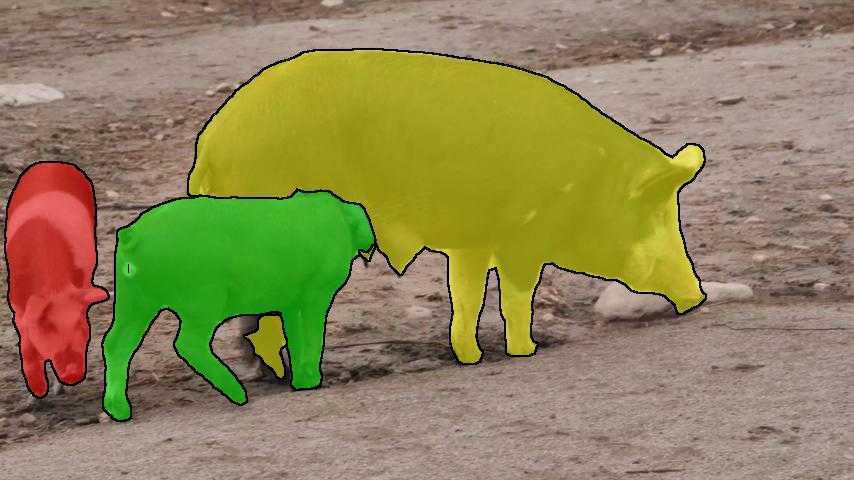} & {\footnotesize{}}
\includegraphics[width=0.15\linewidth]{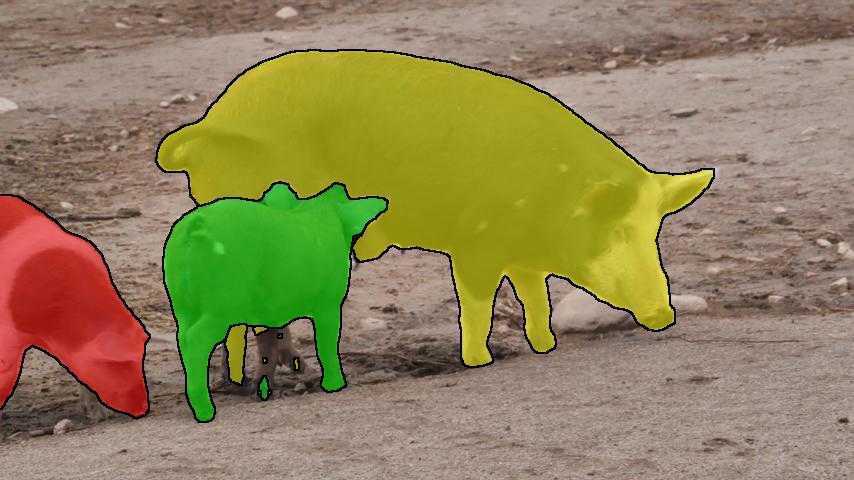} & {\footnotesize{}}
\includegraphics[width=0.15\linewidth]{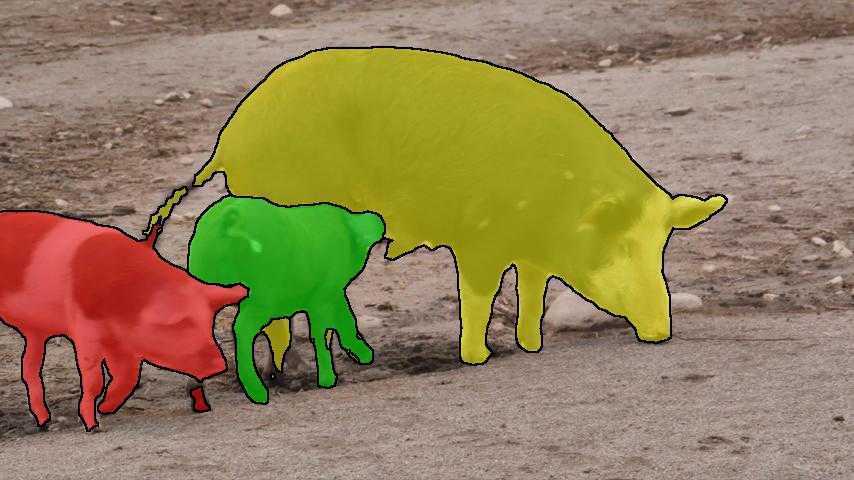} & {\footnotesize{}}
\includegraphics[width=0.15\linewidth]{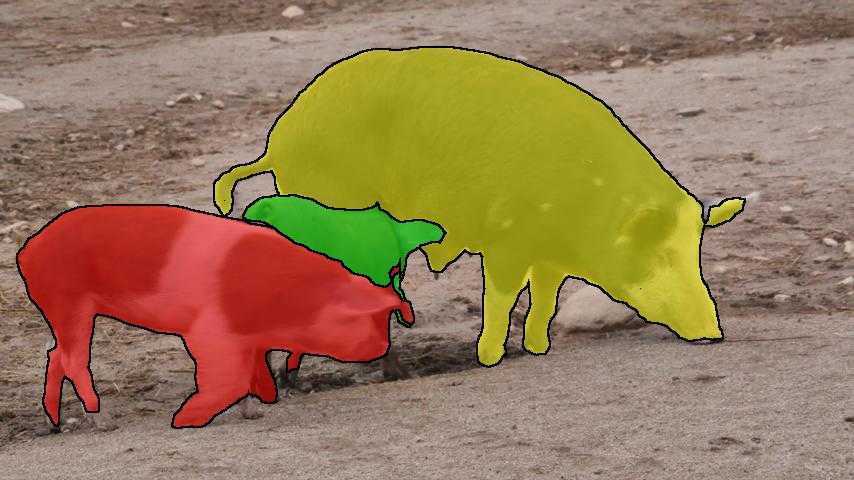} & {\footnotesize{}}
\includegraphics[width=0.15\linewidth]{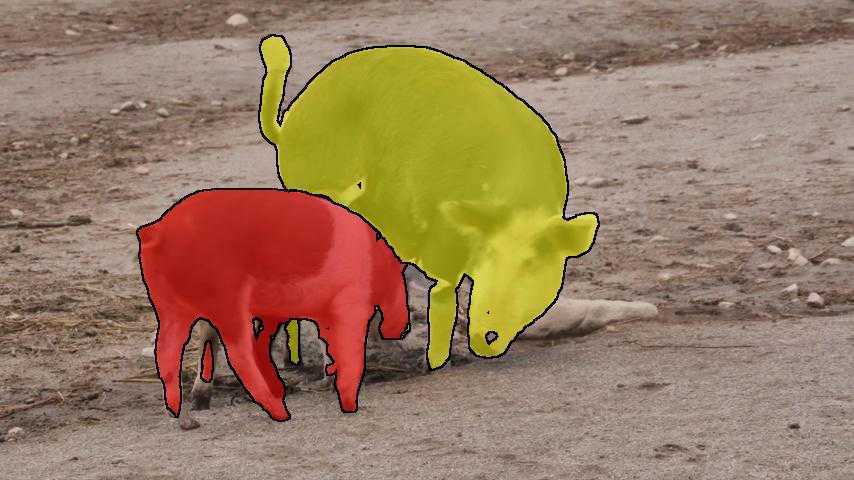} & {\footnotesize{}}
\includegraphics[width=0.15\linewidth]{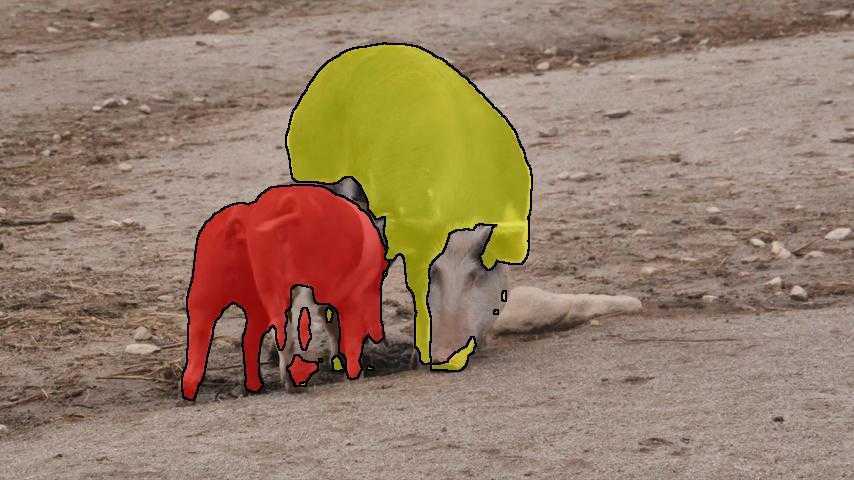}\tabularnewline
\multicolumn{6}{c}{\textit{\vspace{0.05em} }} \tabularnewline

\multicolumn{6}{@{}c@{}}{
\begin{tabular}{@{}c@{}}
\textit{\textcolor{mygreen}{ID 1}: "An orange goldfish in the center next to the largest fish". \textcolor{myred}{ID 2}: "The biggest goldfish".} \tabularnewline
\textit{\textcolor{myyellow}{ID 3}: "The smallest goldfish". \textcolor{myblue}{ID 4}: "A small goldfish in the end". } \tabularnewline
\textit{\textcolor{mypink}{ID 5}: "A goldfish on the bottom".} \tabularnewline
\end{tabular}
} \tabularnewline

\includegraphics[width=0.15\linewidth]{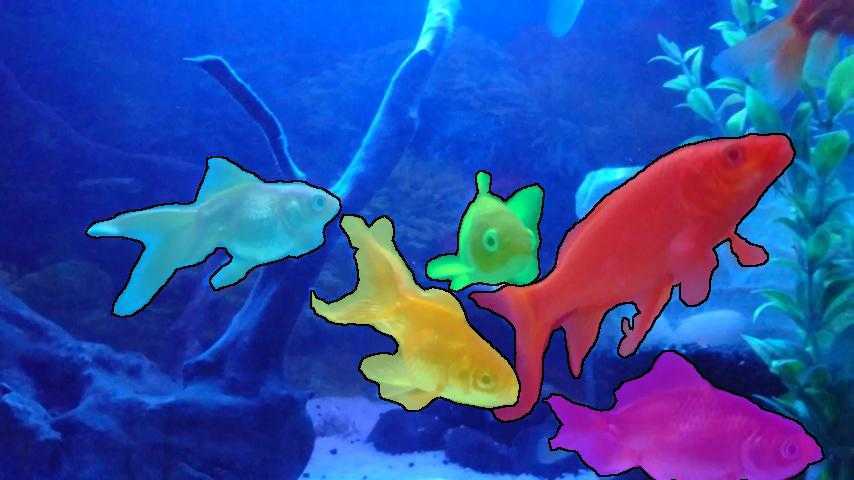} & {\footnotesize{}}
\includegraphics[width=0.15\linewidth]{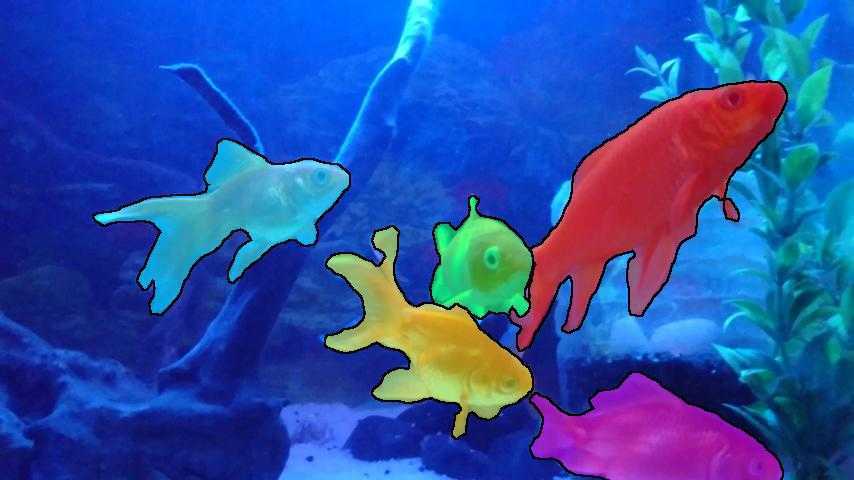} & {\footnotesize{}}
\includegraphics[width=0.15\linewidth]{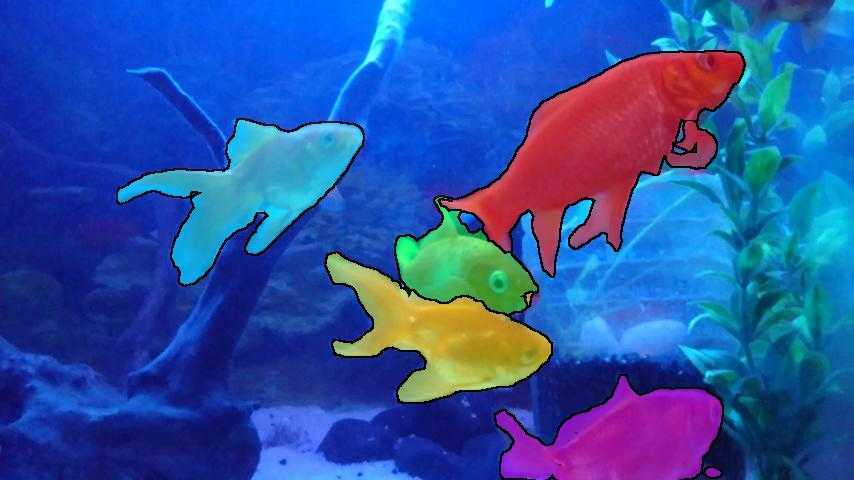} & {\footnotesize{}}
\includegraphics[width=0.15\linewidth]{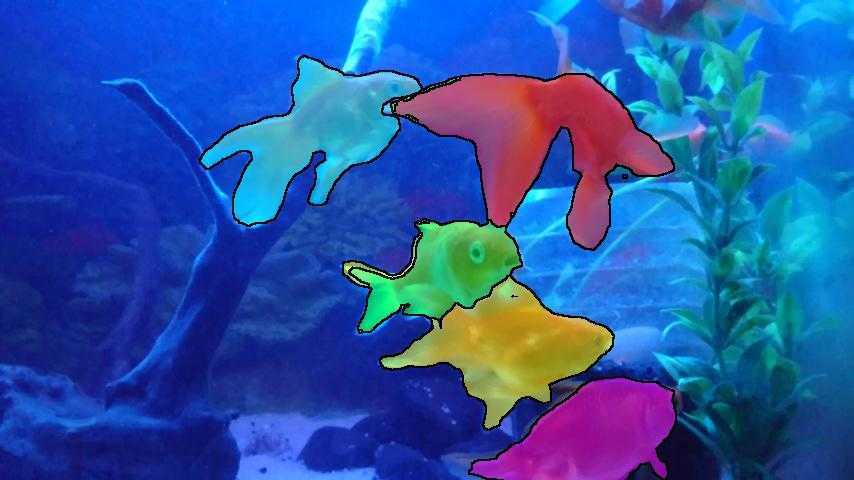} & {\footnotesize{}}
\includegraphics[width=0.15\linewidth]{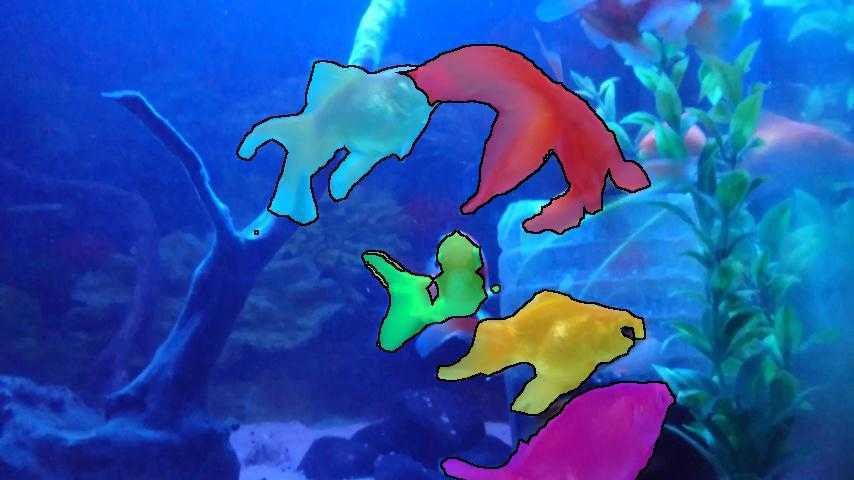} & {\footnotesize{}}
\includegraphics[width=0.15\linewidth]{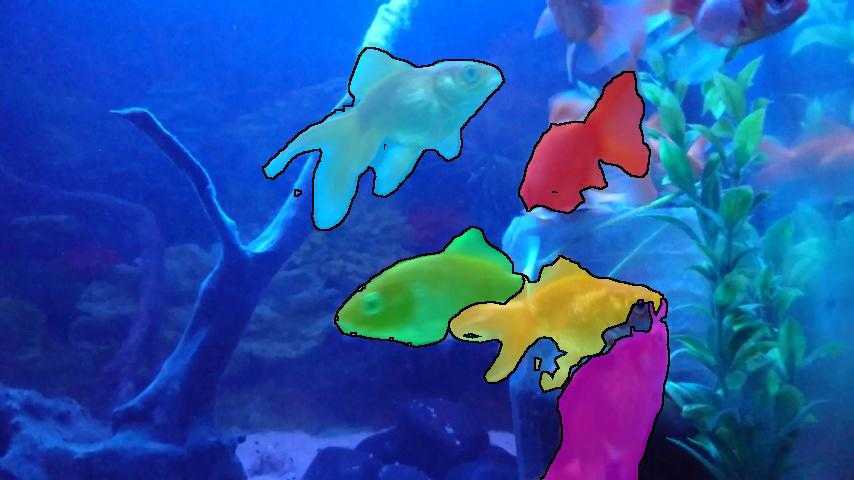}\tabularnewline

\end{tabular}\hfill{}
\par\end{centering}

\caption{\label{fig:sup_qualitative-results-mask-lang-vos}Video object segmentation qualitative results using Mask + Language as supervision
on $\text{DAVIS}_{\text{17}}$, val set. Frames sampled along
the video duration. In the last row we visualize a failure case of the proposed approach.  }
\vspace{0em}
\end{figure*}

Figure \ref{fig:sup_qualitative-results-mask-lang-vos} shows examples of Mask + Language supervision on $\text{DAVIS}_{\text{17}}$, validation set.
We observe high quality instance level segmentation of multiple similar looking objects. 
% In the last row of Figure \ref{fig:sup_qualitative-results-mask-lang-vos} we 
% visualize a failure case of the proposed approach.

\begin{figure*}
\begin{centering}
\setlength{\tabcolsep}{0.1em}
\renewcommand{\arraystretch}{0}
\par\end{centering}
\begin{centering}
\hfill{}%
\begin{tabular}{c@{\hskip 0.05in}c@{\hskip 0.05in}c@{\hskip 0.05in}c@{\hskip 0.05in}c@{\hskip 0.05in}c}

\multicolumn{6}{c}{Language supervision, \textit{ \textcolor{mygreen}{ID 1}: "A brown camel in the front".} } \tabularnewline
\includegraphics[width=0.15\linewidth]{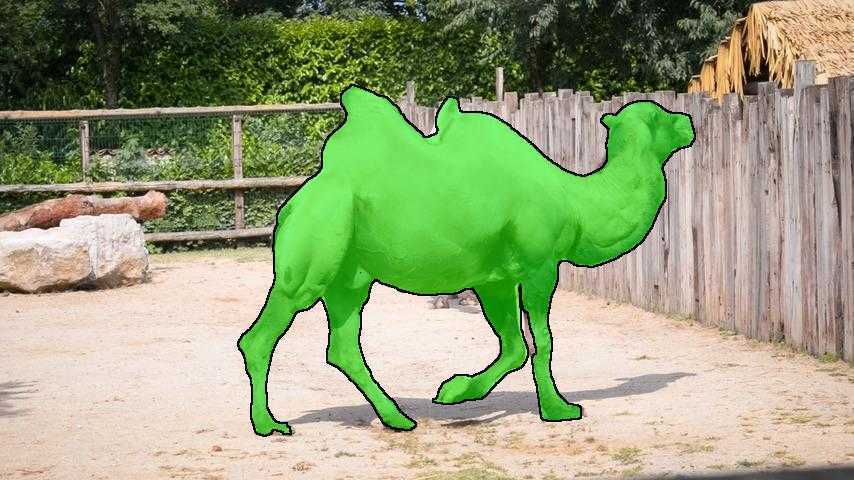} & {\footnotesize{}}
\includegraphics[width=0.15\linewidth]{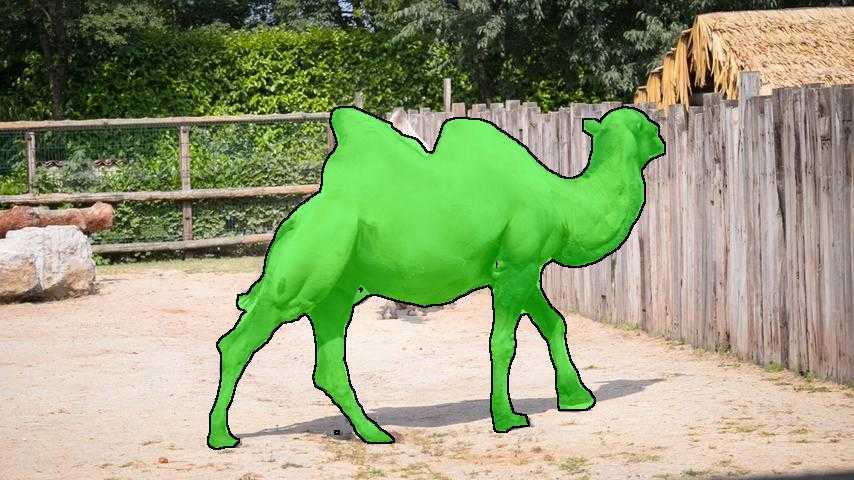} & {\footnotesize{}}
\includegraphics[width=0.15\linewidth]{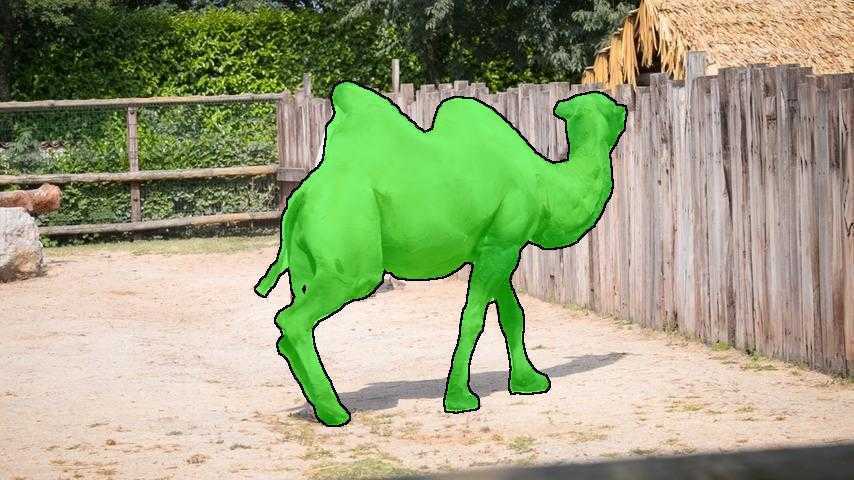} & {\footnotesize{}}
\includegraphics[width=0.15\linewidth]{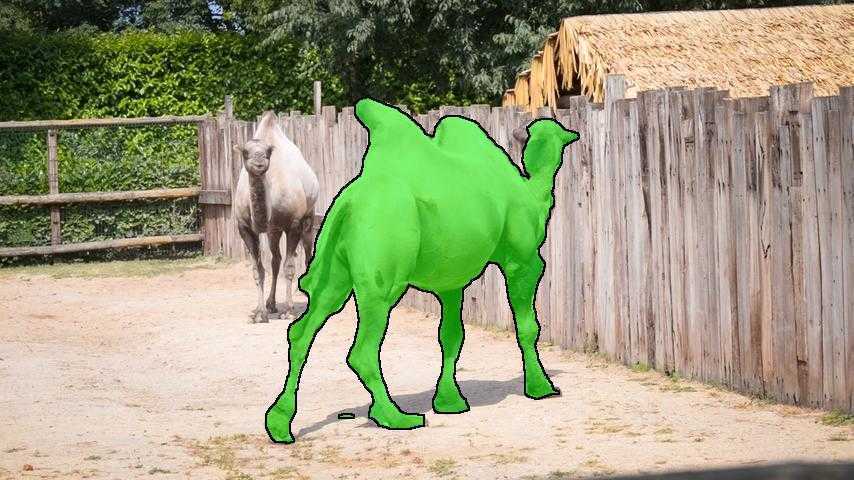} & {\footnotesize{}}
\includegraphics[width=0.15\linewidth]{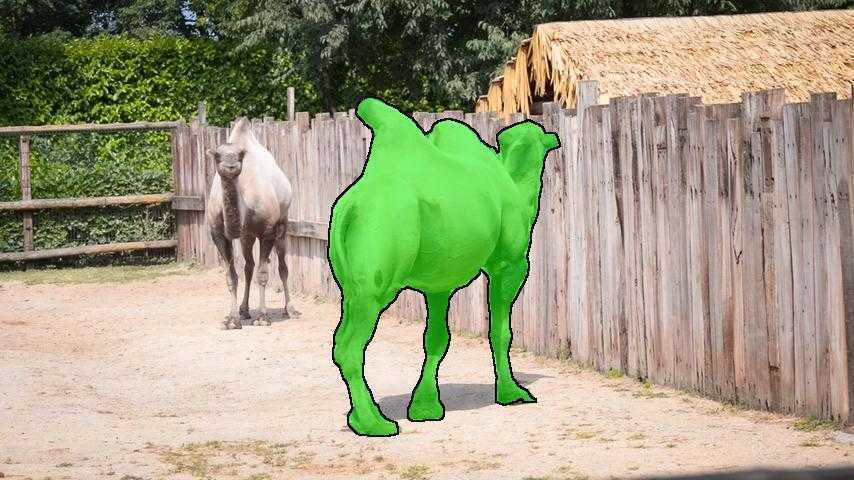} & {\footnotesize{}}
\includegraphics[width=0.15\linewidth]{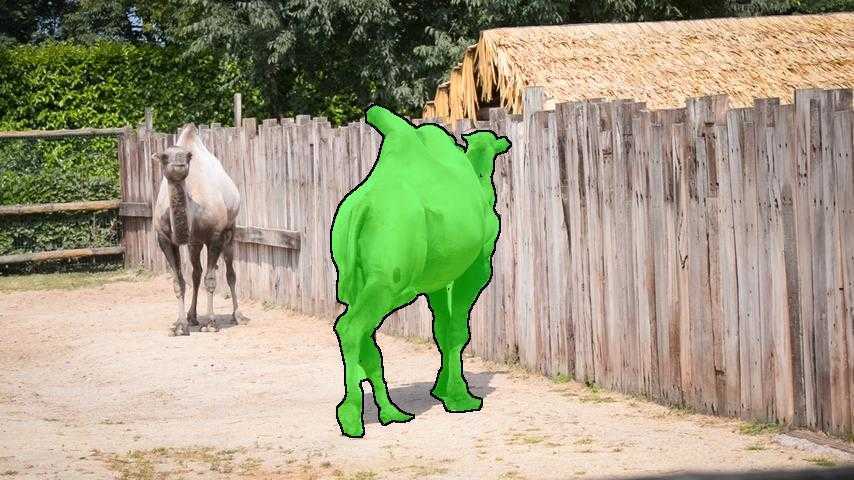}\tabularnewline
\multicolumn{6}{c}{{Pixel-level mask supervision } }\tabularnewline
\includegraphics[width=0.15\linewidth]{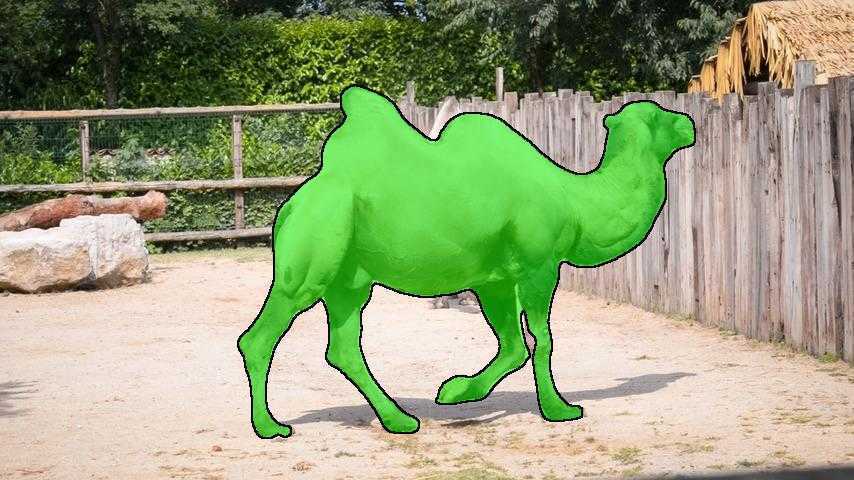} & {\footnotesize{}}
\includegraphics[width=0.15\linewidth]{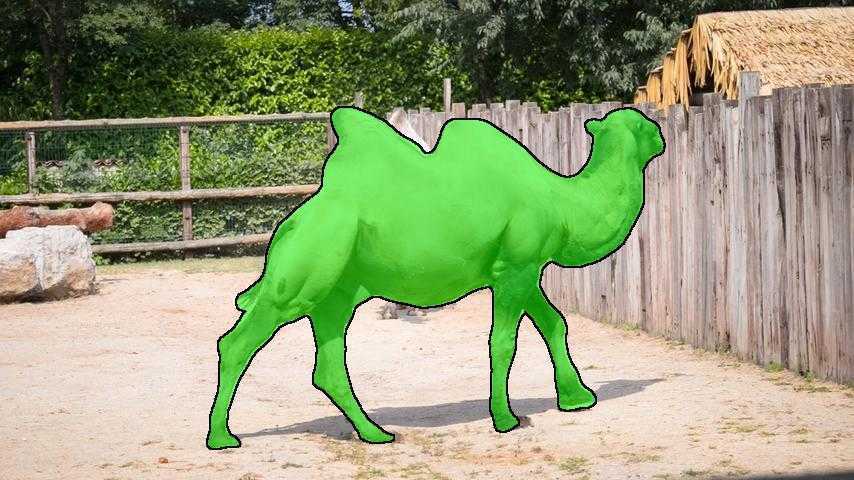} & {\footnotesize{}}
\includegraphics[width=0.15\linewidth]{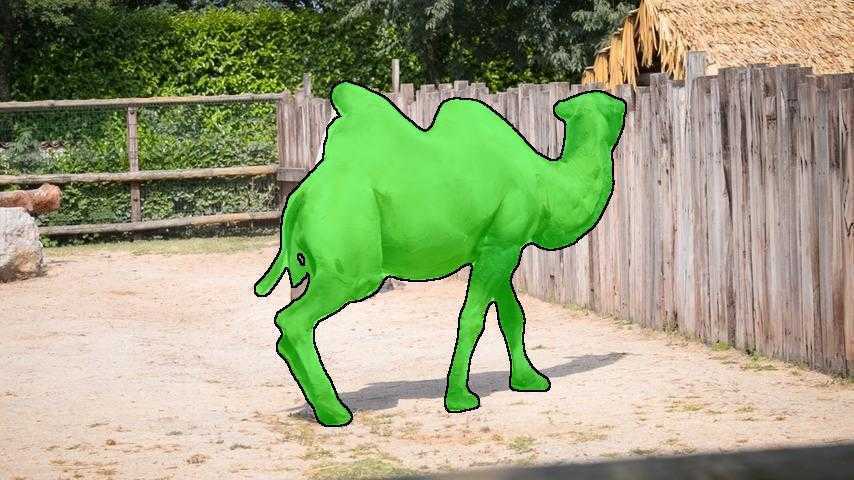} & {\footnotesize{}}
\includegraphics[width=0.15\linewidth]{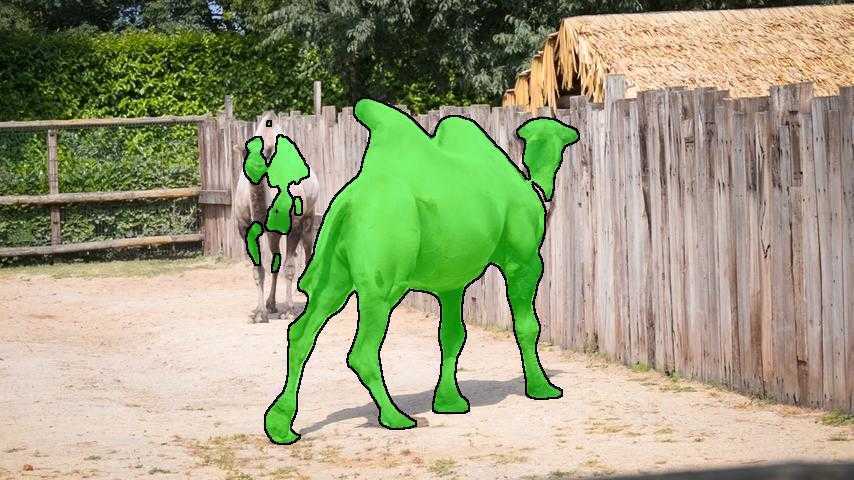} & {\footnotesize{}}
\includegraphics[width=0.15\linewidth]{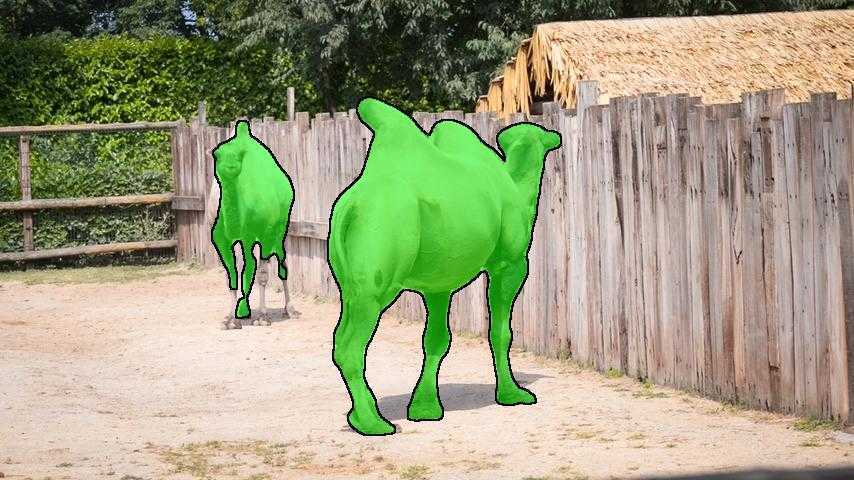} & {\footnotesize{}}
\includegraphics[width=0.15\linewidth]{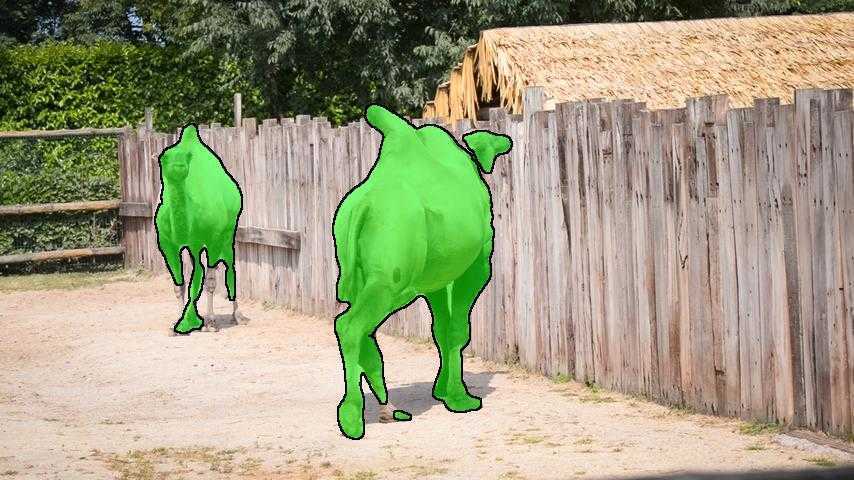}\tabularnewline

\multicolumn{6}{c}{\textit{\vspace{0.05em} }} \tabularnewline

\multicolumn{6}{c}{Language supervision, \textit{ \textcolor{mygreen}{ID 1}: "A silver car".} } \tabularnewline
\includegraphics[width=0.15\linewidth]{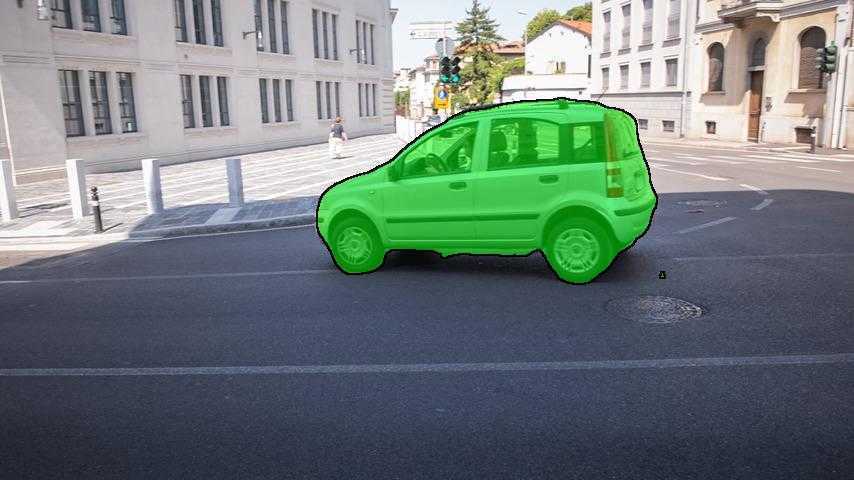} & {\footnotesize{}}
\includegraphics[width=0.15\linewidth]{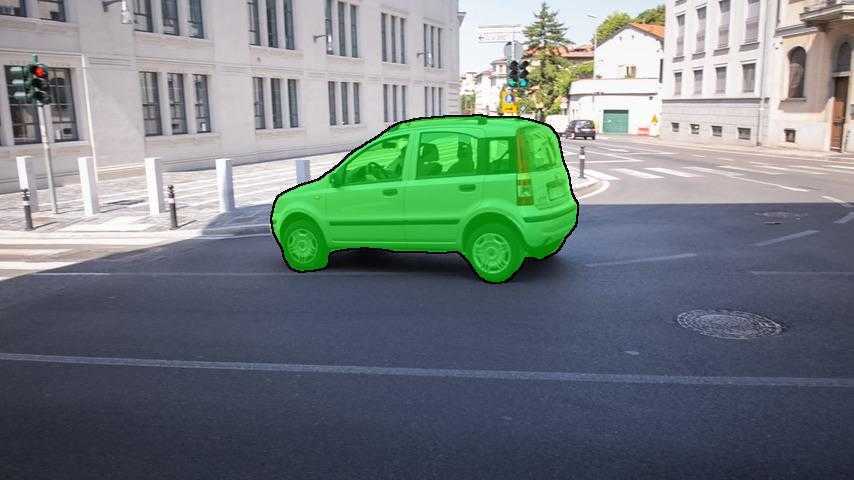} & {\footnotesize{}}
\includegraphics[width=0.15\linewidth]{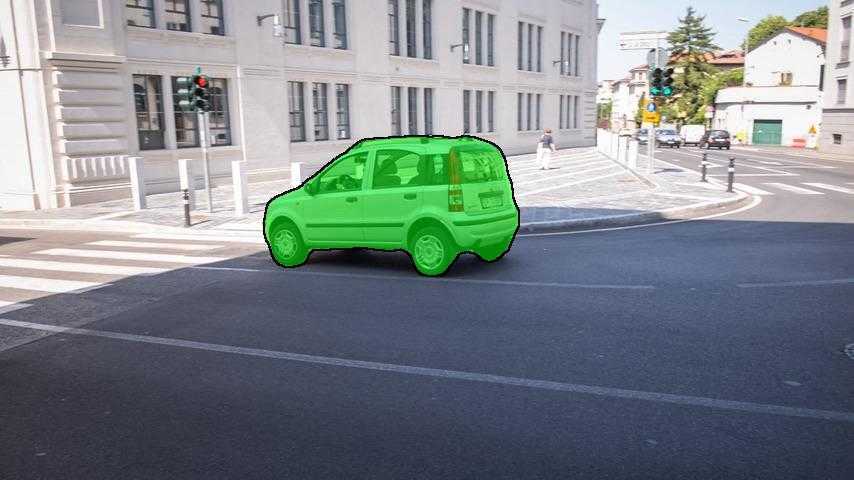} & {\footnotesize{}}
\includegraphics[width=0.15\linewidth]{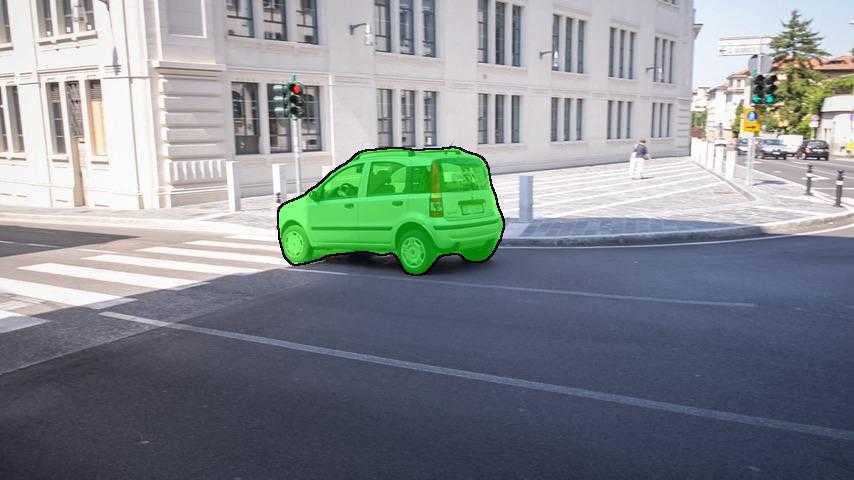} & {\footnotesize{}}
\includegraphics[width=0.15\linewidth]{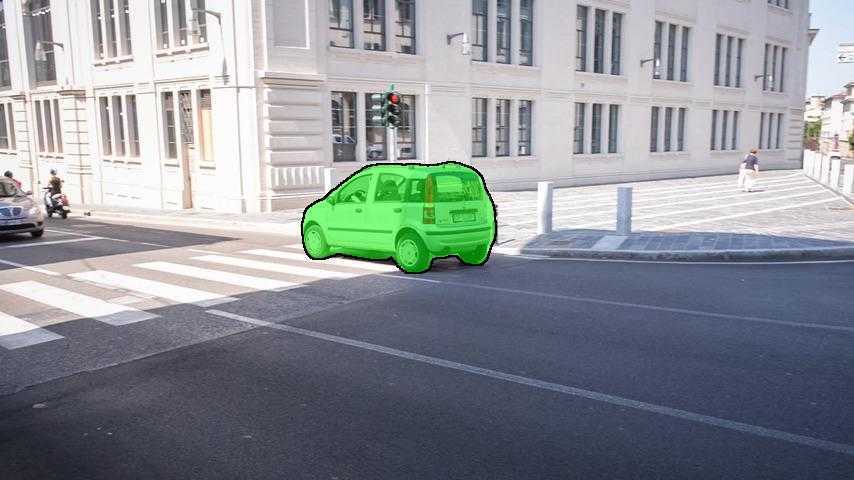} & {\footnotesize{}}
\includegraphics[width=0.15\linewidth]{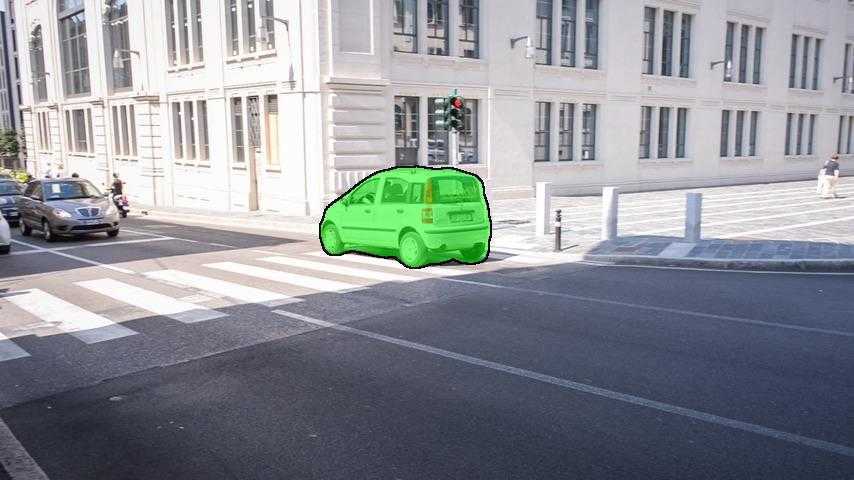}\tabularnewline
\multicolumn{6}{c}{{Pixel-level mask supervision } }\tabularnewline
\includegraphics[width=0.15\linewidth]{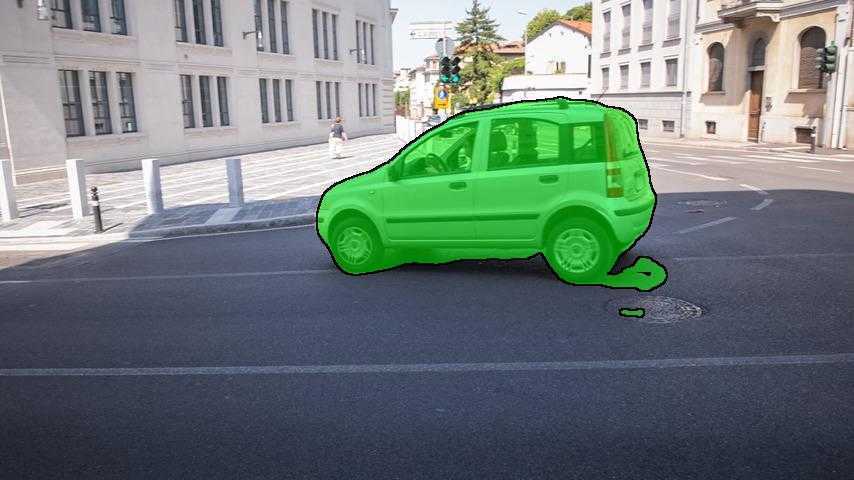} & {\footnotesize{}}
\includegraphics[width=0.15\linewidth]{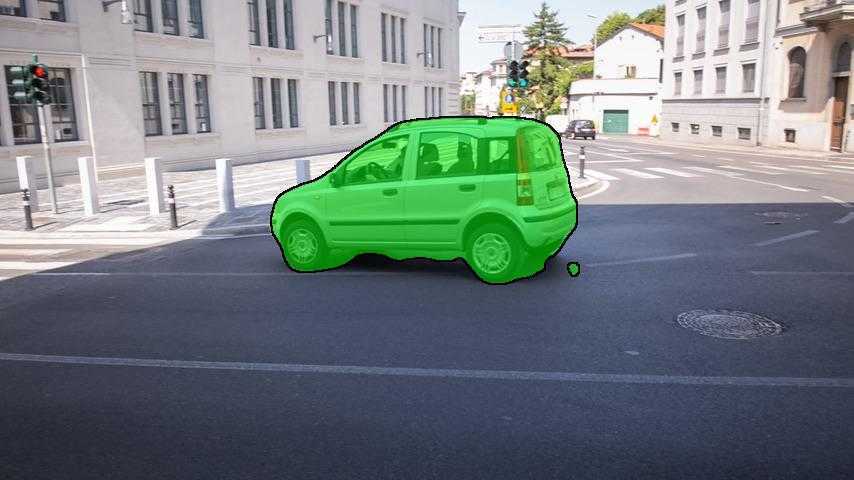} & {\footnotesize{}}
\includegraphics[width=0.15\linewidth]{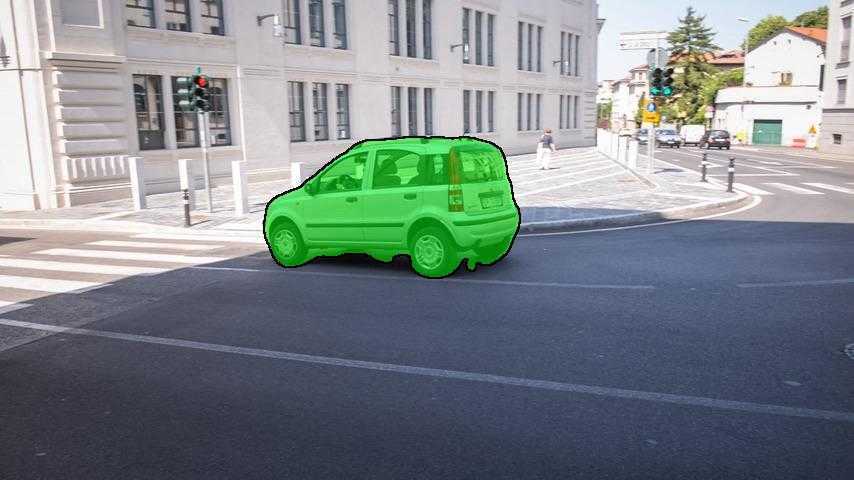} & {\footnotesize{}}
\includegraphics[width=0.15\linewidth]{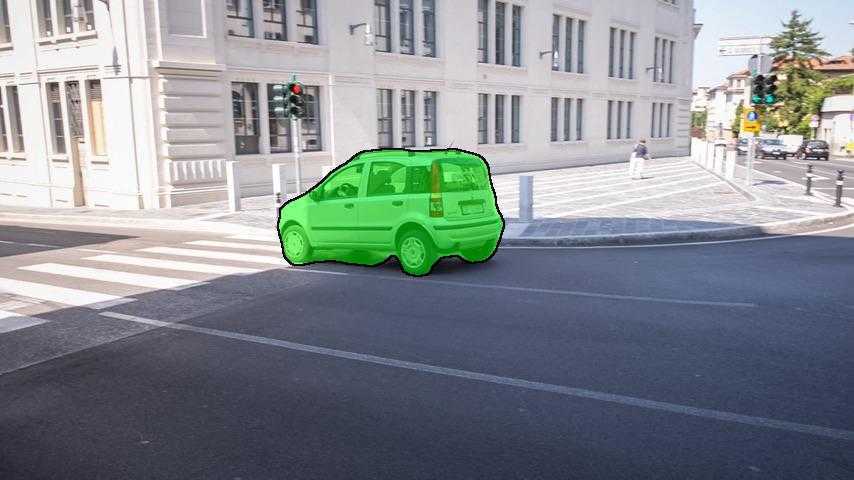} & {\footnotesize{}}
\includegraphics[width=0.15\linewidth]{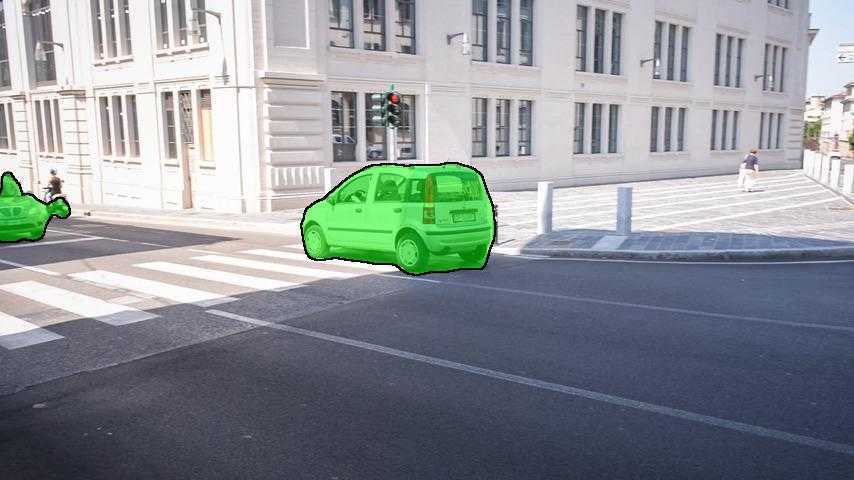} & {\footnotesize{}}
\includegraphics[width=0.15\linewidth]{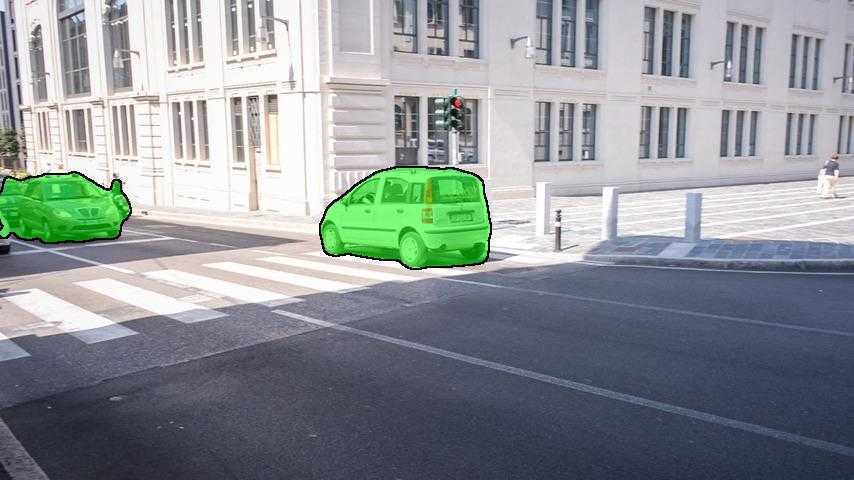}\tabularnewline

\multicolumn{6}{c}{\textit{\vspace{0.05em} }} \tabularnewline

\multicolumn{6}{c}{Language supervision, \textit{ \textcolor{mygreen}{ID 1}: "A black car".} } \tabularnewline
\includegraphics[width=0.15\linewidth]{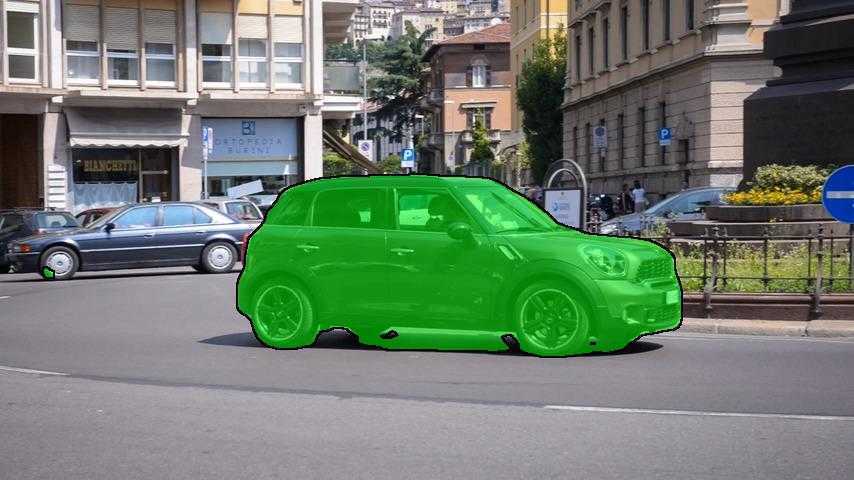} & {\footnotesize{}}
\includegraphics[width=0.15\linewidth]{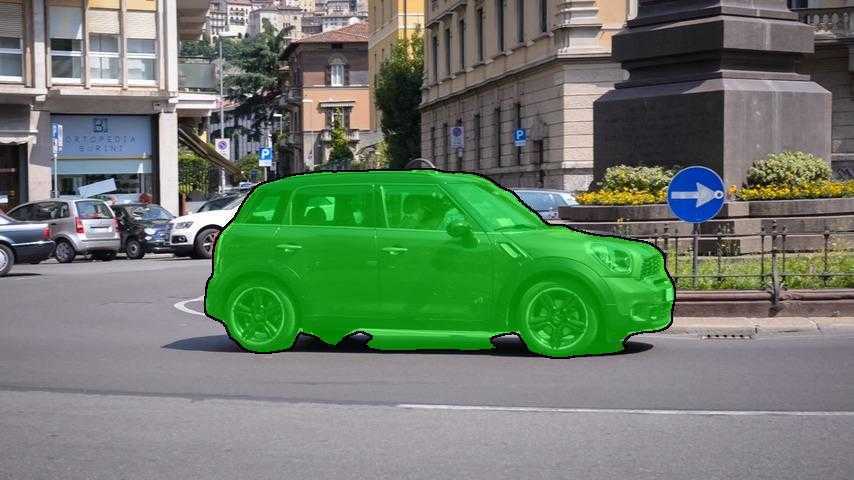} & {\footnotesize{}}
\includegraphics[width=0.15\linewidth]{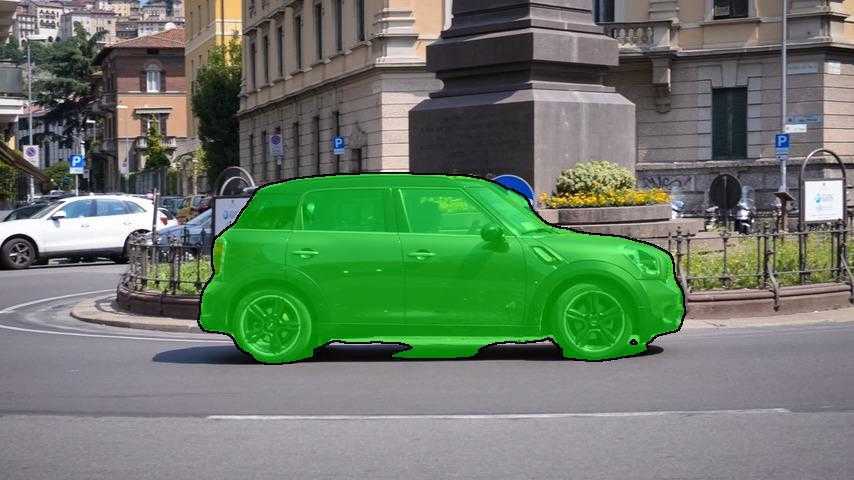} & {\footnotesize{}}
\includegraphics[width=0.15\linewidth]{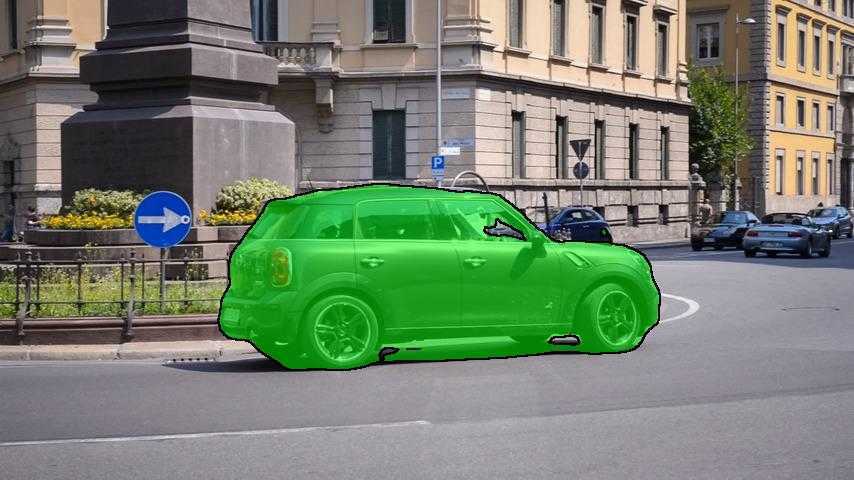} & {\footnotesize{}}
\includegraphics[width=0.15\linewidth]{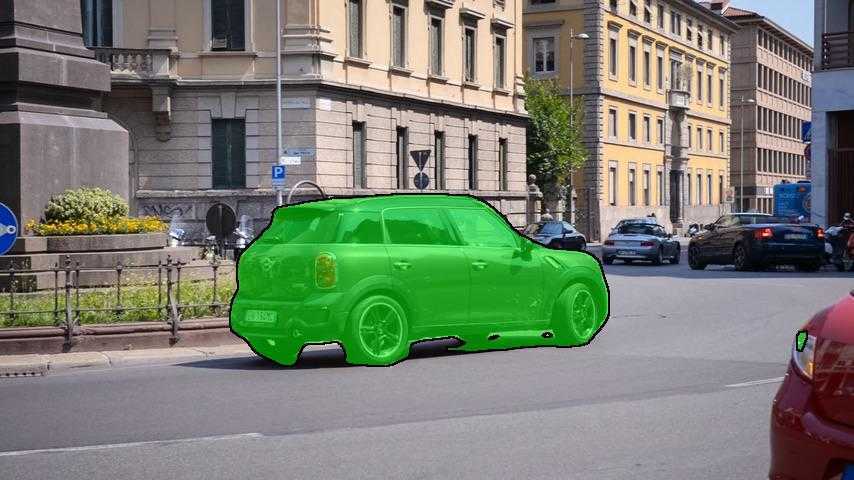} & {\footnotesize{}}
\includegraphics[width=0.15\linewidth]{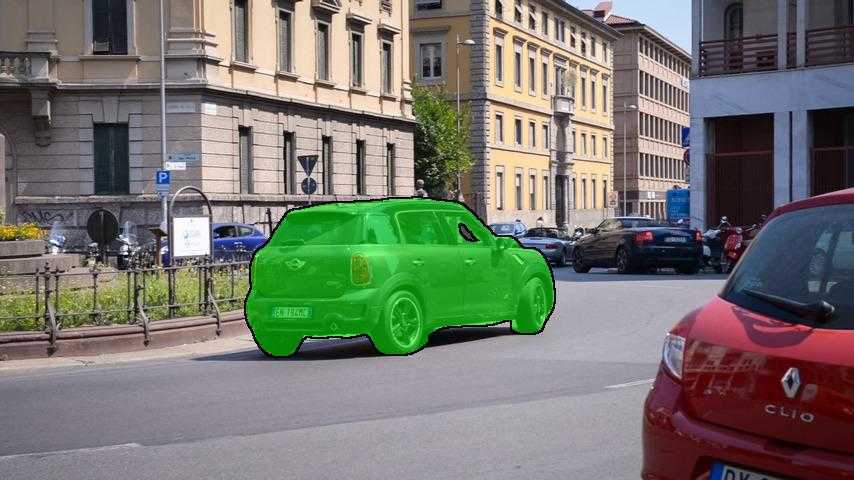}\tabularnewline
\multicolumn{6}{c}{{Pixel-level mask supervision } }\tabularnewline
\includegraphics[width=0.15\linewidth]{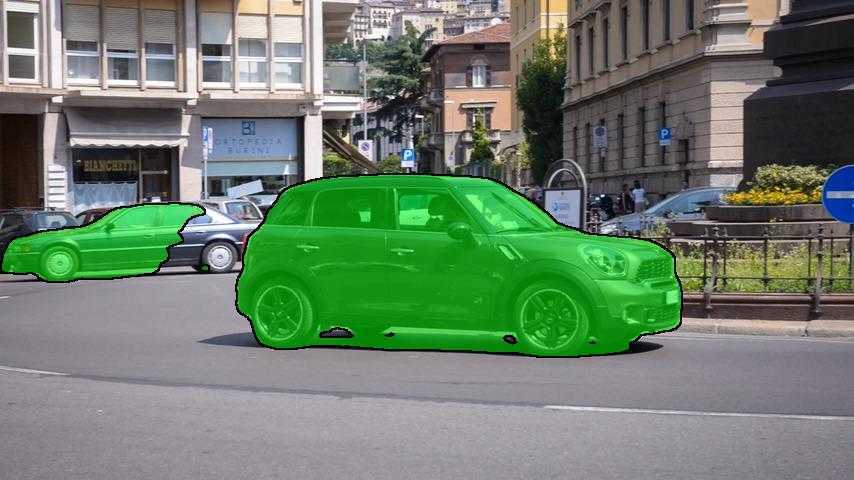} & {\footnotesize{}}
\includegraphics[width=0.15\linewidth]{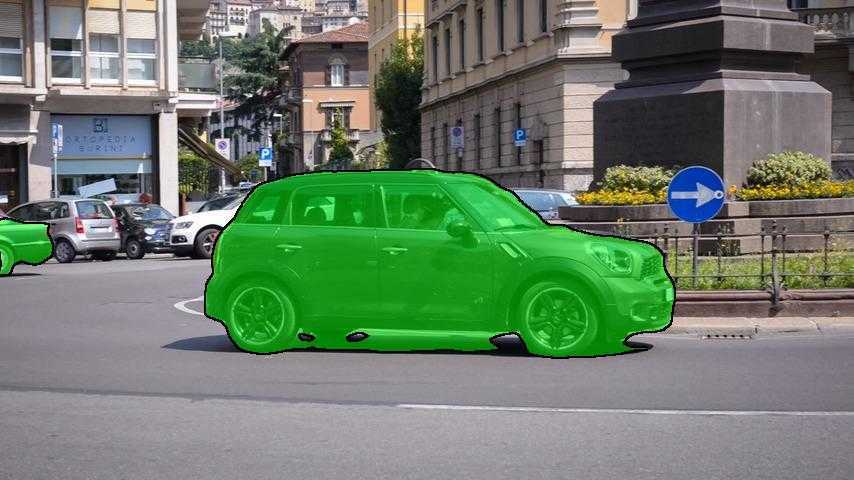} & {\footnotesize{}}
\includegraphics[width=0.15\linewidth]{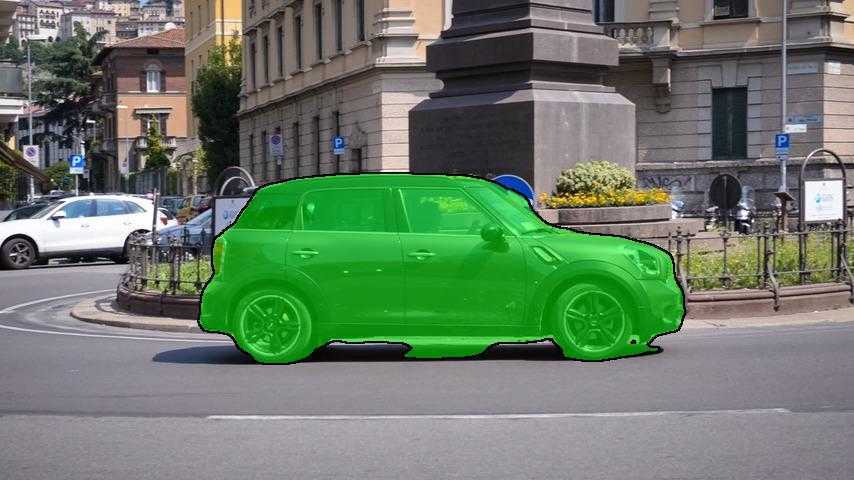} & {\footnotesize{}}
\includegraphics[width=0.15\linewidth]{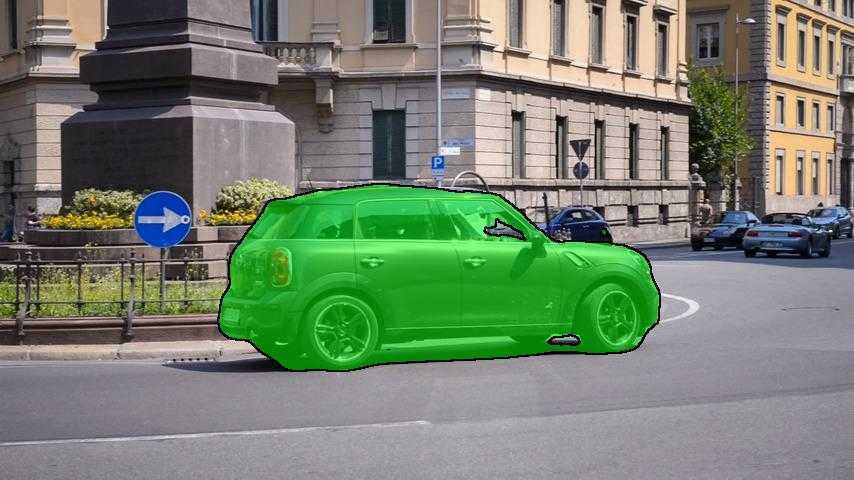} & {\footnotesize{}}
\includegraphics[width=0.15\linewidth]{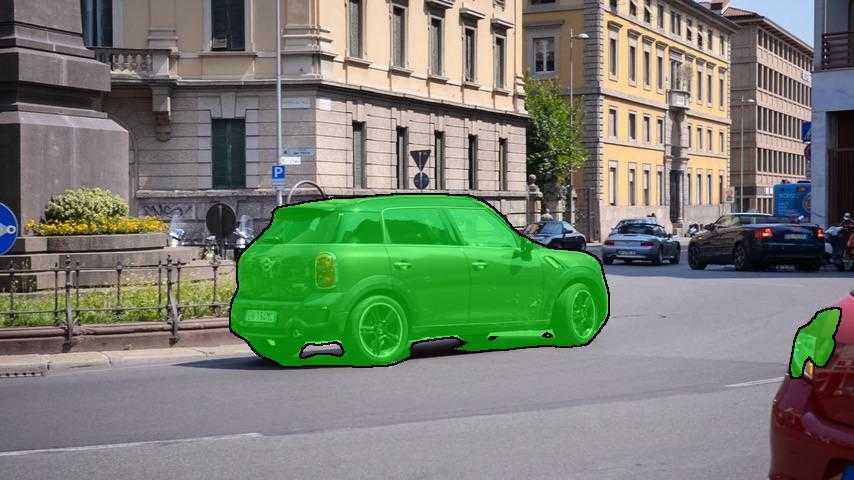} & {\footnotesize{}}
\includegraphics[width=0.15\linewidth]{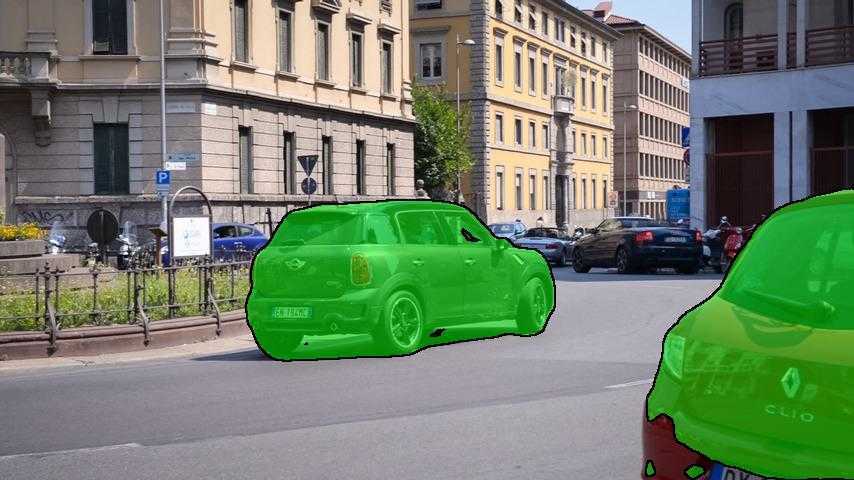}\tabularnewline

\multicolumn{6}{c}{\textit{\vspace{0.05em} }} \tabularnewline

\multicolumn{6}{c}{Language supervision, \textit{  \textcolor{mygreen}{ID 1}: "A green motorbike". \textcolor{myred}{  ID 2}: "A man riding a motorbike".}} \tabularnewline
\includegraphics[width=0.15\linewidth]{figures/VOS_qual/motocross-17/00001} & {\footnotesize{}}
\includegraphics[width=0.15\linewidth]{figures/VOS_qual/motocross-17/00007} & {\footnotesize{}}
\includegraphics[width=0.15\linewidth]{figures/VOS_qual/motocross-17/00012} & {\footnotesize{}}
\includegraphics[width=0.15\linewidth]{figures/VOS_qual/motocross-17/00027} & {\footnotesize{}}
\includegraphics[width=0.15\linewidth]{figures/VOS_qual/motocross-17/00030} & {\footnotesize{}}
\includegraphics[width=0.15\linewidth]{figures/VOS_qual/motocross-17/00038}\tabularnewline

\multicolumn{6}{c}{{Pixel-level mask supervision }} \tabularnewline
\includegraphics[width=0.15\linewidth]{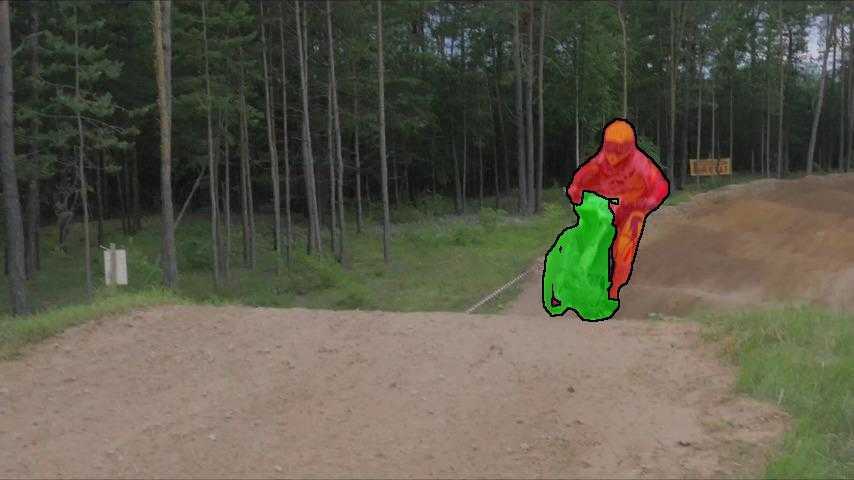} & {\footnotesize{}}
\includegraphics[width=0.15\linewidth]{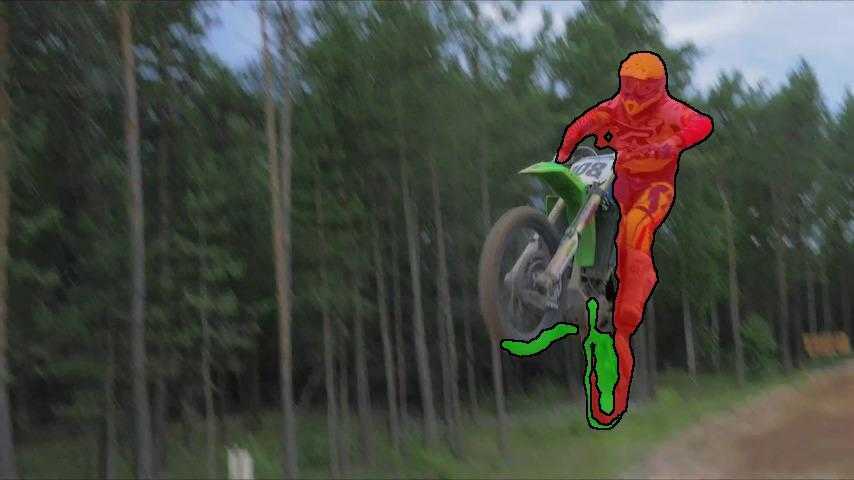} & {\footnotesize{}}
\includegraphics[width=0.15\linewidth]{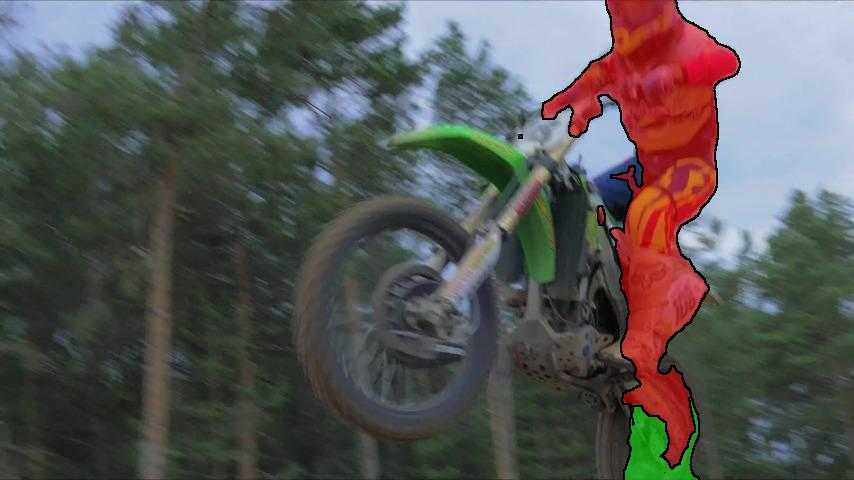} & {\footnotesize{}}
\includegraphics[width=0.15\linewidth]{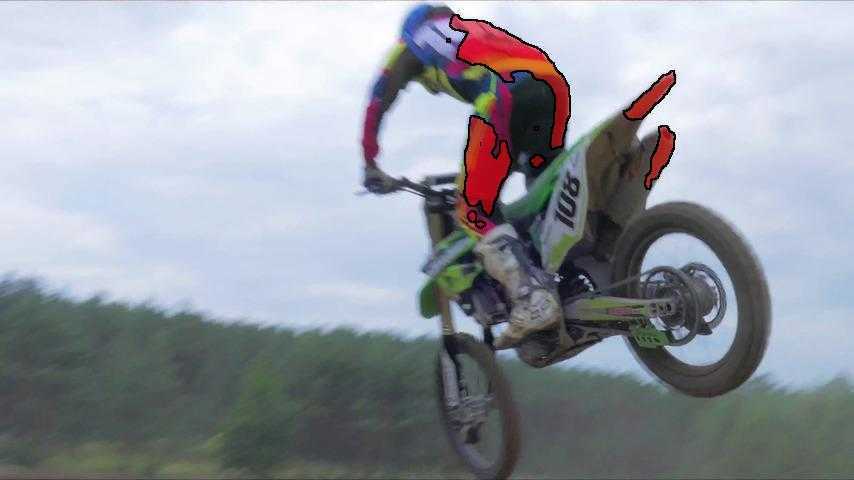} & {\footnotesize{}}
\includegraphics[width=0.15\linewidth]{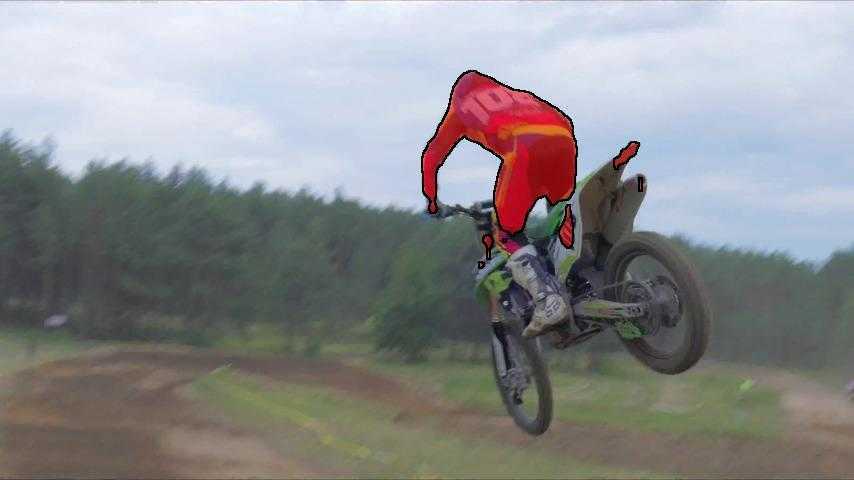} & {\footnotesize{}}
\includegraphics[width=0.15\linewidth]{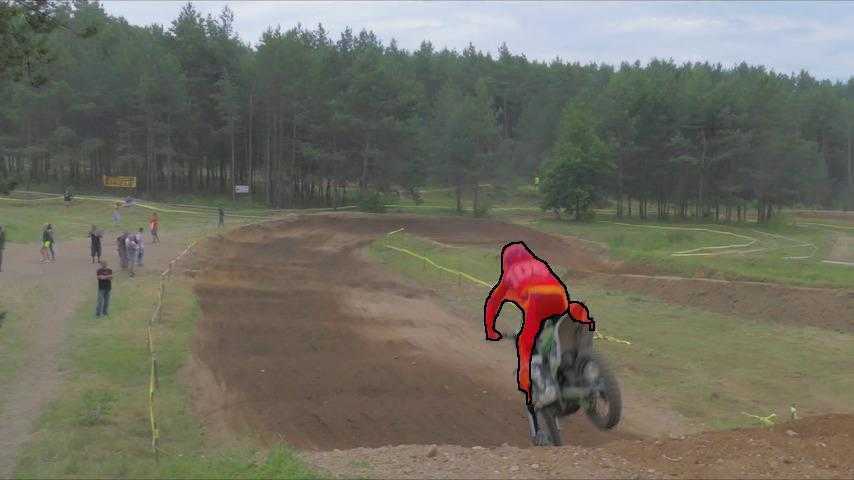}\tabularnewline

\multicolumn{6}{c}{\textit{\vspace{0.05em} }} \tabularnewline

\multicolumn{6}{c}{
\begin{tabular}{c} 
Language supervision, \textit{ \textcolor{mygreen}{ID 1}: "A black scooter ridden by a man".}  \tabularnewline
\textit{\textcolor{myred}{  ID 2}: "A man in a suit riding a scooter".}  \tabularnewline 
\end{tabular}
} \tabularnewline
\includegraphics[width=0.15\linewidth]{figures/VOS_qual/scooter-black_17/00001} & {\footnotesize{}}
\includegraphics[width=0.15\linewidth]{figures/VOS_qual/scooter-black_17/00011} & {\footnotesize{}}
\includegraphics[width=0.15\linewidth]{figures/VOS_qual/scooter-black_17/00025} & {\footnotesize{}}
\includegraphics[width=0.15\linewidth]{figures/VOS_qual/scooter-black_17/00030} & {\footnotesize{}}
\includegraphics[width=0.15\linewidth]{figures/VOS_qual/scooter-black_17/00037} & {\footnotesize{}}
\includegraphics[width=0.15\linewidth]{figures/VOS_qual/scooter-black_17/00042}\tabularnewline
\multicolumn{6}{c}{{Pixel-level mask supervision }} \tabularnewline
\includegraphics[width=0.15\linewidth]{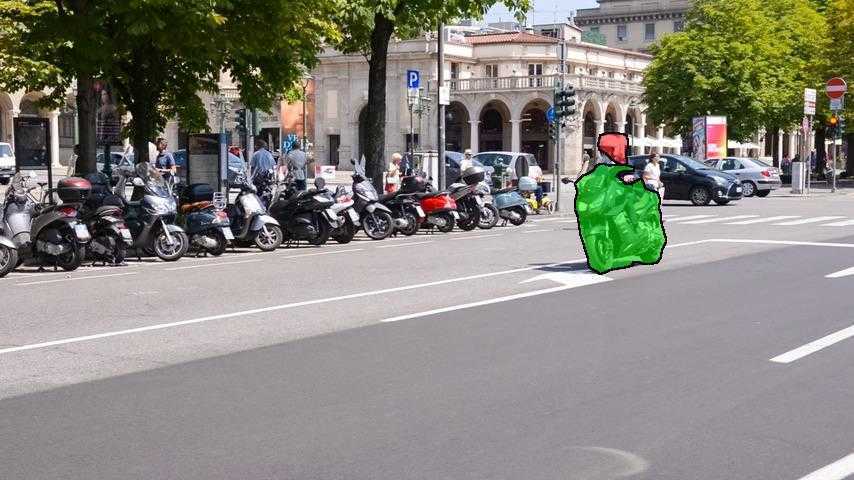} & {\footnotesize{}}
\includegraphics[width=0.15\linewidth]{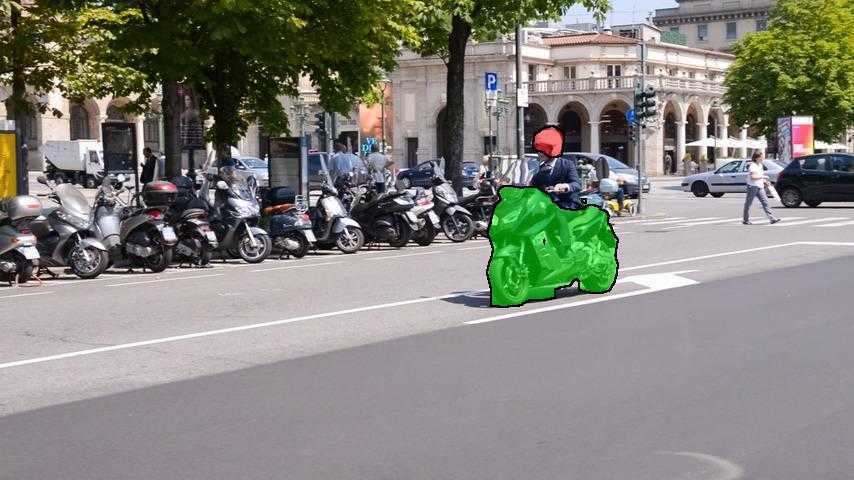} & {\footnotesize{}}
\includegraphics[width=0.15\linewidth]{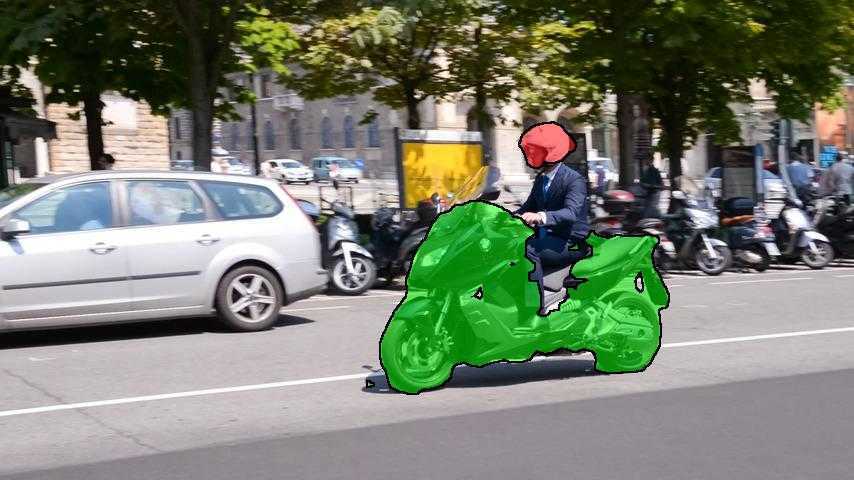} & {\footnotesize{}}
\includegraphics[width=0.15\linewidth]{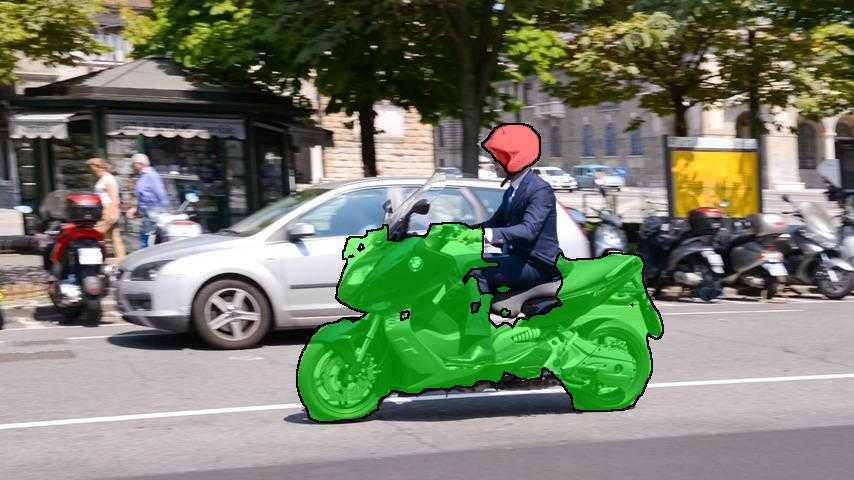} & {\footnotesize{}}
\includegraphics[width=0.15\linewidth]{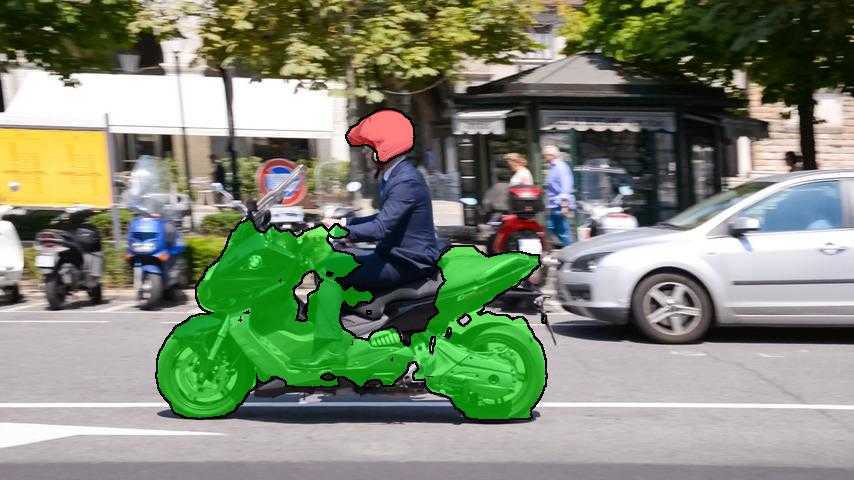} & {\footnotesize{}}
\includegraphics[width=0.15\linewidth]{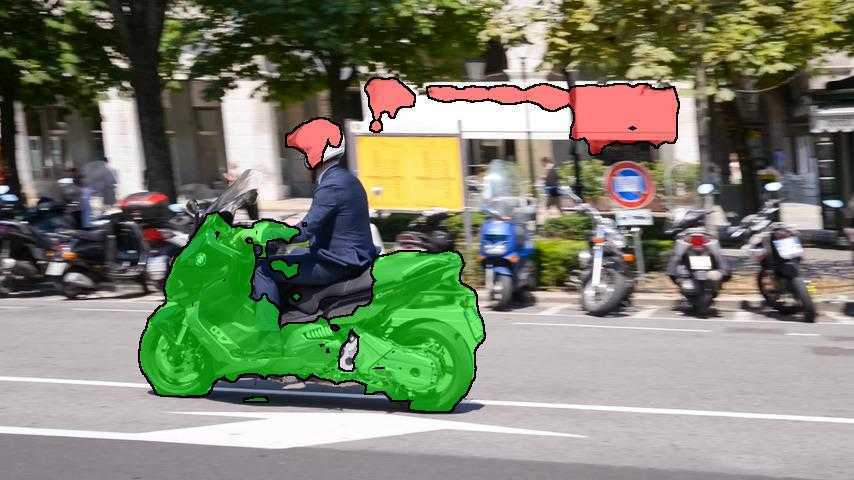}\tabularnewline

\end{tabular}\hfill{}
\par\end{centering}

\caption{\label{fig:sup_qualitative-results-mask-vos} Video object segmentation results using Language versus Mask on the 1st frame as supervision
on $\text{DAVIS}_{\text{16}}$ and $\text{DAVIS}_{\text{17}}$, val sets. Using language only  results in a more robust performance for videos with similar looking instances and camera view changes in comparison to employing pixel-level masks.
Frames sampled along
the video duration. The videos are chosen with the highest mIoU difference.}
\vspace{0em}
\end{figure*}

Figure \ref{fig:sup_qualitative-results-mask-vos} shows comparison of Language versus Mask supervision on $\text{DAVIS}_{\text{16}}$ and $\text{DAVIS}_{\text{17}}$, validation sets.
Note that using only language supervision results in a more robust performance for videos with similar looking instances and camera view changes in comparison to employing pixel-level masks.

\end{document}